\def\buildarxiv{}
\newif\iftmlr
\ifdefined\buildarxiv \tmlrfalse \else \tmlrtrue \fi
\newif\iffinal
\finaltrue
\documentclass[10pt]{article}
\iftmlr\usepackage{tmlr}\else\usepackage[preprint]{tmlr}\fi
\usepackage[utf8]{inputenc}
\usepackage[T1]{fontenc}
\usepackage[english]{babel}
\usepackage{amsmath,amssymb,amsthm}
\usepackage{booktabs,graphicx,array,multirow,longtable}
\usepackage{adjustbox}
\usepackage{enumitem}
\usepackage[section]{placeins}
\usepackage{xcolor}
\usepackage{microtype}
\usepackage[hidelinks]{hyperref}
\usepackage{cleveref}
\usepackage{url}
\hypersetup{pdftitle={Frozen Factor or Spectral Band? In LoRA, the Factor Matters More at Low Rank},
  pdfauthor={}, pdfsubject={}, pdfkeywords={}, pdfcreator={}, pdfproducer={}}
\ifdefined\pdfinfoomitdate \pdfinfoomitdate=1 \fi
\ifdefined\pdftrailerid \pdftrailerid{} \fi
\ifdefined\pdfsuppressptexinfo \pdfsuppressptexinfo=-1 \fi

\iffinal
\makeatletter
\DeclareRobustCommand{\todo}[1]{\PackageError{final}{A TODO remains: \detokenize{#1}}{Resolve every todo before setting finaltrue.}\@@end}
\makeatother
\else
\newcommand{\todo}[1]{\textcolor{red}{\textbf{[TODO: #1]}}}
\fi
\iffinal
\makeatletter
\newcommand{\paperfig}[2][]{\IfFileExists{#2}{\includegraphics[#1]{#2}}{\PackageError{final}{Missing figure file: \detokenize{#2}}{Add every figure.}\@@end}}
\makeatother
\else
\newcommand{\paperfig}[2][]{\IfFileExists{#2}{\includegraphics[#1]{#2}}{\fbox{\parbox[c][2.5cm][c]{0.9\linewidth}{\centering\small missing figure file:\\ \texttt{\detokenize{#2}}}}}}
\fi

\newcommand{\block}[1]{\def\thisblock{#1}\ifx\thisblock\wantedblock\blockbegin}
\newcommand{\blockbegin}{\ifhmode\unskip\unskip\space\fi\ignorespaces}
\let\endblock\fi
\newwrite\secposfile
\immediate\openout\secposfile=\jobname.secpos
\AtBeginDocument{\immediate\write\secposfile{textheight \number\textheight}%
  \immediate\write\secposfile{texttop \number\dimexpr\paperheight-1in-\voffset-\topmargin-\headheight-\headsep\relax}}
\newcommand{\secmark}[1]{\pdfsavepos\write\secposfile{mark #1 \thepage\space\the\pdflastypos}}
\newcommand{\Crefresults}{\Cref{sec:rank}}

\iftmlr\setlist[itemize]{itemsep=0pt,topsep=3pt}\fi

\newcommand{\posthoc}{\textsuperscript{\dag}}

\title{\bfseries Frozen Factor or Spectral Band? In LoRA, the Factor Matters More at Low Rank\thanks{The authors used an AI assistant to write and test the experiment code and to check that every number in the text matches the logged results. The research question, the design decisions and the claims are the authors'. All results come from the runs described in the paper.
}}
\iftmlr
\author{\name Anonymous authors \\ \addr Paper under double-blind review}
\else
\author{\name Adnan Slimane Ali \email adnan.slimane-ali@student-cs.fr \\ \addr CentraleSup\'elec, Universit\'e Paris-Saclay
\AND \name David Ngwe Pouth \email david.ngwepouth@student-cs.fr \\ \addr CentraleSup\'elec, Universit\'e Paris-Saclay}
\fi
\date{}

\begin{document}
\maketitle

\begin{abstract}\noindent
On two main task--model pairs, at ranks 1 to 8 with few optimiser steps, the factor a LoRA variant freezes matters more than its band of singular directions.
MiCA and PiCa are compared by band, minor against dominant, yet MiCA freezes the output factor $B$ and PiCa, in our notation, the input factor $A$.
We cross factor and band in a $2 \times 2$ design on models from 1B to 8B parameters.
Freezing $A$ beats freezing $B$ on both bands, by $+12$ to $+20$ points at rank 2 on a format task, our primary evidence, and still by about $+7$ points at rank 8 on OpenBookQA.
A Holm correction over 258 primary tests, added post hoc, keeps this advantage in 15 of 16 tests and the lead of factor over band at ranks 1, 2 and 4, rank 1 of the format task being itself post hoc.
At rank 8 that lead passes its own test but not the correction.
The advantage repeats at rank 2 with two 7B models, by $+11$ to $+13$ points.
Outside the correction, it repeats on GSM8K with a 1B model and, in a post hoc experiment, with an 8B model at ranks 8 and 16.
It holds when $B$ frozen gets as many parameters or more, with a random frozen $A$, and with sixteen times the format-task budget.
It narrows above rank 8 on most pairs, at rank 8 with the 8B model on the two main tasks, on ARC and in a task mixture, and it closes at the last step of longer training on OpenBookQA.
At MiCA's and PiCa's own settings, it is smaller or not detected.
\end{abstract}

\let\layoutFloatBarrier\FloatBarrier\let\FloatBarrier\relax
\secmark{intro}
\section{Introduction}
\label{sec:intro}

LoRA adapts a pretrained weight by adding a low-rank update, the product of two factors
\citep{hu2021lora}.
Frozen-subspace variants fix one factor on a band of singular directions of that weight and train
the other.

MiCA \citep{rudiger2026mica} and PiCa \citep{hwang2026pica} differ in factor and band.
MiCA freezes the output factor $B$ on the minor band, with the smallest singular values.
PiCa, in our notation, freezes the input factor $A$ on the dominant band, with the largest.
The two methods are usually compared by their band, yet the comparison changes two things at
once.

We ask which of the two explains a difference in accuracy at low rank: the frozen factor or
the band?

We cross the two choices: we freeze $A$ or $B$, on the dominant or the minor band
(\Cref{fig:design}).
Inside one experiment, the four configurations share the task, model, rank, training data and
number of training steps.
The trainable-parameter count is not, and we control for it in \Cref{sec:robustness}.
A free LoRA, with both factors trained, is the reference.
We run the design on two main pairs: OpenBookQA with Qwen2.5-1.5B, and a format task with
Llama-3.2-1B.
The format task asks for a fixed JSON structure that the base model never produces.
On the format task at rank 2, $A$ frozen on PiCa's band reaches 77.08\% and $B$ frozen on MiCA's
band reaches 57.92\%.
The gap is 19 points.
Changing only the band moves $A$ frozen by 1 point and $B$ frozen by 7 points.
Changing only the factor moves accuracy by 12 to 20 points.

At ranks 1 to 8 and with short training, the frozen factor matters more.
We find five things.
\Cref{tab:statusterms} defines the status words, such as established.
\begin{itemize}[nosep]
\item \textbf{At ranks 1 to 8, $A$ frozen beats $B$ frozen on both bands of the two main pairs.}
  At rank 8 on OpenBookQA, the advantage is still about $+7$ points on both bands (\Cref{sec:rank}).
\item \textbf{At ranks 1, 2 and 4, the factor matters more than the band.}
  On both main pairs, that verdict survives a post hoc Holm correction.
  At rank 8 it is established and not Holm-robust, and rank 1 of the format task is post hoc.
\item \textbf{The advantage repeats on larger models and on a reasoning task.}
  At rank 2 on OpenBookQA with Qwen2.5-7B and Mistral-7B, $A$ frozen leads by $+11.02$ to $+13.15$
  points, all four Holm-robust.
  On GSM8K with Llama-3.1-8B, post hoc, the factor leads the band at ranks 8 and 16.
\item \textbf{At rank 2, the advantage survives our controls on parameters and seeds, and a post hoc
  control on the learning rate.}
  It also holds with a random frozen $A$ (\Cref{sec:robustness}).
\item \textbf{At MiCA's and PiCa's own settings, the advantage is smaller or not detected.}
  At rank 16 with MiCA's recipe, flipping its frozen factor helps on the format task only.
  At rank 256 with PiCa's recipe, we detect no advantage of $A$ frozen (\Cref{sec:methods}).
\end{itemize}

\begin{figure}[!ht]
\centering
\paperfig[width=0.95\linewidth]{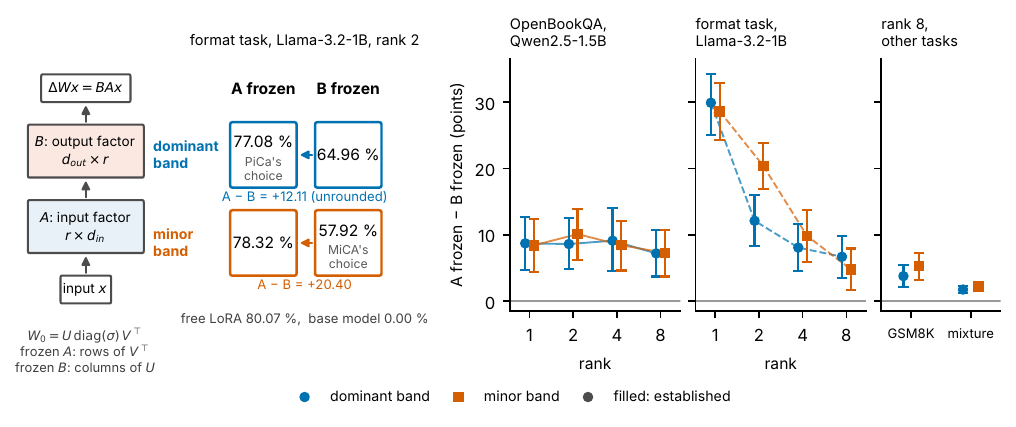}
\caption{$A$ frozen beats $B$ frozen on both bands at ranks 1, 2, 4 and 8 on both main pairs.
\emph{Left:} a LoRA layer with $W_0 = U \operatorname{diag}(\sigma) V^\top$.
A frozen $A$ takes rows of $V^\top$, and a frozen $B$ takes columns of $U$.
\emph{Middle:} the four configurations on the format task with Llama-3.2-1B at rank 2.
Each cell is the mean test accuracy of eight seeds, in percent.
The column, the factor, moves accuracy more than the row, the band.
The base model scores 0\% and the free LoRA scores 80.07\%.
\emph{Right:} A $-$ B, the accuracy of $A$ frozen minus that of $B$ frozen, in points, against
the rank.
Each bar is the interval of one test, and the last panel shows two other tasks at rank 8.
In every figure, circles are the dominant band and squares the minor band.
A filled mark is an established difference (\Cref{tab:statusterms}).
Rank 1 of the format task is post hoc.}
\label{fig:design}
\end{figure}

\paragraph{Scope.}
The finding holds at ranks 1 to 8, with short training, on the two main pairs and in the rank-2
replications.
Short training means 49 optimiser steps on OpenBookQA and 274 on the format task.
The format task is our primary evidence, because we formed the hypothesis on the OpenBookQA pair.
On the format task, the advantage persists with 16 times the budget.
Outside that range the finding narrows (\Cref{sec:narrow}).
Above rank 8, A $-$ B is established mostly on the minor band.
On OpenBookQA with Qwen2.5-1.5B, the observed band effect exceeds the factor effect at
ranks 16, 32 and 64.
On OpenBookQA at rank 2, the gap closes at the last step of a four times longer budget, and we
did not test why.
\iftmlr\else
\Cref{app:supported} shows how that result depends on the protocol.
\fi%
With Llama-3.1-8B, our largest model, A $-$ B is established on one band of each main task only.
On ARC and on a mixture of eight tasks, the advantage is small.
We did not test an interaction between rank and factor.

\paragraph{Contributions.}
We propose no new method, and the advantage of training $B$ over $A$ is known
\citep{zhu2024asymmetry}.
We contribute a crossed design that separates the frozen factor from the band, with a learning
rate selected for each configuration and a free LoRA of matched size.
We add the evidence above, with the places where it stops, and an audit of MiCA and PiCa at their
own ranks and recipes.
\Cref{sec:background,sec:protocol} set up the comparison, \Cref{sec:rank,sec:methods,sec:robustness}
give the results, and \Cref{sec:discussion} the limits.
\iftmlr\else
The appendices give the statistics and the learning-rate rules (\Cref{app:stats}), every contrast
(\Cref{app:contrasts}) and every setting and rank (\Cref{app:rankexp}).
They also give the controls (\Cref{app:robust}), the overlap analysis (\Cref{sec:mechanism}) and the
data, budgets and scoring rules (\Cref{app:data}).
\def\wantedblock{code}\immediate\write\secposfile{block arxiv_only code}\block{claimrank2}
At rank 2, freezing the output factor $B$ costs more than the choice of band on our two main pairs, on OpenBookQA with Qwen2.5-7B, Mistral-7B and Qwen3-1.7B-Base, and on the format task with Qwen2.5-7B\posthoc.
In these settings, A $-$ B is established on both bands, and $A$ frozen on either band beats $B$ frozen on the other band ($F$ exceeds $|S|$).
With the instruction-tuned Qwen2.5-1.5B on OpenBookQA, A $-$ B is established on the dominant band only, and $F$ exceeds $|S|$.
On ARC with Qwen2.5-1.5B, A $-$ B is established on the minor band only, and $F$ against $|S|$ is not established.
With Llama-3.1-8B at rank 8, A $-$ B is established on one band of each task, and $F$ against $|S|$ is not established.
\endblock
\block{claimranks}
At ranks 1 to 8 on our two main pairs (the format task at rank 1\posthoc), freezing the output factor $B$ costs more than the choice of band (established).
On OpenBookQA, this holds at the standard 49 optimiser steps (600,000 supervised tokens, about 5.8 passes over the packed training set), where $B$ frozen stays near the base model, and the gap is small under the answer-only generation protocol (no test; \Cref{app:oc}).
At the higher ranks tested, the effect is not established on most bands.
\endblock
\block{code}
Code and logs are available from the authors on request (\Cref{sec:repro}).
\endblock
\block{repro}
Code and logs are available from the authors on request: the code, the protocol of every experiment and the reading of every experiment.
Each protocol was committed before the first run of its experiment (\Cref{sec:prespec}).
A reading is a text file from which the numbers of this paper are taken.
A script checks the numbers of the paper against these readings before the paper is built.
\endblock
\block{ack}

\endblock
\block{contributions}

\endblock
\block{ptrprot}
\Cref{app:supported} shows how this result depends on the protocol.
\endblock
\block{guide}
The appendices give the statistics and the rules for the learning rate (\Cref{app:stats}), every contrast (\Cref{app:contrasts}), every robustness check and control (\Cref{app:robust}), every setting and rank (\Cref{app:rankexp}), the overlap analysis (\Cref{sec:mechanism}) and the data, budgets and scoring rules (\Cref{app:data}).
\endblock
\block{ptrrelated}
\Cref{app:related} discusses further related work, on VeRA, DoRA and small singular values.
\endblock
\block{ptrposthoc}
\Cref{tab:posthoc} lists the analyses decided after seeing the results, and \Cref{app:stats} says how to read the mark \dag.
\endblock
\block{ptrrank2}
\paragraph{Every setting at rank 2.}
\Cref{fig:forest} shows A $-$ B, $F$ and $S$ at rank 2 for every task and model, and \Cref{app:contrasts} gives every contrast with its interval.
\Cref{app:rankexp} gives the crossed pairs, ARC, the instruction-tuned model, Qwen3-1.7B-Base and the eight-task mixture.
\endblock
\block{ptrranks}
\Cref{app:rankexp} gives every setting and rank: the replication at ranks 16 and 32, the LoRA scale at rank 16, Mistral-7B up to rank 64 and the eight-task mixture.
\Cref{app:collapse} gives the diverged run at rank 16 on the format task, and \Cref{app:notest} the bounds where A $-$ B is not established.
With four times the training budget at rank 16, A $-$ B stays established on both bands of the format task ($+4.13$ and $+4.11$ points) and is not established on OpenBookQA (\Cref{app:cvr16}).
\endblock
\block{ptrmethods}
\Cref{app:methodarms} gives them with their parameter counts.
\endblock
\block{ptrrobust}
\paragraph{Side effects of freezing $A$.}
When adapters are merged on the minor band, $A$ frozen loses more accuracy on its own task than $B$ frozen.
\Cref{app:robust} gives this result and the other side effects: forgetting, intruder dimensions and transfer to a neighbouring task.
It also gives the controls on the training setup (\Cref{app:controls}) and two results that go against parts of our reading.
\endblock
\block{ptrholds}
\Cref{app:holds} lists the settings where the main result holds.
\endblock
\block{ptrmech}
\Cref{app:overlap,app:mech,app:gradsides} give these analyses.
\endblock

\fi%

\secmark{background}
\section{Background and related work}
\label{sec:background}

What exactly differs between MiCA and PiCa, and what did earlier work settle?

\paragraph{LoRA and its two factors.}
LoRA \citep{hu2021lora} keeps a pretrained weight $W_0$ of shape
$d_{\text{out}} \times d_{\text{in}}$ fixed and adds a low-rank update of rank $r$:
\[
  W = W_0 + \Delta W, \qquad \Delta W = BA .
\]
The matrix $A$ has shape $r \times d_{\text{in}}$ and reads the layer input, so we call it the
\emph{input factor}.
The matrix $B$ has shape $d_{\text{out}} \times r$ and writes to the output, so we call it the
\emph{output factor}.
Freezing a factor means that we set it before training and never update it, while the other
factor trains from zero.

\paragraph{Bands.}
Write the singular value decomposition $W_0 = U \operatorname{diag}(\sigma) V^\top$, with
$\sigma$ in decreasing order.
\iftmlr\else
The columns of $U$ are directions of the output space, and the rows of $V^\top$ are directions of
the input space.
\fi%
A \emph{band} is a set of $r$ singular directions of $W_0$.
The \emph{dominant band} holds the first $r$ directions, with the largest singular values.
The \emph{minor band} holds the last $r$ directions, however small their singular values.

\paragraph{MiCA and PiCa differ in two ways.}
MiCA \citep{rudiger2026mica} freezes $B$ on the minor band and trains $A$.
PiCa \citep{hwang2026pica} freezes the $r$ leading singular vectors $U_r$ and trains a factor
shared across layers.
Its paper writes $Y = X(W_0 + U_r B)$, with the inputs $X$ multiplied from the left.
The frozen $U_r$ therefore sits on the input side.
In our notation, PiCa freezes $A$ on the dominant band (\Cref{app:methodarms}).
\begin{center}
\small
\begin{tabular}{lll}
\toprule
Method & Frozen factor & Band \\
\midrule
MiCA & output factor $B$ & minor \\
PiCa & input factor $A$ & dominant \\
\bottomrule
\end{tabular}
\end{center}
Moving from one row to the other changes both columns, so the comparison alone
cannot say which matters.

\paragraph{Related work.}
One line of work freezes a factor.
\citet{zhu2024asymmetry} showed that training $B$ matters more than training $A$, and that a random,
untrained $A$ performs close to a trained one.
LoRA-FA \citep{zhang2023lorafa} freezes a random $A$ to save memory, and FFA-LoRA
\citep{sun2024ffalora} freezes $A$ for privacy in federated learning.
Our factor result points the same way, so the asymmetry itself is not new.
Our addition is to cross factor and band.
A second line chooses a band and trains both factors.
PiSSA \citep{meng2024pissa} starts from the dominant components and MiLoRA
\citep{wang2024milora} from the minor ones.
LoRA-RLPO and LoRA-RLMO \citep{rlvrortho2026} are two further initialisations of that kind, and
our $A$ frozen starts from them.
A third line freezes both factors.
LoRA-XS \citep{balazy2024loraxs} trains a small $r \times r$ matrix between them, and LoRA-SB
\citep{ponkshe2024lorasb} takes its frozen factors from the first update (\Cref{app:core}).
\iftmlr\else
\Cref{sec:mechanism} asks whether the input basis or the output basis limits such an adapter.
\fi%

\secmark{protocol}
\section{Protocol}
\label{sec:protocol}

How do we compare the four configurations fairly, and when do we call a difference real?

\subsection{What is compared: four frozen configurations and a free LoRA}
\label{sec:configs}

We compare four frozen configurations, which cross the frozen factor with the band.
\emph{$B$ frozen on a band} sets $B$ to the $r$ columns of $U$ in that band and trains $A$ from
zero.
\emph{$A$ frozen on a band} sets $A$ to the $r$ rows of $V^\top$ in that band and trains $B$ from
zero.
$B$ frozen on the minor band is MiCA's configuration.
$A$ frozen on the dominant band has the factor and the band that PiCa freezes.

The reference is the \emph{free LoRA}: both factors train, from LoRA's usual initialisation
\citep{hu2021lora}.
The \emph{matched free LoRA} is a free LoRA with about the same trainable-parameter count as
$B$ frozen (\Cref{app:data}).
In \Cref{sec:methods} only, it matches the method under comparison instead.
\iftmlr\else
On the format task, it is the free LoRA with the smallest count at least as large as that of
$B$ frozen.
We chose that reference because one of our questions compares MiCA's configuration with a LoRA of
about the same size.
\Cref{app:data} gives the ratios both ways.
\fi%

Every configuration adapts the seven linear projections of every block, with scaling
$\alpha/r = 1$.
One experiment varies only the frozen factor and the band, under one rule that picks the learning
rate.
Each rank is a separate experiment, never pooled, with its own learning-rate selection and its own
tests.

\subsection{Is the comparison fair in budget and parameters?}
\label{sec:budgets}

The training budget is equal across configurations, and the parameter count is not.
The \emph{training budget}, the number of optimiser steps, is 49 on OpenBookQA and 274 on the
format task.
Both budgets are short in optimiser steps.
On OpenBookQA, the 49 steps still make about 5.8 passes over the packed training set.
\iftmlr\else
The 49 steps see 600,000 supervised tokens.
\fi%

The \emph{parameter count} is the number of trainable parameters.
Per module, $A$ frozen trains the $r\,d_{\text{out}}$ entries of $B$, and $B$ frozen trains the
$r\,d_{\text{in}}$ entries of $A$.
Summed over the adapted modules, $A$ frozen trains more parameters than $B$ frozen at equal rank,
and more than the matched free LoRA.
The cause is the large output of the up and gate projections of the MLP.
\iftmlr\else
On Qwen2.5-1.5B at rank 2, $A$ frozen trains 1,290,240 parameters and $B$ frozen 1,017,856.
\fi%
The extra parameters are a possible advantage for $A$ frozen.
We check them on each task in \Cref{sec:robustness}, and $A$ frozen stays ahead.
$A$ frozen at rank 1 also beats $B$ frozen at rank 2 with fewer trainable parameters, on both
tasks and both bands (\Cref{tab:rank1budget}).
That observation is descriptive, post hoc and crosses experiments.
\iftmlr\else
On OpenBookQA, $A$ frozen at rank 1 trains 645,120 parameters, against 1,017,856 for $B$ frozen at
rank 2.
Its mean accuracy on the dominant band is 53.50\%, against 44.65\% for $B$ frozen at rank 2.
\fi%

\subsection{How learning rates are chosen}
\label{sec:rates}

We select the learning rate separately for every configuration, rank, task and model, with rules
fixed in advance.
A shared rate would be unfair, because \citet{lee2026learningrate} find that the best rate differs
between LoRA variants.
Most selections start from six candidate rates, trained with seeds 0 and 1.
We score them on a validation set disjoint from the test set and keep the rate with the best
validation mean.
Fixed rules handle grid edges and near-ties (\Cref{app:stats}).
Over all experiments, 128 selections ended on an unresolved tie and 34 at the grid edge (\Cref{tab:ties}).
In the eight-task mixture at ranks 2, 16 and 32, rerunning the other tied rate leaves A $-$ B positive in all six cases
(\Cref{app:othertied}).
\iftmlr\else
\paragraph{The grid and the four rules.}
In most experiments the six candidates are $10^{-4}$, $5 \times 10^{-4}$, $10^{-3}$,
$2 \times 10^{-3}$, $5 \times 10^{-3}$ and $10^{-2}$.
Four rules choose among them.
\def\wantedblock{ratesrules}\immediate\write\secposfile{block sec3_protocol ratesrules}\block{holm}
\paragraph{The same tests after a Holm correction.}
\label{app:holm}
The body reads each test at its own level.
Here we apply Holm's correction, post hoc, to the 258 primary tests, at $0.05$.
Of the 151 tests established at their own level, 117 survive and 34 do not (\Cref{tab:holm}).
The low-rank result survives.
The verdict \emph{factor} survives at ranks 1, 2 and 4 on both main pairs.
A $-$ B survives on both bands at rank 2 with Qwen2.5-7B and with Mistral-7B.
At rank 8 on the main pairs, A $-$ B survives on three of the four bands (all but the minor band of the format task), and the verdict \emph{factor} does not survive on either pair.
Flipped MiCA does not survive on the format task.
GSM8K, the replication's tests of A $-$ B at ranks 16 and 32, the longer budgets on the format task, longer training at rank 16 (\Cref{app:cvr16}), flipped MiCA with a 7B model (\Cref{app:flip7}) and the test on the input side (\Cref{app:ig}) are outside the family and keep the level of their own tests.
Every bootstrap p-value under $10^{-3}$ was recomputed with 1,000,000 resamples: with 10,000, the p-values move in steps of $2 \times 10^{-4}$, which is coarser than Holm's first threshold, $0.05/258$.
Each regenerated interval reproduces the printed one; for two of them a bound differs by $0.007$, within one unit of the last printed digit.
The correction changes no status in the text: ``established'' always refers to a test's own level.

\begin{table}[h!]
\centering\small
\caption{\textbf{Status of the main results cited in the body, after the Holm correction.}
``Established only'': established at the level of its own test and not Holm-robust.
``Outside'': the test is not among the 258 primary tests and is read at its own level.}
\label{tab:status}
\begin{tabular}{p{8.6cm}p{6.2cm}}
\toprule
Result & Status \\
\midrule
A $-$ B, main pairs, ranks 1, 2, 4 and 8 (16 tests) & 15 Holm-robust; format task, rank 8, minor band: established only \\
Factor against band, main pairs & Holm-robust at ranks 1, 2 and 4; established only at rank 8 \\
A $-$ B at rank 2, OpenBookQA with Qwen2.5-7B and Mistral-7B (4 tests) & Holm-robust \\
A $-$ B at rank 2, task and model crossed (4 tests) & 3 Holm-robust, 1 established only \\
A $-$ B on ARC, minor band; factor against band, instruction-tuned model & established only \\
A $-$ B above rank 8, minor band: Qwen2.5-7B at rank 16, Mistral-7B at rank 32, mixture at ranks 16 and 32 & Holm-robust \\
A $-$ B with Mistral-7B at rank 64 & established only \\
Flipped MiCA minus MiCA, format task & established only \\
Free LoRA minus MiCA's configuration (\Cref{tab:gap}) & OpenBookQA at rank 16: Holm-robust; the three other rows of the main pairs: established only; the 7B rows: outside \\
LoRA+ against $A$ frozen & established only \\
GSM8K with both models; Llama-3.1-8B; random bases on eight seeds; longer budgets on the format task; longer training at rank 16 (\Cref{app:cvr16}), flipped MiCA with a 7B model (\Cref{app:flip7}) and the test on the input side (\Cref{app:ig}) & outside \\
Format task at rank 1; factor against band on three pairs & among the 258 tests; post hoc, tests fixed after related results \\
Every learning rate of $B$ frozen & outside; post hoc \\
\bottomrule
\end{tabular}
\end{table}

\begin{table}[h!]
\centering
\scriptsize
\caption{\textbf{The 34 established tests that do not survive the Holm correction.}
Estimates are in accuracy points, or a correlation for the overlap tests; $F$ against $|S|$ gives its two cross differences.
The adjusted p is the Holm-adjusted value over the 258 tests.}
\label{tab:holm}
\begin{tabular}{llp{5.6cm}ll}
\toprule
Task and model & rank & test & estimate & adjusted p \\
\midrule
Format task, Llama-3.2-1B & 16 & Free LoRA $-$ MiCA's configuration & $+4.78$ & $0.0505$ \\
Format task, Llama-3.2-1B & -- & Oracle directions for $A$ & $+4.82$ & $0.057$ \\
OpenBookQA, Qwen2.5-1.5B & -- & A $-$ B, minor band, one eighth of the data & $+5.02$ & $0.070$ \\
OpenBookQA, Mistral-7B & 64 & A $-$ B, minor band & $+4.88$ & $0.087$ \\
Format task, Llama-3.2-1B & 8 & A $-$ B, minor band & $+4.78$ & $0.100$ \\
OpenBookQA, Llama-3.2-1B & -- & A $-$ B, minor band & $+5.37$ & $0.117$ \\
Eight-task mixture, Qwen2.5-1.5B & 16 & A $-$ B, dominant band & $+0.73$ & $0.135$ \\
OpenBookQA, Qwen2.5-1.5B & 32 & Forgetting, $B$ frozen $-$ $A$ frozen, minor band & $-3.61$ & $0.135$ \\
Format task, Llama-3.2-1B & -- & Activation-band control, second test & $+15.09$ & $0.145$ \\
OpenBookQA, Llama-3.2-1B & -- & $F$ against $|S|$ & $+7.60$, $+6.07$ & $0.150$ \\
OpenBookQA, Mistral-7B & 32 & $F$ against $|S|$ & $+7.83$, $+4.97$ & $0.157$ \\
Format task, Llama-3.2-1B & 2 & Input-side overlap correlation (not the input-side test of \Cref{app:ig}) & $+0.703$ & $0.182$ \\
OpenBookQA, Qwen2.5-1.5B & -- & cost of freezing $B$ in attention modules & $+2.78$ & $0.182$ \\
ARC-Easy, Qwen2.5-1.5B & 2 & A $-$ B, minor band, over the rates of $B$ frozen (a test of the ARC experiment, not the control of \Cref{tab:robustdraft}) & $+1.63$ & $0.230$ \\
Format task, Llama-3.2-1B & 8 & $F$ against $|S|$ & $+6.43$, $+5.03$ & $0.254$ \\
ARC-Challenge, Qwen2.5-1.5B & 2 & A $-$ B, minor band & $+3.18$ & $0.277$ \\
ARC-Easy, Qwen2.5-1.5B & 2 & A $-$ B, minor band & $+1.63$ & $0.277$ \\
Format task, Llama-3.2-1B & -- & Flipped MiCA $-$ MiCA, same rank & $+3.48$ & $0.277$ \\
Format task, Llama-3.2-1B & -- & Stitched factors, second test & $-6.03$ & $0.277$ \\
OpenBookQA, Mistral-7B & 32 & A $-$ B, dominant band & $+4.38$ & $0.277$ \\
Eight-task mixture, Qwen2.5-1.5B & 2 & $F$ against $|S|$ & $+2.14$, $+0.98$ & $0.484$ \\
ARC-Easy, Qwen2.5-1.5B & -- & A $-$ B, minor band, best checkpoint & $+1.48$ & $0.528$ \\
Format task, Llama-3.2-1B & -- & Flipped MiCA $-$ MiCA, MiCA's parameter count & $+3.50$ & $0.528$ \\
OpenBookQA, Qwen2.5-1.5B & -- & Core-only control, second test & $+2.98$ & $0.543$ \\
OpenBookQA, Qwen2.5-1.5B & -- & Activation-band control, third test & $+3.88$ & $0.679$ \\
OpenBookQA, Qwen2.5-1.5B & 32 & Free LoRA $-$ MiCA's configuration & $+4.25$ & $0.789$ \\
OpenBookQA, Qwen2.5-1.5B & 8 & $F$ against $|S|$ & $+9.88$, $+4.55$ & $0.789$ \\
OpenBookQA, Qwen2.5-1.5B & -- & Free LoRA $-$ MiCA (PEFT), tuned rates & $+4.00$ & $0.798$ \\
Format task, Llama-3.2-1B & 16 & $F$ against $|S|$ & $+4.11$, $+3.62$ & $0.949$ \\
OpenBookQA, Qwen2.5-1.5B-Instruct & -- & $F$ against $|S|$ & $+5.83$, $+6.80$ & $0.986$ \\
Format task, Llama-3.2-1B & 32 & Forgetting, free LoRA $-$ $B$ frozen, minor band & $-2.93$ & $0.999$ \\
Format task, Llama-3.2-1B & -- & $A$ frozen (dominant) $-$ LoRA+ free LoRA & $-3.01$ & $1.000$ \\
Format task, Llama-3.2-1B & 32 & Free LoRA $-$ MiCA's configuration & $+3.09$ & $1.000$ \\
OpenBookQA, Qwen2.5-1.5B & 8 & Output-overlap correlation & $+0.524$ & $1.000$ \\
\bottomrule
\end{tabular}
\end{table}
\endblock
\block{opening}
In each experiment we fix the task, the model, the rank, the training data, the number of training steps and the rule that picks the learning rate.
We vary only which factor is frozen and on which band.
The design has four frozen configurations (two factors crossed with two bands) and a free LoRA, whose two factors both train as the reference.
We run them on the two main pairs of \Cref{sec:intro} (\Cref{sec:tasks}).
\endblock
\block{configs}
LoRA adds a low-rank update $\Delta W = BA$ to a pretrained weight $W_0$ of shape $d_{\text{out}} \times d_{\text{in}}$.
The factor $A$ ($r \times d_{\text{in}}$) reads the input of the layer, and we call it the \emph{input factor}.
The factor $B$ ($d_{\text{out}} \times r$) writes to the output, and we call it the \emph{output factor}.
Let $W_0 = U \operatorname{diag}(\sigma) V^\top$ be the singular value decomposition (SVD), with $\sigma$ in decreasing order.
\iftmlr\else
The columns of $U$ are directions of the output space, and the rows of $V^\top$ are directions of the input space.
\fi
The \emph{dominant band} is the first $r$ singular directions.
The \emph{minor band} is the last $r$ singular directions, however small their singular values.
\emph{$B$ frozen on a band} sets $B$ to the $r$ columns of $U$ in that band and trains $A$ from zero.
$B$ frozen on the minor band is MiCA's configuration~\citep{rudiger2026mica}.
\emph{$A$ frozen on a band} sets $A$ to the $r$ rows of $V^\top$ in that band and trains $B$ from zero.
In our notation, $A$ frozen on the dominant band is the factor and band that PiCa freezes~\citep{hwang2026pica}.
$A$ frozen starts from the initialisation of LoRA-RLPO on the dominant band and of LoRA-RLMO on the minor band~\citep{rlvrortho2026}, then keeps $A$ fixed.
The reference is a \emph{free LoRA}: both factors train, from LoRA's usual initialisation~\citep{hu2021lora}.
\iftmlr We adapt the seven linear projections of every block, with $\alpha/r = 1$.
\fi\iftmlr\else
As a control, $A$ or $B$ can be frozen on a random orthonormal basis instead of a band (\Cref{tab:results}).
\fi
\endblock
\block{configsfixed}
In all four configurations the update starts at zero and is scaled by $\alpha/r = 1$, and we adapt the seven linear projections of every block.
\endblock
\block{budgets}
The \emph{training budget} is the number of optimiser steps.
Every configuration trains for 49 steps on OpenBookQA, about six passes over its training questions, and 274 on the format task.
\iftmlr\else On OpenBookQA, 49 steps are 600,000 supervised tokens, about 5.8 passes over the packed training set. \fi
The \emph{parameter count} is the number of trainable parameters.
The matched free LoRA is the free LoRA with about the same parameter count as $B$ frozen (\Cref{app:data} gives the rule per task).
It is the reference for every configuration, except in \Cref{sec:methods}, where it matches the compared method.
$A$ frozen trains $B$, with $r\,d_{\text{out}}$ parameters per module, and $B$ frozen trains $A$, with $r\,d_{\text{in}}$.
Summed over the adapted modules, $A$ frozen trains more parameters than $B$ frozen at the same rank, and than the matched free LoRA, which could favour it.
\iftmlr The up and gate projections of the MLP have a large output, which explains the difference.\fi
\endblock
\block{budgetscounts}
$A$ frozen trains more parameters than $B$ frozen because the up and gate projections of the MLP have a large output.
On Qwen2.5-1.5B at rank 2, the counts are 1,290,240 against 1,017,856.
\endblock
\block{budgetrule}
On the format task, the matched free LoRA is the free LoRA with the smallest count at least as large as that of $B$ frozen.
We choose this reference because comparing MiCA's configuration with a LoRA of about the same size is one of our questions.
\endblock
\block{budgetsend}
\iftmlr
Two checks, one per task, address the extra parameters.
In the OpenBookQA check, we adapt only the query and output projections of attention, whose input and output sizes are equal.
There, both frozen configurations train the same number of parameters, and $A$ frozen still beats $B$ frozen, by $+7.93$ points on OpenBookQA.
On the format task this check was not possible, because $B$ frozen collapsed on these modules: its mean accuracy fell below $45\,\%$, our collapse threshold for this task only.
We instead gave $B$ frozen more parameters than $A$ frozen, and $A$ frozen stayed ahead (\Cref{tab:robustdraft}).
\else
\Cref{app:controls} tests the effect of the extra parameters on modules where the two frozen configurations train exactly the same number.
On OpenBookQA, A $-$ B stays established on these modules.
On the format task, $B$ frozen collapsed on them, so we instead gave $B$ frozen 1.09 to 1.30 times the parameters of $A$ frozen, and $A$ frozen stayed ahead (\Cref{app:controls}).
\fi
\endblock
\block{budgetslow}
\iftmlr
We also observe that $A$ frozen at rank 1 beats $B$ frozen at rank 2 while training fewer parameters, on both tasks and both bands (\Cref{tab:rank1budget}).
\else
$A$ frozen at rank 1 beats $B$ frozen at rank 2 while training fewer parameters (no test; we chose this comparison after the results, and it crosses two experiments).
On OpenBookQA, $A$ frozen at rank 1 trains 645,120 parameters, fewer than the 1,017,856 of $B$ frozen at rank 2.
Its mean accuracy on the dominant band is still higher: 53.50, against 44.65 for $B$ frozen at rank 2.
The minor band and the format task give the same picture (\Cref{tab:rank1budget}).
\fi
\endblock
\block{budgetslowdetail}
\begin{table}[t]
\centering
\small
\begin{tabular}{llrrr}
\toprule
Task & Configuration & Parameters & Dominant band & Minor band \\
\midrule
OpenBookQA & $A$ frozen, rank 1 & 645,120 & 53.50 & 51.80 \\
OpenBookQA & $B$ frozen, rank 2 & 1,017,856 & 44.65 & 42.55 \\
Format task & $A$ frozen, rank 1\posthoc & 376,832 & 79.46 & 75.51 \\
Format task & $B$ frozen, rank 2 & 655,360 & 64.96 & 57.92 \\
\bottomrule
\end{tabular}
\caption{\textbf{$A$ frozen at rank 1 beats $B$ frozen at rank 2 with fewer trainable parameters, on both tasks and both bands.} Mean test accuracy (descriptive). The format task at rank 1 is post hoc.}
\label{tab:rank1budget}
\end{table}
\endblock
\block{rates}
Freezing a factor changes which learning rate works best, so we select the rate separately for every configuration, rank, task and model, with rules fixed in advance; \citet{lee2026learningrate} also find that the best rate differs between LoRA variants, PiSSA needing a lower one than vanilla LoRA.
\endblock
\block{ratesgrid}
In most experiments the candidates are $10^{-4}$, $5 \times 10^{-4}$, $10^{-3}$, $2 \times 10^{-3}$, $5 \times 10^{-3}$ and $10^{-2}$.
\endblock
\block{rates2}
\iftmlr
We try six candidate rates and choose among them with four rules: keep the best validation mean, extend the grid by one rate at an edge, flag a rate still at the edge, and settle near-ties with three more seeds.
Each candidate trains with seeds 0 and 1 and is scored on a validation set disjoint from the test set.
Ties are settled on seeds 30 to 32, and the main results use eight other seeds, 2 to 9 (\Cref{app:rankexp,app:controls} give the exceptions).
Where two rates tied in the eight-task mixture, rerunning the other rate leaves A $-$ B positive in all six cases (\Cref{app:othertied}).
\else
Four rules choose among them (\Cref{app:rules} gives the details):
\fi
\endblock
\block{ratessum}
The four rules keep the best validation mean, extend the grid by one rate at an edge, and decide near-ties with three more seeds.
\endblock
\block{ratesrules}
\iftmlr The four rules are:\fi
\begin{enumerate}
\item Each candidate is trained with seeds 0 and 1 and scored on a validation set disjoint from the test set. The best mean is kept.
\item If the best rate is the smallest or the largest of the grid, we add one more rate beyond it and train it in the same way.
\item A best rate that is still at the new edge is reported as \emph{at the edge of the grid}.
\item Two rates within half an accuracy point are \emph{tied}, and three more seeds decide between them. A tie that remains is an \emph{unresolved tie}.
\end{enumerate}
\endblock
\block{seeds}
Results use eight seeds, 2 to 9, distinct from the seeds 0 and 1 of the rate grid, except the replication (ranks 16 and 32 of the main pairs, run again with new seeds), which uses seeds 12 to 19 (\Cref{sec:rank}).
The seed sets the order of the training examples.
For the free LoRA, it also sets the random initial $A$, drawn module after module from one seeded generator with nothing mixed in per module.
Our code draws this $A$ with the default initialisation of the PEFT library (Kaiming uniform), and the random-$A$ control freezes exactly this initial $A$.
The frozen factors are set by the pretrained weight, not by the seed.
\endblock
\block{ratestability}
\paragraph{The intervals and the choice of learning rate.}
Each interval takes the learning rate chosen on validation as fixed, and it does not cover other tasks or models.
A post hoc check asks how stable that choice is at rank 2 on the main pairs.
It redraws the validation counts of seeds 0 and 1 in 10,000 binomial resamples and applies the rules for the learning rate to each.
This check is not an interval: the resample holds the noise of the 800 validation answers, not the variation of training across seeds.
On OpenBookQA, the rules pick another rate than ours in $12.0\,\%$ to $66.0\,\%$ of the resamples, depending on the configuration.
On the format task, in the resamples where both chosen rates have test scores, A $-$ B on the dominant band is $+13.24$ points, against $+12.13$ at our rates.
These two values come from the per-seed scores rounded to one decimal, so the second one differs slightly from the $+12.11$ of \Cref{tab:results}.
\endblock
\block{testsdetail}
\paragraph{Statuses.}
The appendix gives each result a status, and the body says it in words.
A result is \emph{established} when its interval excludes zero, and \emph{not established} when it does not.
A result is \emph{descriptive} when it has no test.
A result is \emph{post hoc}, marked \dag, when its experiment or comparison was decided after related results were seen.
A \emph{reversal} is an established negative result against the direction the protocol expects.
The seeds vary only the data order and the free LoRA's initial $A$.
When an experiment makes several tests, each interval is widened by a Bonferroni correction ($97.5\,\%$ or $98.75\,\%$ instead of $95\,\%$).
When two families of tests without a joint correction differ, as for PiCa at its published and at selected learning rates (\Cref{sec:methods}), we state the weaker result.
When the end of the interval nearest zero is within 0.5 points of zero, we call the result \emph{suggestive}, or ``barely'' in \Cref{tab:gap} (the end is a lower bound for a positive result and an upper bound for a negative one).
Below the floor of 2 points over the base model, the task is not learned well enough for a difference to mean anything.
\endblock
\block{fs}
The \emph{factor effect} $F$ is the gain from freezing $A$ instead of $B$, averaged over the two bands.
The \emph{band effect} $S$ is the gain from using the minor band instead of the dominant band, averaged over the two factors.
The sign of $S$ only says which band is better, so we compare $F$ with its size $|S|$.
$F$ and $S$ are computed from mean accuracies and carry no test of their own.
\iftmlr On the format task at rank 2, for example, $F = +16.26$ and $S = -2.90$.
\fi\iftmlr\else
Take OpenBookQA at rank 2.
There, $A$ frozen beats $B$ frozen by $+8.60$ points on the dominant band and by $+10.10$ on the minor band (A $-$ B, both positive).
Changing the band, with the frozen factor fixed, moves accuracy much less.
\fi
\endblock
\block{fsvalues}
In the OpenBookQA example at rank 2, the factor effect is $F = +9.35$ points and the band effect is $S = -1.35$ points (descriptive, 95\,\% interval $[-3.16, +0.51]$).
\endblock
\block{fssub}
$S$ is the mean of two changes: $S_A$, the accuracy of $A$ frozen on the minor band minus that of $A$ frozen on the dominant band, and $S_B$, the same difference for $B$ frozen.
Here $S_A = -0.60$ and $S_B = -2.10$ (descriptive).
\endblock
\block{fsalgebra}
In algebra, $F$ exceeds $|S|$ exactly when $F - S$ and $F + S$ are both positive.
$F - S$ is the accuracy of $A$ frozen on the dominant band minus that of $B$ frozen on the minor band.
$F + S$ is the accuracy of $A$ frozen on the minor band minus that of $B$ frozen on the dominant band.
Levels are written as percentages, such as $97.5\,\%$ or $98.75\,\%$, or as fractions (0.975) in the tables.
\endblock
\block{fstest}
\iftmlr
If the factor matters more than the band, $A$ frozen beats $B$ frozen even when $B$ frozen gets the other band.
We therefore test two \emph{cross differences}.
The first is $A$ frozen on the dominant band minus $B$ frozen on the minor band, and it equals $F - S$.
The second is $A$ frozen on the minor band minus $B$ frozen on the dominant band, and it equals $F + S$.
Both are positive exactly when $F > |S|$ (in the example, $+19.16$ and $+13.36$, both established).
The test \emph{$F$ against $|S|$} reads the intervals of the two cross differences and gives one of four verdicts:
\begin{itemize}[nosep]
\item \emph{factor}: both intervals exclude zero in favour of $A$ frozen;
\item \emph{band}: either interval excludes zero in favour of $B$ frozen;
\item \emph{open}: every other case;
\item \emph{no test}: the experiment has no such test.
\end{itemize}
In the text, ``the factor matters more than the band'' means the verdict \emph{factor}.
\else
The test \emph{$F$ against $|S|$} compares $A$ frozen on each band with $B$ frozen on the other band.
``$F$ exceeds $|S|$'' is established when both differences are established positive.
``$|S|$ exceeds $F$'' is established when either difference is established negative.
The verdict is \emph{open} when neither is established, and \emph{no test} when the experiment has no such test.
We formed the hypothesis that $F$ exceeds $|S|$ on OpenBookQA with Qwen2.5-1.5B at rank 2, so its confirmation comes from the format task, the 7B models and the other ranks.
\fi
\iftmlr\else\Cref{app:stats} gives the levels of the intervals and says when this test is weak.\fi
\endblock
\block{fsweak}
\paragraph{When the test of $F$ against $|S|$ is weak.}
The test is weak when $B$ frozen stays near the base model's accuracy.
Then $|S|$ is bounded by how little $B$ frozen moves, and the test passes whatever the band does.
The format task (base model $0\,\%$) is the case where this does not apply.
\endblock
\block{prespec}
\iftmlr
We wrote every test of the main design before its experiment ran; \Cref{tab:posthoc} lists the ones we added after seeing related results.
A test runs only if the free LoRA gains at least 2 points over the base model, which shows that the task is learned.
The intervals take the chosen learning rate as fixed (\Cref{app:stats} checks how stable that choice is).
Each test is read at its own pre-specified level.
A post hoc Holm correction over the 258 primary tests keeps 117 of the 151 established results, among them 15 of the 16 tests of A $-$ B on the main pairs at ranks 1 to 8 (four ranks, two pairs, two bands).
Rank 1 of the format task is among the 258, although it is post hoc; GSM8K, the 8B model and the random bases on eight seeds are not.
The comparison with the band survives at ranks 1, 2 and 4, not at rank 8 (\Cref{app:holm}).
An established result that survives this correction is \emph{Holm-robust}.
The text marks these results, and \Cref{tab:status} gives the status of each result cited.

\paragraph{What is exploratory.}
The hypothesis that $F$ exceeds $|S|$ came from OpenBookQA with Qwen2.5-1.5B at rank 2, so that setting is exploratory.
Its confirmation comes from the format task, the two 7B models and the other ranks, whose tests were written before their runs, with post hoc exceptions: rank 1 of the format task, the replication at ranks 16 and 32, two 7B rows of \Cref{fig:forest}, the OpenBookQA points of \Cref{fig:budget} after 49 steps, and the gradient analysis of \Cref{sec:discussion}.
The tables mark these added tests ``after the results''.
\else
We wrote every test of the main design before its experiment ran; \Cref{tab:posthoc} lists the ones we added after seeing related results.
A test runs only if the free LoRA gains at least 2 points over the base model, which shows that the task is learned.
The intervals take the chosen learning rate as fixed (\Cref{app:stats} checks how stable that choice is).
Each test is read at its own pre-specified level.
A post hoc Holm correction over the 258 primary tests keeps 117 of the 151 established results, among them A $-$ B on both main pairs at every rank from 1 to 8, in 15 of its 16 tests; the comparison with the band survives at ranks 1, 2 and 4, not at rank 8 (\Cref{app:holm}).
An established result that survives this correction is \emph{Holm-robust}.
The text marks these results, and \Cref{tab:status} gives the status of each result cited.
A result whose experiment or comparison was decided after related results were seen is \emph{post hoc}, marked \dag.
\Cref{app:deviations} lists every departure from a protocol, and \Cref{app:tests} every pre-specified test.
\fi
\endblock

\block{prespecfixed}
\paragraph{What a protocol fixes.}
Each experiment has a protocol written before its runs.
A protocol fixes the data, the configurations, the rate rules, the tests, their levels and the conclusion allowed by each outcome.
\endblock
\block{prespeckinds}
\paragraph{Reading the mark \dag.}
The mark \dag\ has three kinds.
In the first kind, the experiment was chosen after related results were seen, but its tests were written before its runs.
\dag\ is of this kind on an established or not established result.
In the second kind, the test itself was fixed after results that it compares had been read.
The five cases are the format task at rank 1, every learning rate of $B$ frozen, factor against band on three pairs, forgetting at rank 16 and the least-squares predictor (\Cref{tab:deviations}).
In the third kind, the analysis was chosen after seeing the results, and it has no test.
\dag\ is of this kind on a descriptive number, and no conclusion rests on it.
\endblock
\block{tasks}
The main pairs run at ranks 1, 2, 4, 8, 16, 32 and 64.
They are OpenBookQA~\citep{mihaylov2018openbookqa} with
Qwen2.5-1.5B~\citep{qwen2024qwen25} and a \emph{format task} with Llama-3.2-1B~\citep{grattafiori2024llama3}.
In the \emph{main protocol} on OpenBookQA (500 test questions, 49 training steps), the answer is the option with the highest mean log-likelihood over its tokens.
\endblock
\block{scoring}
On OpenBookQA, the score of an option is its summed log-likelihood divided by the number of its tokens.
Adapters are trained and scored in this format (\Cref{app:letter} gives the results with a letter prompt).
\endblock
\block{tasks2}
The format task takes its sentences from CoNLL-2003~\citep{tjongkimsang2003conll}, and an output is either correct or not (614 test items).
\iftmlr
Other pairs test whether the result carries over, with ARC-Easy and ARC-Challenge~\citep{clark2018arc}, the instruction-tuned Qwen2.5-1.5B-Instruct, the more recent Qwen3-1.7B-Base (\Cref{app:rankexp}), two 7B models, Qwen2.5-7B and Mistral-7B~\citep{jiang2023mistral}, Llama-3.1-8B, and GSM8K~\citep{cobbe2021gsm8k} with Llama-3.2-1B.
An \emph{eight-task mixture} trains one adapter on eight commonsense tasks and scores its mean accuracy.
\fi
Most controls run at rank 2.
\endblock
\block{tasks2why}
The two pairs cover choosing among given answers and producing a new structured output, and fine-tuning gains much on both.
\endblock
\block{tasks2detail}
\iftmlr\else
Other pairs test whether the result carries over: ARC-Easy and ARC-Challenge~\citep{clark2018arc} (ARC, the AI2 Reasoning Challenge, grade-school science questions), an instruction-tuned model, a
more recent base model (Qwen3-1.7B-Base), Qwen2.5-7B and Mistral-7B~\citep{jiang2023mistral}.
An \emph{eight-task mixture} trains one adapter on eight commonsense tasks and scores its mean accuracy (\Cref{app:data}).
\fi
\endblock
\block{letterprompt}
\paragraph{OpenBookQA: the controls and the letter prompt.}
Four controls (the $\alpha/r$ sweep, SGD, the $4\times$ convergence and adapter merging) ran on OpenBookQA only.
With a letter prompt, a transfer the adapters were not trained for, the base model answers $76.6\,\%$.
There, $A$ frozen gets fewer items right than $B$ frozen, by $-621$ and $-1{,}496$ of 4,000 on the two bands (descriptive, post hoc).
When the adapters are trained on the letter and scored by generation, A $-$ B is $+1.70$ and $+2.48$ (descriptive, 95\,\% intervals
$[-0.23, +3.70]$ and $[+0.07, +4.98]$).
\Cref{tab:oqaprot} gives the design trained and scored with the letter prompt.
\endblock
\block{posthoctable}
\begin{table}[h]
\centering
\small
\caption{\textbf{Analyses decided after seeing the results, and the section that reports each one.}
An experiment in this list was chosen after related results, and its tests were written before its runs.
Five entries are of a second kind, with tests fixed after results that they compare had been read: the format task at rank 1, every learning rate of $B$ frozen, factor against band on three pairs, forgetting at rank 16 and the predictor (\Cref{tab:deviations}).
A comparison gives observations only.}
\label{tab:posthoc}
\begin{tabular}{@{}>{\raggedright\arraybackslash}p{0.42\linewidth}l>{\raggedright\arraybackslash}p{0.42\linewidth}l@{}}
\toprule
Analysis & Sec. & Analysis & Sec. \\
\midrule
$A$ frozen at rank 1 against $B$ frozen at rank 2 & \ref{sec:budgets} & Eight-task mixture at ranks 16 and 32 & \ref{sec:rank} \\
Format task at rank 1 & \ref{sec:rank} & Flipped MiCA & \ref{sec:methods} \\
Format task and ARC-Challenge, Qwen2.5-7B & \ref{sec:rank} & Matched free LoRA vs.\ dominant-band $B$ frozen & \ref{sec:methods} \\
Letter prompt & \ref{sec:rank} & Last step at four times the training budget & \ref{sec:narrow}, \ref{app:robust} \\
Replication at ranks 16 and 32, its tests & \ref{sec:rank} & Predictor from \citet{zhu2024asymmetry} & \ref{sec:robustness} \\
OpenBookQA, Qwen2.5-7B, rank 16 & \ref{sec:rank}, \ref{sec:methods} & Forgetting at matched accuracy and at rank 16 & \ref{app:robust} \\
Mistral-7B at rank 64 & \ref{sec:rank} & Transfer from OpenBookQA to ARC-Easy & \ref{app:transfer} \\
Every learning rate of $B$ frozen & \ref{sec:robustness} & Gradient subspaces at step 0 and at the last step & \ref{sec:discussion} \\
Factor against band on three pairs & \ref{app:tests} & GSM8K with Llama-3.1-8B & \ref{sec:rank} \\
Output overlap at ranks 8 and 16 & \ref{app:overlap} & Scaling $\alpha$ and unfreezing a factor, at rank 16 & \ref{app:rankexp} \\
\bottomrule
\end{tabular}
\end{table}
\endblock

\fi%

\subsection{How we measure factor against band}
\label{sec:tests}

We measure the factor with one difference per band, and we compare factor and band with two cross
differences.
Let $a_{\text{dom}}$ and $a_{\text{min}}$ be the accuracies of $A$ frozen on the dominant and the
minor band, and $b_{\text{dom}}$ and $b_{\text{min}}$ those of $B$ frozen.
\emph{A $-$ B} is the accuracy of $A$ frozen minus the accuracy of $B$ frozen on the same band.
The \emph{factor effect} $F$ is the gain from freezing $A$ instead of $B$, averaged over the two
bands.
The \emph{band effect} $S$ is the gain from the minor band instead of the dominant band, averaged
over the two factors:
\[
  F = \tfrac{1}{2}\bigl[(a_{\text{dom}} - b_{\text{dom}}) + (a_{\text{min}} - b_{\text{min}})\bigr],
  \qquad
  S = \tfrac{1}{2}\bigl[(a_{\text{min}} - a_{\text{dom}}) + (b_{\text{min}} - b_{\text{dom}})\bigr].
\]
A positive $S$ means that the minor band is better, and we compare $F$ with the size $|S|$.
$F$ and $S$ come from mean accuracies and carry no test of their own, so they are descriptive.

The test uses the two \emph{cross differences}, which change the factor and the band together:
\[
  a_{\text{dom}} - b_{\text{min}} = F - S,
  \qquad
  a_{\text{min}} - b_{\text{dom}} = F + S .
\]
Both cross differences are positive exactly when $F > |S|$.
The test of $F$ against $|S|$ reads their two intervals and returns one of the four verdicts of
\Cref{tab:statusterms}.
``The factor matters more'' means the verdict \emph{factor}.
\iftmlr\else

Take OpenBookQA at rank 2 as an example.
A $-$ B is $+8.60$ points on the dominant band and $+10.10$ on the minor band.
Changing the band with the frozen factor fixed moves accuracy much less.
\Cref{app:stats} continues the example and says when the test of $F$ against $|S|$ is weak.
\fi%

\subsection{Statistical protocol}
\label{sec:prespec}

\begin{table}[t]
\centering
\small
\caption{The status words and the four verdicts, defined once for the whole paper.}
\label{tab:statusterms}
\begin{tabular}{@{}l>{\raggedright\arraybackslash}p{0.7\linewidth}@{}}
\toprule
established & The interval of the test excludes zero at the test's own pre-specified level. \\
Holm-robust & Still established after the post hoc Holm correction over the 258 primary tests. \\
not detected (at most $x$ points) & The interval includes zero, and $x$ is the largest effect in the expected direction that it allows. A real effect of a few points is not excluded. \\
post hoc & Decided, or its test fixed, after we had seen related results (\Cref{tab:posthoc}). \\
descriptive & Reported without a test; no conclusion rests on it. \\
outside the Holm family & Not among the 258 primary tests, and read at its own level. \\
\midrule
\multicolumn{2}{@{}l}{Verdicts of the test of $F$ against $|S|$, read from the intervals of the two cross differences} \\
factor & Both intervals exclude zero in favour of $A$ frozen. \\
band & Either interval excludes zero in favour of $B$ frozen. \\
open & Every other case. \\
no test & The experiment has no such test. \\
\bottomrule
\end{tabular}
\end{table}

The tests of the main design were written before their experiments ran.
Two parts of the main design are post hoc: rank 1 of the format task, and the replication at ranks 16 and 32, decided
after ranks 4 and 8 had been read.
Every interval is a paired bootstrap interval over seeds and test items (\Cref{app:stats}).
It takes the selected learning rate as fixed.
\Cref{tab:statusterms} defines the status words that we give to each result.
A test runs only if it passes a gate: the free LoRA gains at least 2 points over the base model, so the task is learned.
\iftmlr\else
\Cref{app:deviations} lists every departure from a protocol, and \Cref{app:tests} every
pre-specified test.
\fi%

One setting is exploratory: OpenBookQA with Qwen2.5-1.5B at rank 2, where we formed the hypothesis
$F > |S|$.
The format task, the two 7B models and the other ranks serve as confirmation.
We ran many tests, so we add a post hoc Holm correction over the 258 primary tests.
It keeps 117 of the 151 tests established at their own level, and \Cref{app:holm} lists the 34
that do not survive.
\Cref{tab:status} gives the main statuses.

\subsection{Which tasks and models}
\label{sec:tasks}

Two main pairs carry the design at ranks 1, 2, 4, 8, 16, 32 and 64.
The first pair is OpenBookQA \citep{mihaylov2018openbookqa} with Qwen2.5-1.5B
\citep{qwen2024qwen25}.
OpenBookQA is a multiple-choice task with 500 test questions.
The answer is the option with the highest mean log-likelihood over its tokens, and the base
model scores 42.20\% under that protocol.
The second pair is the format task with Llama-3.2-1B \citep{grattafiori2024llama3}.
The \emph{format task} rewrites the named entities of a sentence into a fixed one-line JSON
structure.
Its sentences come from CoNLL-2003 \citep{tjongkimsang2003conll}, and it has 614 test items (\Cref{app:data}).
An output is correct when its format is (entities unchecked); base models never produce one.
\iftmlr\else
The two pairs cover two kinds of task: choosing among given answers and producing a new structured
output.
Fine-tuning gains much on both.
\fi%

Other tasks and models test how far the finding extends.
The tasks are ARC-Easy and ARC-Challenge \citep{clark2018arc}, GSM8K \citep{cobbe2021gsm8k} and an
eight-task mixture.
The mixture trains one adapter on eight commonsense tasks and reports their mean accuracy.
The other models are an instruction-tuned Qwen2.5-1.5B, Qwen2.5-7B, Mistral-7B
\citep{jiang2023mistral}, Llama-3.1-8B and Qwen3-1.7B-Base.
\iftmlr\else
Qwen3-1.7B-Base, a more recent base model, runs on OpenBookQA at rank 2 only.
ARC, the AI2 Reasoning Challenge, holds grade-school science questions.
\Cref{app:data} lists the eight tasks of the mixture.
\fi%

We focus on low rank, where a frozen factor constrains the update most.
Rank 8 is the default of the PEFT library \citep{mangrulkar2022peft}.
The metric is test accuracy in percent everywhere, and differences are in accuracy points.
Main results use eight seeds, 2 to 9, and \Cref{app:rankexp,app:controls,app:x16} list the exceptions.
\iftmlr\else
One exception is the second run of ranks 16 and 32 on the main pairs, which uses seeds 12 to 19
(\Cref{sec:rank}).
The seed sets the order of the training examples.
For the free LoRA, it also sets the random initial $A$.
We draw that $A$ with the default initialisation of the PEFT library (Kaiming uniform), module
after module from one seeded generator, with nothing mixed in per module.
The random-$A$ control of \Cref{sec:robustness} freezes exactly that initial $A$.
The pretrained weight sets the frozen factors, and the seed does not.
\fi%

\secmark{results}
\begin{figure}[t]
\centering
\paperfig[width=0.80\linewidth]{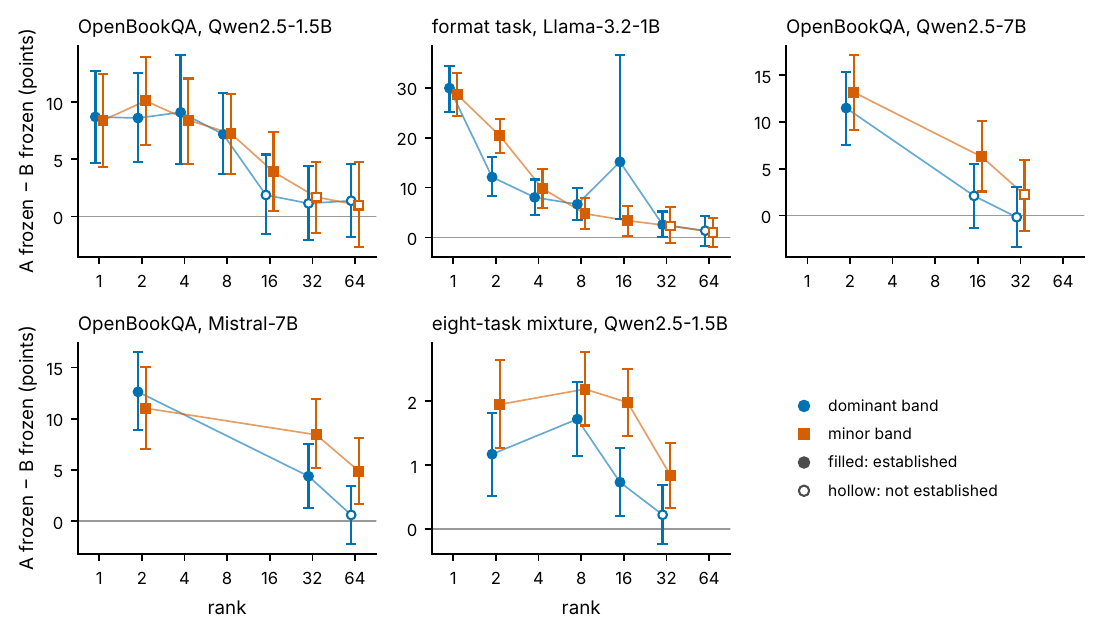}
\caption{$A$ frozen beats $B$ frozen at ranks 1 to 8 on both main pairs.
Above rank 8, the advantage is established mostly on the minor band.
Each panel plots A $-$ B in accuracy points against the rank for one task and model, with its own
vertical scale.
Bars are the intervals of the tests, and thin lines join tested ranks only.
Blue circles are the dominant band and orange squares the minor band.
A filled mark is established and a hollow mark is not (\Cref{tab:statusterms}).
On the main pairs, ranks 16 and 32 show the replication (\Cref{app:rankexp}).
Five sets of points are post hoc: rank 1 of the format task, that replication, Qwen2.5-7B at
rank 16, Mistral-7B at rank 64 and the mixture above rank 8.
One diverged run inflates the format task's dominant band at rank 16.}
\label{fig:rank}
\end{figure}

\section{At ranks 1 to 8, the frozen factor matters more than the band}
\label{sec:rank2}\label{sec:rank}

Which choice, the frozen factor or the band, explains the gap between the four frozen
configurations?
And how far does the answer hold across ranks, models, tasks and training lengths?
\Cref{sec:advholds} gives the settings where the advantage holds, and \Cref{sec:narrow} those where
it narrows.

\subsection{Where the advantage holds}
\label{sec:advholds}

\paragraph{One case in full: at rank 2 on the format task, the factor moves accuracy and the band
barely does.}
We take the format task with Llama-3.2-1B at rank 2 (\Cref{fig:design}, middle panel).
Every configuration trains for 274 steps at its own selected learning rate, and $A$ frozen trains
the most parameters (\Cref{sec:robustness}).
The matched free LoRA reaches 80.07\%, above all four frozen configurations.
With $A$ frozen, accuracy is 77.08\% on the dominant band and 78.32\% on the minor band.
With $B$ frozen, it is 64.96\% and 57.92\%.
A $-$ B is $+12.11$ points on the dominant band and $+20.40$ on the minor band, both
Holm-robust.

The test of \Cref{sec:tests} agrees.
The factor effect is $F = +16.26$ and the band effect is $S = -2.90$.
The two cross differences are $+19.16$ and $+13.36$ points.
Both are established, so the verdict is \emph{factor}.
PiCa's cell and MiCA's cell are 19 points apart in this case, and most of that gap comes with the
frozen factor.

\paragraph{The same pattern holds at ranks 1, 2, 4 and 8 on the two main pairs.}
We ask whether rank 2 is special, and repeat the experiment at ranks 1, 4 and 8 on both main pairs
(\Cref{fig:rank}).
$A$ frozen beats $B$ frozen on both bands at each of these ranks, on both pairs.
On the format task the gap is largest at the smallest ranks: $+29.93$ points at rank 1, which is
post hoc, and still $+6.68$ and $+4.78$ points at rank 8.
On OpenBookQA the gap is about $+7$ points on both bands at rank 8, and looks flat from rank 1 on
(not tested).

All 16 differences are established, and the Holm correction keeps 15 of them.
The exception is the minor band of the format task at rank 8.
The verdict \emph{factor} is Holm-robust at ranks 1, 2 and 4 on both pairs, and at rank 8 it is
established and not Holm-robust.
The format task is our primary evidence, because the hypothesis came from the OpenBookQA pair at
rank 2, and that pair is fragile (\Cref{sec:prespec,sec:narrow}).
On the two main pairs, the factor therefore matters more than the band at ranks 1 to 8, Holm-robustly at ranks 1, 2 and 4.

\begin{figure}[t]
\centering
\paperfig[width=0.79\linewidth]{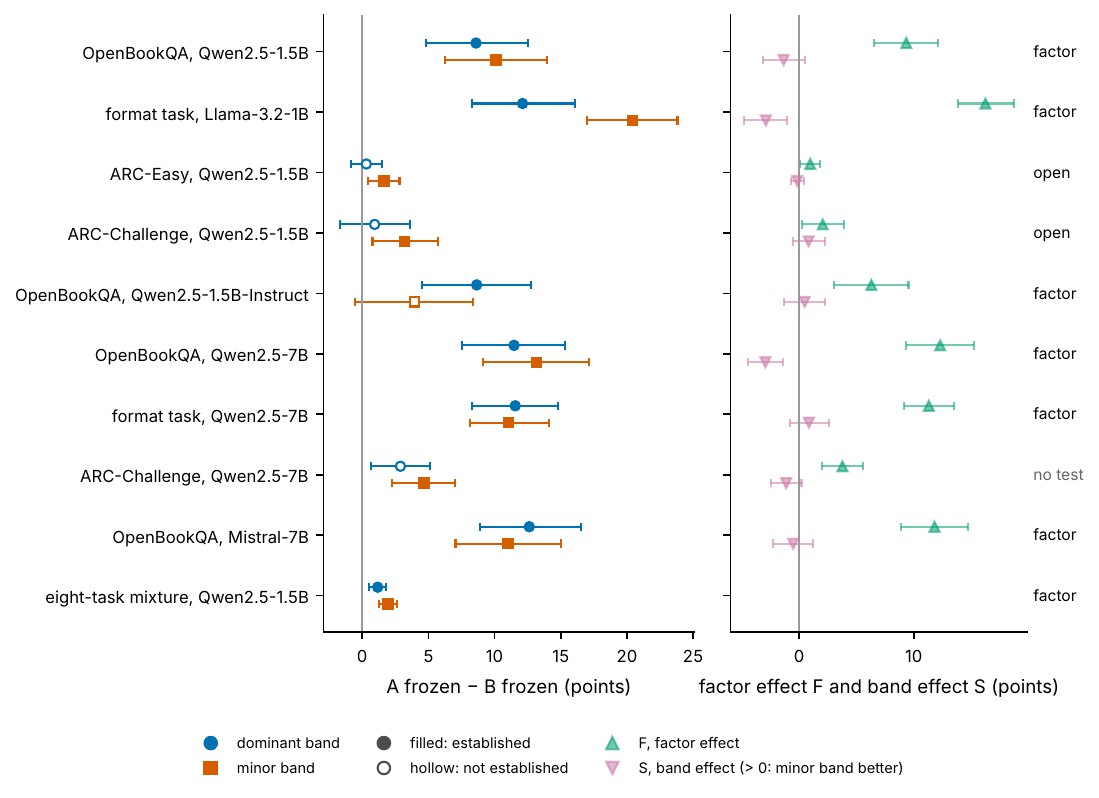}
\caption{At rank 2, the factor matters more than the band in seven of the nine settings shown (ten of the twelve of \Cref{tab:results}).
A tenth setting has no such test.
Left panel: A $-$ B on each band, in accuracy points, with the interval of its test.
Blue circles are the dominant band and orange squares the minor band.
A filled mark is established and a hollow mark is not.
One hollow mark has no test, the dominant band of ARC-Challenge with Qwen2.5-7B.
Right panel: the factor effect $F$ as green triangles pointing up, the band effect $S$ as pink
triangles pointing down, with 95\% descriptive intervals.
$S > 0$ means the minor band is better.
The word at the right edge is the verdict of \Cref{tab:statusterms}.
The mixture row shows its verdict only, because no intervals were computed for its $F$ and $S$.
The two Qwen2.5-7B rows on the format task and on ARC-Challenge are post hoc (\Cref{app:contrasts}).}
\label{fig:forest}
\end{figure}

\begin{table}[t]
\centering
\small
\caption{Beyond the two main pairs, A $-$ B is established on both bands with the 7B models at rank 2 and
on GSM8K; with Llama-3.1-8B on the two main tasks the verdict is open.
ARC, the instruction-tuned model and the mixture are in \Cref{tab:results}.
A $-$ B in accuracy points, bold when established; statuses as in \Cref{tab:statusterms},
``outside'' is outside the Holm family.}
\label{tab:beyond}
\begin{tabular}{@{}lrrrll@{}}
\toprule
Setting & rank & dominant & minor & verdict & status \\
\midrule
OpenBookQA, Qwen2.5-7B & 2 & \textbf{+11.47} & \textbf{+13.15} & factor & Holm-robust \\
OpenBookQA, Mistral-7B & 2 & \textbf{+12.62} & \textbf{+11.02} & factor & Holm-robust \\
format task, Qwen2.5-7B & 2 & \textbf{+11.56} & \textbf{+11.07} & factor & post hoc \\
GSM8K, Llama-3.2-1B & 2 & \textbf{+5.62} & \textbf{+6.57} & factor & outside \\
 & 8 & \textbf{+3.75} & \textbf{+5.24} & open & outside \\
GSM8K, Llama-3.1-8B & 8 & \textbf{+4.63} & \textbf{+6.89} & factor & post hoc, outside \\
 & 16 & \textbf{+4.62} & \textbf{+7.53} & factor & post hoc, outside \\
OpenBookQA, Llama-3.1-8B & 8 & +2.95 & \textbf{+7.02} & open & outside \\
format task, Llama-3.1-8B & 8 & \textbf{+4.38} & +1.69 & open & outside \\
\bottomrule
\end{tabular}
\end{table}

\paragraph{The advantage repeats on two 7B models and on GSM8K.}
Next we ask whether the result holds on other models and tasks (\Cref{tab:beyond} and \Cref{fig:forest}).
At rank 2, the verdict is \emph{factor} in the three 7B settings tested.
The 7B runs on OpenBookQA keep its 49-step budget, and we did not train them longer.
We also swapped the models of the two main pairs.
The advantage stays on both bands, with 3 of 4 differences Holm-robust.
GSM8K is our only reasoning task, trained for two epochs.
With Llama-3.2-1B the verdict is \emph{factor} at rank 2 and \emph{open} at rank 8.
With Llama-3.1-8B, chosen post hoc, it is \emph{factor} at ranks 8 and 16
(\Cref{tab:gsm,tab:gsm8b}).
\iftmlr\else
With Qwen3-1.7B-Base, a more recent base model, A $-$ B is established on both bands of
OpenBookQA, and the verdict is \emph{factor}.
\Cref{tab:results} gives A $-$ B for every setting and rank, and \Cref{app:contrasts} gives every
contrast with its interval.
\fi%

\begin{figure}[t]
\centering
\paperfig[width=0.77\linewidth]{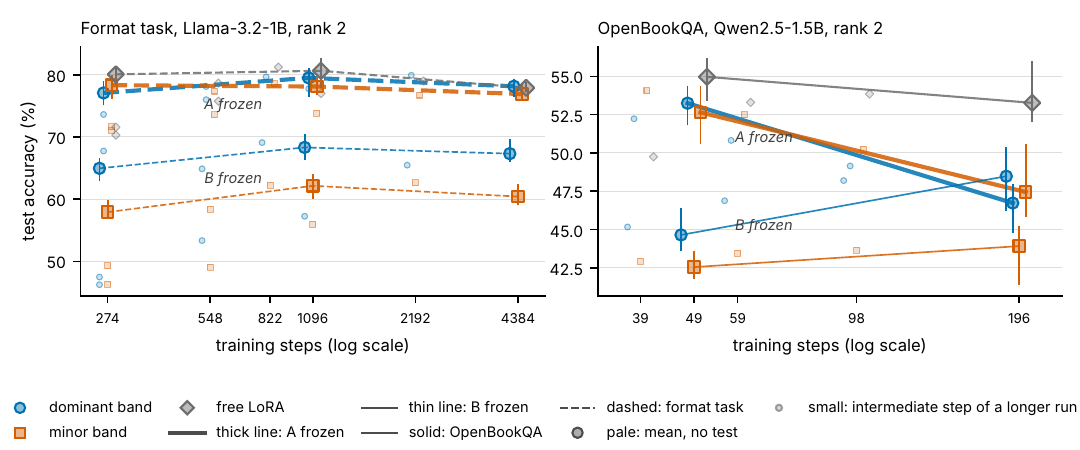}
\caption{On the format task the gap persists with longer training, and on OpenBookQA it closes as
$A$ frozen declines.
Both panels show test accuracy (\%, mean over seeds) against the number of training steps on a
log scale, at rank 2.
The left panel is the format task with Llama-3.2-1B and the right panel is OpenBookQA with
Qwen2.5-1.5B.
Thick lines are $A$ frozen and thin lines are $B$ frozen, one colour per band, and the remaining
curve is the free LoRA.
Lines join the last steps of separate runs, each with its own selected learning rate.
Bars span the lowest and the highest seed and are not confidence intervals.
Small marks are intermediate steps of a longer run, which are not equivalent to shorter runs.
Pale marks are means with no test.
The OpenBookQA points after 49 steps are post hoc (\Cref{app:budgetfig}).}
\label{fig:budget}
\end{figure}

\paragraph{On the format task, the gap persists with sixteen times the budget.}
The main pairs train briefly, so we ask whether $B$ frozen only needs more steps to catch up
(\Cref{fig:budget}).
On the format task at rank 2, sixteen times the budget does not close the gap, though on the minor band it shrinks after step 274 of that schedule (\Cref{app:x16}).
A $-$ B is then $+10.83$ and $+16.48$ points, both established and outside the Holm family, against $+12.11$ and $+20.40$ at the main budget.
$B$ frozen reaches a training loss of 0.001 and $A$ frozen 0.000 (\Cref{app:x16}), so slow
training alone does not explain the gap.
That reading rests on six seeds, with rates of $B$ frozen at the grid edge.

\subsection{Where the advantage narrows}
\label{sec:narrow}

\paragraph{Above rank 8, the advantage is established mostly on the minor band.}
We ran ranks 16, 32 and 64 on the main pairs, with a post hoc replication of ranks 16 and 32 on new
seeds and a finer rate grid (\Cref{fig:rank}).
At rank 16 the advantage is established on the minor band, on both main pairs, with Qwen2.5-7B
and in the eight-task mixture.
With Mistral-7B it lasts up to rank 64 on that band.
Four minor-band results above rank 8 are Holm-robust: Qwen2.5-7B at rank 16, Mistral-7B at
rank 32 and the mixture at ranks 16 and 32 (\Cref{tab:status}).
On the dominant band of OpenBookQA, the difference at rank 16 is not established.
At rank 16 on the format task, the verdict stays \emph{factor} in both runs, but one diverged run
of $B$ frozen inflates the replication's dominant-band value and its verdict (\Cref{app:collapse}).
With four times the budget at rank 16, the format task keeps the advantage on both bands and
OpenBookQA does not (\Cref{app:cvr16}).
\iftmlr\else
On the format task, A $-$ B is then $+4.13$ points on the dominant band and $+4.11$ on the minor
band.
\Cref{app:rankexp} gives every setting and rank, among them the LoRA scale at rank 16 and
Mistral-7B up to rank 64.
\Cref{app:notest} gives the bounds where A $-$ B is not established.
\fi%
We tested no interaction between rank and factor.
Above rank 8, we therefore state the advantage band by band, not as a rule.

\paragraph{On OpenBookQA above rank 8, the observed band effect exceeds the factor effect.}
On OpenBookQA with Qwen2.5-1.5B, $B$ frozen does better on the dominant band at every rank from 16
on.
The observed $|S|$ exceeds $F$ in all five comparisons we have at ranks 16, 32 and 64
(\Cref{tab:reversal}).
\iftmlr\else
$S$ ranges from $-3.25$ to $-5.60$ points and $F$ from $+1.19$ to $+2.89$, with no test.
\fi%
The test confirms the verdict \emph{band} in one of the five, the replication at rank 32.
In that case the second cross difference, $-4.17$ points, is established.
That case is post hoc, and in the other four the test does not confirm the observation.
\iftmlr\else
In the confirmed case, A $-$ B is established on neither band.
\citet{lee2026learningrate} see the same direction in methods that train both factors.
In their results, PiSSA, on the dominant band, overtakes vanilla LoRA as the rank grows.
MiLoRA, on the minor band, loses its lead, within 1 to 2 points.
\fi%
On this pair, the statement that the factor matters more than the band stops at rank 8.
\iftmlr\else
\begin{table}[!htb]
\centering
\small
\caption{\textbf{On OpenBookQA with Qwen2.5-1.5B above rank 8, the observed band effect $|S|$ is larger than the factor effect $F$ in all five comparisons; one of the five tests finds it.}
$F$ and $S$ are observed values in accuracy points (\Cref{sec:tests}), and only the verdict is a test.
$S$ is negative: the dominant band is the better one.
Verdict and note: as in \Cref{tab:results}.}
\label{tab:reversal}
\begin{tabular}{@{}rlrrll@{}}
\toprule
rank & run & $F$ & $S$ & verdict & note \\
\midrule
16 & first exp. & $+2.78$ & $-3.25$ & open & \\
16 & replication & $+2.89$ & $-3.44$ & open & after the results \\
32 & first exp. & $+1.91$ & $-5.04$ & open & \\
32 & replication & $+1.43$ & $-5.60$ & band & after the results \\
64 & & $+1.19$ & $-4.26$ & open & \\
\bottomrule
\end{tabular}
\end{table}

\fi%

\paragraph{With the largest model, the advantage is established on one band of each main task.}
Llama-3.1-8B is the largest model we ran, at rank 8 on both main tasks.
A $-$ B is established on the minor band of OpenBookQA and on the dominant band of the format
task (\Cref{tab:beyond}); on the other bands we detect no difference (at most 6.45 and 4.09 points).
At rank 16 on OpenBookQA, we detect no difference on either band.
The verdict is \emph{open} in all three settings, which are outside the Holm family
(\Cref{app:llama8b}).
On GSM8K the same model gives the verdict \emph{factor} (\Cref{tab:beyond}).

\paragraph{On ARC and in the eight-task mixture, the rank-2 advantage is small.}
At rank 2, the verdict is \emph{factor} in ten of the twelve settings tested (\Cref{tab:results}).
On ARC-Easy and ARC-Challenge with Qwen2.5-1.5B, A $-$ B is established on the minor band only, by a
small margin.
The verdict is \emph{open} on both ARC tasks.
With the instruction-tuned Qwen2.5-1.5B, A $-$ B is established on the dominant band only.
The verdict \emph{factor} still holds for that model, established and not Holm-robust.
In the eight-task mixture the advantage is small, $+1.17$ and $+1.95$ points on the two bands.
Inside the mixture, $A$ frozen loses to $B$ frozen on ARC-Easy, dominant band.

\paragraph{With longer training on OpenBookQA, the gap closes as $A$ frozen declines.}
On OpenBookQA at rank 2, with four times the budget, $A$ frozen falls from 53.25\% to 46.73\% on
the dominant band, and $B$ frozen rises from 44.65\% to 48.48\%.
At that last step, A $-$ B is $-1.75$ points on the dominant band, and its descriptive interval
includes zero on both bands.
At the checkpoint of each run with the best validation accuracy, the gap holds on the minor band
only, at $+8.45$ points.
The gap closes mostly because $A$ frozen declines, while the free LoRA goes from 54.97\% to
53.27\%.
We read the decline as overfitting and did not test that reading.
The OpenBookQA points after 49 steps are post hoc.

\paragraph{OpenBookQA with Qwen2.5-1.5B at rank 2 is the most fragile pair.}
At 49 steps A $-$ B is established on both bands, and $B$ frozen stays near the base model.
The base model scores 42.20\% and $B$ frozen reaches 44.65\% on the dominant band.
A configuration that barely learns says little about the band, so the test of $F$ against $|S|$ is
weak there (\Cref{tab:oqaprot,app:stats}).
The scoring prompt matters as well.
A letter prompt lists the options by letter and scores the answer letter; the base model reaches
76.60\% with it.
When we train and score with the letter prompt, A $-$ B is established on both bands at 49 steps (the dominant band rests on one run of
$B$ frozen, \Cref{app:letter}) and on neither at 893 steps, and we do not find $F$ above $|S|$.
When we score adapters trained under the main protocol with the letter prompt, a format they were
not trained for, the sign reverses.
A $-$ B is then $-15.5$ points on the dominant band and $-37.4$ on the minor band.
That reversal is descriptive and post hoc, and we have no explanation for it (\Cref{app:letter}).
\iftmlr\else
We also trained the same questions as an answer-only generation task, for 1,187 steps.
A $-$ B is then small, $+1.70$ points on the dominant band and $+2.48$ on the minor band, with no
test (\Cref{app:oc}).
\fi%
On this pair, the gap depends on the protocol, and its sign reverses with the format.
\iftmlr\else
\def\wantedblock{oqatable}\immediate\write\secposfile{block sec4_results oqatable}\block{claim}
This section asks three questions.
Does freezing $A$ instead of $B$ raise accuracy on the main pairs, and at which ranks?
Does the frozen factor matter more than the band?
And where does the answer stop holding: at higher ranks, with other models and tasks, with longer training?
\endblock
\block{others}
\paragraph{In the other settings, the advantage is smaller or holds on one band.}
Swapping the models of the two main pairs (the format task on Qwen2.5-1.5B, OpenBookQA on Llama-3.2-1B) keeps the advantage on both bands.
Its size follows the task more than the model, and three of the four differences are Holm-robust.
On ARC with Qwen2.5-1.5B, $A$ frozen wins on the minor band only, by a small margin.
With the instruction-tuned Qwen2.5-1.5B, it wins on the dominant band only, and the factor still matters more than the band.
In the eight-task mixture the gain is small, and on ARC-Easy inside that mixture $A$ frozen loses to $B$ frozen on the dominant band.
\endblock
\block{aboveeightshort}
\paragraph{Above rank 8, A $-$ B holds mostly on the minor band.}
We ran ranks 16 and 32 of the main pairs twice: a \emph{first experiment}, and a \emph{replication} with new seeds and a finer grid of learning rates.
The gain survives mostly on the minor band, the band where MiCA freezes $B$.
At rank 16, it is established there on both main pairs, with Qwen2.5-7B and in the mixture; with Mistral-7B, it lasts up to rank 64.
Four of these results are Holm-robust: Qwen2.5-7B at rank 16, Mistral-7B at rank 32, and the mixture at ranks 16 and 32.
On the dominant band at rank 16, the gain is not established on OpenBookQA, and on the format task one diverged run drives it.
With four times the training budget at rank 16, the format task keeps the gain on both bands ($+4.13$ and $+4.11$ points) and OpenBookQA does not (\Cref{app:rankexp,app:cvr16}).
The test of an interaction between rank and factor is pending.
\endblock
\block{tableintro}
\iftmlr
\paragraph{At ranks 1 to 8, freezing $A$ beats freezing $B$ on both main pairs, and the factor matters more than the band.}
Freezing $A$ wins at ranks 1, 2, 4 and 8, on both bands of both main pairs (\Cref{fig:design,fig:rank}).
At rank 8, the default of the PEFT library, it leads by about $+7$ points on both bands of OpenBookQA, and by $+6.68$ and $+4.78$ on the format task.
The format task shows its largest gaps at the smallest ranks, up to $+29.93$ points at rank 1, a point we added post hoc.
On OpenBookQA, the gap stays about the same from rank 1 to 8, a trend we did not test.
The Holm correction keeps 15 of these 16 differences; the exception is the $+4.78$.
It keeps the lead of the factor over the band at ranks 1, 2 and 4, not at rank 8.
\else
\Cref{tab:results} in \Cref{app:rankexp} gives A $-$ B for every setting and rank, and its caption explains how to read it.
On the format task, with random bases in place of a spectral band (\Cref{tab:controls}), $A$ frozen beats $B$ frozen by $+11.07$ points ($[+9.35, +12.84]$ at 97.5\,\%, eight seeds).
\fi
\endblock
\block{tableintrodetail}
The first experiment also ran rank 16 on the format task, and shows no diverged run there.
On the main pairs, the first experiment ran ranks 2, 16, 32 and 64.
The replication re-ran ranks 16 and 32 with new seeds and a finer grid of learning rates.
Ranks 1, 4 and 8 come from two further experiments.
The random-band control uses 24 random orthonormal bases per factor, seeds 2 to 9, and learning rates fixed at 0.01 ($B$ frozen) and 0.02 ($A$ frozen).
\endblock
\block{figintro}
\iftmlr
\paragraph{At rank 2, the advantage holds in most tasks and models.}
\Cref{fig:forest} shows ten of the settings.
\else
\Cref{fig:rank} shows A $-$ B against the rank, on every setting tested at several ranks.
\fi
\endblock
\block{figintro2}
Each point is a separate experiment with its own learning-rate selection, its own tests and eight seeds; we never pool ranks, and \iftmlr the thin lines join tested ranks only.\else we do not interpolate between them.
The thin lines join tested ranks only (three for each 7B model), and each panel has its own vertical scale.\fi
\endblock
\block{figrank}
\begin{figure}[t]
\centering
\paperfig[width=\iftmlr0.88\linewidth\else0.9\linewidth\fi]{figures/fig3_rank_profile.pdf}
\caption{\textbf{A $-$ B against the rank: established at ranks 1 to 8 on both main pairs; above rank 8, mostly on the minor band.}
\iftmlr Bars: the intervals of the tests.
Small markers at ranks 16 and 32: the first experiment, or a learning rate that tied in the selection; the large ones there come from the post hoc replication (\Cref{app:rankexp}).
Format task, rank 16, dominant band: one diverged run of $B$ frozen inflates the value (\Cref{app:collapse}).
\else Dominant band: blue circles. Minor band: vermillion squares.
Filled marks: established. Hollow marks: not established. Bars: the intervals of the pre-specified tests, each at its own level.
Small markers: the first experiment at rank 32 on the main pairs. The other main-pair points at ranks 16 and 32 come from the replication\posthoc.
Small pale circle\posthoc\ at rank 16: the same comparison at the learning rate that tied with the chosen one in the selection (no test, $95\,\%$ interval).
Format task, rank 16, dominant band: one diverged run of $B$ frozen inflates the value (\Cref{app:collapse}).
\posthoc: post hoc, listed in \Cref{app:rankexp}.
\fi}
\label{fig:rank}
\end{figure}
\endblock
\block{example}
\iftmlr
\paragraph{On the format task and on two 7B models, the frozen factor decides most of the outcome.}
These three settings give the largest gaps at rank 2 (\Cref{fig:forest}).
On the format task, $A$ frozen leads by $+12.11$ points on the dominant band and $+20.40$ on the minor band.
On OpenBookQA with Qwen2.5-7B and Mistral-7B, it leads by $+11.02$ to $+13.15$ points, and the four differences are Holm-robust.
In all three, the factor matters more than the band.
\else
\paragraph{An example: the format task with Llama-3.2-1B at rank 2.}
On this task, which factor is frozen decides most of the outcome, and the band decides little.
Freezing $A$ instead of $B$ gains $+20.40$ points on the minor band and $+12.11$ on the dominant band (A $-$ B, intervals in \Cref{app:contrasts}).
So $A$ frozen on either band beats $B$ frozen on the other band, and the factor effect $F = +16.26$ exceeds the band effect $|S|$.
$A$ frozen stays below the matched free LoRA by $1.75$ points on the minor band, $2.99$ on the dominant (no test).
\fi
\endblock
\block{obqa}
\paragraph{On OpenBookQA at rank 2, the gap closes with longer training, as $A$ frozen declines.}
OpenBookQA at rank 2 is our most fragile pair.
At the standard 49 steps, $A$ frozen wins on both bands, but $B$ frozen stays near the base model, so the comparison says little about the band (\Cref{tab:oqaprot}).
With four times the budget, $A$ frozen falls on the dominant band, from $53.25\,\%$ to $46.73\,\%$, while $B$ frozen rises from $44.65\,\%$ to $48.48\,\%$.
The gap closes: A $-$ B is $-1.75$ there, not established (\Cref{fig:budget}).
We read the fall of $A$ frozen as overfitting, a reading we have not tested; the free LoRA falls much less, from $54.97\,\%$ to $53.27\,\%$.
At each run's best validation checkpoint, the gap still holds on the minor band ($+8.45$ points) and is not established on the dominant band.

\paragraph{The way of scoring can change the sign.}
A \emph{letter prompt} lists the options by letter and scores the answer letter.
The base model answers $76.60\,\%$ with it, against $42.20\,\%$ under our main protocol.
Trained and scored with the letter prompt, $A$ frozen wins on both bands at 49 steps and on neither at 893 steps.
When adapters trained under the main protocol are scored with it, the gap reverses at rank 2, to $-15.5$ and $-37.4$ points, and we do not know why (\Cref{app:letter}).
\endblock
\block{figbudget}
\begin{figure}[t]
\centering
\paperfig[width=\iftmlr0.88\linewidth\else\linewidth\fi]{figures/budget_gap.pdf}
\caption{\textbf{With longer training, the gap persists on the format task and closes on OpenBookQA, as $A$ frozen declines.}
Rank 2, test accuracy (\%), mean over seeds; bars span the lowest and highest seed.
Lines join the last steps of separate runs.
On OpenBookQA, points after 49 steps are post hoc (\Cref{app:budgetfig}).}
\label{fig:budget}
\end{figure}
\endblock
\block{oqatable}
\begin{table}[t]
\centering
\small
\caption{\textbf{OpenBookQA, Qwen2.5-1.5B, rank 2: A $-$ B is large under likelihood at 49 steps, and small after long training in generation or in a letter format.}
Accuracy (\%), mean of eight seeds. Bold: A $-$ B is established; generation has no test. Generation and the letter prompt score the answer letter, which the base model already gives.
Letter: the design trained and scored with the letter prompt.}
\label{tab:oqaprot}
\begin{tabular}{lrrrr}
\toprule
& likelihood & generation & letter & letter \\
& 49 steps & 1,187 steps & 49 steps & 893 steps \\
\midrule
Base model & 42.20 & 76.20 & 76.60 & 76.60 \\
$B$ frozen, dominant band & 44.65 & 83.70 & 76.25 & 84.85 \\
$B$ frozen, minor band & 42.55 & 81.27 & 78.95 & 81.97 \\
$A$ frozen, dominant band & 53.25 & 85.40 & 82.20 & 85.22 \\
$A$ frozen, minor band & 52.65 & 83.75 & 81.80 & 84.40 \\
Matched free LoRA & 54.97 & 86.17 & 81.17 & 86.05 \\
\midrule
A $-$ B, dominant / minor & \textbf{+8.60} / \textbf{+10.10} & $+1.70$ / $+2.48$ & \textbf{+5.95} / \textbf{+2.85} & $+0.37$ / $+2.42$ \\
\bottomrule
\end{tabular}
\end{table}
\endblock
\block{crossed}
\iftmlr
\paragraph{The size of A $-$ B follows the task more than the model.}
At rank 2, we also ran the format task with Qwen2.5-1.5B and OpenBookQA with Llama-3.2-1B.
$A$ frozen beats $B$ frozen on both bands of both new pairs (\Cref{tab:results}).
The gains, on the dominant then the minor band, are $+14.23$ and $+12.70$ points on the format task with Qwen2.5-1.5B, and $+8.30$ and $+5.37$ on OpenBookQA with Llama-3.2-1B.
We observe that the gap is larger on the format task than on OpenBookQA with either model, on both bands.
\else
\paragraph{The size of A $-$ B follows the task more than the model.}
At rank 2, we also ran the format task with Qwen2.5-1.5B and OpenBookQA with Llama-3.2-1B.
A $-$ B is established on both bands of both new pairs: $+14.23$ and $+12.70$ on the format task with Qwen2.5-1.5B, $+8.30$ and $+5.37$ on OpenBookQA with Llama-3.2-1B.
On both bands, A $-$ B is larger on the format task than on OpenBookQA with either model (descriptive), so its size follows the task more than the model.
\fi
\endblock
\block{exampleformatdetail}
The base model never produces the format ($0.00\,\%$), so every gain is a gain over zero.
The matched free LoRA reaches $80.07\,\%$.
Freezing $B$ reaches $57.92\,\%$ on the minor band, as MiCA does, and $64.96\,\%$ on the dominant band.
Freezing $A$ reaches $78.32\,\%$ on the minor band and $77.08\,\%$ on the dominant, close to the free LoRA.
The band effect is $S = -2.90$ points (descriptive, 95\,\% interval $[-4.80, -1.02]$).
It splits into $S_A = +1.24$ points with $A$ frozen and $S_B = -7.04$ points with $B$ frozen (descriptive).
The band matters more when $B$ is frozen.
\endblock
\block{exampleobqadetail}
The base model answers $42.20\,\%$ of the 500 test questions, and the matched free LoRA reaches $54.97\,\%$.
On OpenBookQA, freezing $B$ reaches $42.55\,\%$ on the minor band, as MiCA does, and $44.65\,\%$ on the dominant band.
Freezing $A$ reaches $52.65\,\%$ on the minor band and $53.25\,\%$ on the dominant band, close to the free LoRA.
The factor effect, the mean of the two gains, is $F = +9.35$ points.
The band effect is only $S = -1.35$ points (descriptive, 95\,\% interval $[-3.16, +0.51]$).
It splits into $S_A = -0.60$ $[-1.67, +0.47]$ with $A$ frozen and $S_B = -2.10$ $[-3.01, -1.19]$ with $B$ frozen (descriptive; paired $t$ over seeds).
The band matters only when $B$ is frozen, and more so on the format task: $S_A = +1.24$ $[-0.59, +3.09]$, $S_B = -7.04$ $[-8.57, -5.50]$.
So $F$ exceeds $|S|$.
\iftmlr
This is the plain reading of \Cref{sec:tests}: each A-frozen configuration beats the B-frozen one on the other band.
\else
In plain words, $F$ exceeds $|S|$ when each A-frozen configuration beats the B-frozen one on the other band, that is, when changing the factor moves accuracy more than changing the band.
\fi
Here $A$ frozen on the minor band ($52.65\,\%$) beats $B$ frozen on the dominant band ($44.65\,\%$).
Also, $A$ frozen on the dominant band ($53.25\,\%$) beats $B$ frozen on the minor band ($42.55\,\%$).
With four times the updates (batch 8), A $-$ B stays established on both bands, so more updates at the same tokens do not remove the gap. The control does not separate the protocol from the training length (\Cref{par:floor}).
\endblock
\block{exampleobqaoc}
\label{app:oc}
The answer-only generation task is the first level (L1) of a ladder of four targets for the same OpenBookQA questions with Qwen2.5-1.5B at rank 2.
The target is the answer letter at L1, the letter and the answer text at L2, a fixed JSON object at L3, and the same object with the supporting fact at L4.
Every level trains for 1,187 optimiser steps (about 7.7 epochs of the 4,957 training questions, with one question per sequence, where the main protocol packs several rows into each sequence), not 49, and generation is scored with a parser that extracts the letter.
This differs from the main protocol (a generation target, more training steps, a parser), so these results are descriptive and not pre-specified tests of A $-$ B.
\Cref{tab:oclevels} gives A $-$ B per band and level, with $F$ and $S$ computed from the mean accuracies (descriptive).
At L1, $F$ is about equal to $|S|$.
At L2, $F$ is below $|S|$, and at L3 and L4, $F$ is above $|S|$.
Across the four levels, A $-$ B is positive in 7 of 8 combinations of band and level, and the 95\,\% interval excludes zero in 4 (the minor band at every level).

We also trained levels L1 and L3 for 74 and 297 steps, with tests fixed in advance (one family of 12).
At 74 steps, A $-$ B is established on both bands at both levels: on the dominant band $+3.48$ $[+0.02, +7.53]$ at L1 and $+5.95$ $[+2.47, +9.77]$ at L3, and on the minor band $+3.05$ and $+4.50$.
At 297 steps it is established on the minor band at both levels ($+4.35$ and $+3.20$) and not on the dominant band ($+2.75$ and $+2.90$, at most $5.90$ and $5.95$).
The change of A $-$ B between 74 and 1,187 steps is established at L3 on the dominant band only ($+4.55$ $[+0.32, +9.03]$). The other three changes are not established.
\endblock
\block{rankeight}
At rank 8, the default rank of the PEFT library~\citep{mangrulkar2022peft}, the effect holds on both main pairs.
\endblock
\block{rankeightdetail}
All four differences (A $-$ B on two bands and two pairs) are established, and $F$ exceeds $|S|$ on both pairs (\Cref{tab:results}).
On OpenBookQA with Qwen2.5-1.5B, $B$ frozen reaches $49.15\,\%$ on the dominant band and $46.45\,\%$ on the minor band.
$A$ frozen reaches $56.33\,\%$ and $53.70\,\%$, so A $-$ B is $+7.18$ points on the dominant band and $+7.25$ on the minor band.
The format task gives values of the same sign and similar size on both bands (\Cref{tab:results}).
\endblock
\block{sameholds}
\paragraph{The same holds on the format task, at 7 billion parameters and, on one band, on an instruction-tuned model.}
All results in this paragraph are at rank 2.
On both 7B models, A $-$ B is established on both bands of OpenBookQA\iftmlr.
The format task with Qwen2.5-7B, the instruction-tuned Qwen2.5-1.5B and Qwen3-1.7B-Base are in \Cref{app:rankexp}\fi.
\endblock
\block{sameholds7b}
On the format task, which the base model never produces, the effect is larger than on OpenBookQA.
There A $-$ B is established on both bands, and $F$ exceeds $|S|$.
With Qwen2.5-7B, A $-$ B is also established on the format task (post hoc) and on the minor band of ARC-Challenge (post hoc).
For lack of power, no test was planned on the dominant band of ARC-Challenge with Qwen2.5-7B (descriptive).
A $-$ B on the 7B models is of the same order as on the smaller models.
We do not test the difference between models.
\endblock
\block{sameholds2}
On OpenBookQA with the instruction-tuned Qwen2.5-1.5B, A $-$ B is established on the dominant band only ($+3.97$ on the minor band, not established, at most $8.35$), and $F$ exceeds $|S|$.
On Qwen3-1.7B-Base, a more recent model of the Qwen family, A $-$ B is established on both bands of OpenBookQA, and $F$ exceeds $|S|$.
\endblock
\block{arc}
\iftmlr
\paragraph{On ARC, the effect is small.}
On ARC with Qwen2.5-1.5B at rank 2, $A$ frozen beats $B$ frozen on the minor band only.
On the dominant band we detect no difference (at most $1.48$ points on ARC-Easy and $3.63$ on ARC-Challenge).
The test of $F$ against $|S|$ does not find the factor larger at any rank from 2 to 64 (\Cref{app:tests}).
We observe that $B$ frozen already gets $51\,\%$ (ARC-Challenge, minor band) to $94\,\%$ (ARC-Easy, dominant band) of the free LoRA's gain over the base model, which leaves $A$ frozen little to add.
\else
\paragraph{On ARC, the effect is small.}
On ARC with Qwen2.5-1.5B, A $-$ B is tested at rank 2 only, and $F$ against $|S|$ is tested at ranks 2 to 64 and not established at any of them.
At rank 2, A $-$ B is established on the minor band only (suggestive on ARC-Easy).
On the dominant band it is not established (at most $1.48$ points on ARC-Easy and $3.63$ on ARC-Challenge).
On ARC-Easy, $B$ frozen already takes $94\,\%$ of the free LoRA's gain over the base model on the dominant band ($+7.77$ of $+8.25$ points) and $84\,\%$ on the minor band, which leaves $A$ frozen little to add.
On ARC-Challenge, $B$ frozen takes $56\,\%$ and $51\,\%$ ($+3.77$ and $+3.48$ of $+6.77$ points).
\fi
\endblock
\block{arcgain}
Fine-tuning also gains less on ARC.
At rank 2 with Qwen2.5-1.5B, the free LoRA gains $+8.25$ points over the base model on ARC-Easy and $+6.77$ on ARC-Challenge, against $+12.77$ on OpenBookQA.
The share of that gain taken by $B$ frozen, not a general lack of room, is what limits A $-$ B (\Cref{tab:results}).
\endblock
\block{lowranks}
\paragraph{From rank 1 to rank 8, the result holds on both main pairs.}
A $-$ B is established on both bands at ranks 1, 2, 4 and 8, and $F$ exceeds $|S|$ at each of these ranks (\Cref{tab:results}).
One of these comparisons is post hoc: the format task at rank 1, whose experiment was decided after related results were seen.
\endblock
\block{replication}
\iftmlr
\paragraph{At rank 16, $A$ frozen still beats $B$ frozen on the minor band, where MiCA freezes $B$.}
Ranks 16 and 32 of the main pairs were run twice: a first experiment, and a replication with new seeds and a finer grid of learning rates.
\else
\paragraph{Ranks 16 and 32 were also run in a replication.}
\Cref{app:rankexp} describes the replication.
At rank 32 we report both results.
The replication's tests of A $-$ B and of $F$ against $|S|$ are post hoc.
\fi
\endblock
\block{replicationwhy}
The first experiment tests A $-$ B only at ranks 2, 32 and 64.
At rank 16, A $-$ B therefore comes from the replication only.
The first experiment's grid was too coarse in one case: on OpenBookQA, the rate selected for $B$ frozen on the minor band lay one or two grid steps below a rate at which its accuracy collapsed.
A coarse grid next to such a collapse can leave a configuration away from its best rate.
The replication uses a finer grid for this reason.
\endblock
\block{aboveeight}
\textbf{Replication (where A $-$ B is tested at rank 16).}
On the minor band, $A$ frozen beats $B$ frozen by $+3.90$ points on OpenBookQA and $+3.40$ on the format task.
On the dominant band, we detect no difference on OpenBookQA with Qwen2.5-1.5B (at most $5.42$ points).
On the format task, one diverged run of $B$ frozen inflates the dominant-band value (\Cref{app:collapse}).
\textbf{First experiment (only $F$ against $|S|$ is tested; no diverged run).}
The test finds the factor larger on the format task (\Cref{tab:results}) and neither larger on OpenBookQA (\Cref{tab:reversal}).
\textbf{Qwen2.5-7B (a separate experiment at rank 16).}
On OpenBookQA, the gain on the minor band is $+6.27$.
\textbf{Longer training (rank 16, four times the budget).}
A $-$ B stays established on both bands of the format task ($+4.13$ and $+4.11$ points, with no diverged run among the reported runs) and is not established on OpenBookQA ($+0.80$ and $+2.85$).
\endblock
\block{reversal}
\iftmlr
\paragraph{On OpenBookQA with Qwen2.5-1.5B above rank 8, the observed band effect is larger than the factor effect.}
From rank 16 on, $B$ frozen does better on the dominant band at every rank.
In all five comparisons at ranks 16, 32 and 64, the observed band effect is larger than the factor effect (\Cref{tab:reversal}).
The test is more cautious and confirms it in one of the five, rank 32 of the replication.
There $A$ frozen on the minor band trails $B$ frozen on the dominant band: the second cross difference, $F + S$, is $-4.17$ points.
\else
On OpenBookQA with Qwen2.5-1.5B, from rank 16 on, $S$ is negative and larger in size than $F$ in all five comparisons of the two experiments ($S$ from $-3.25$ to $-5.60$, $F$ from $+1.19$ to $+2.89$, no test, \Cref{tab:reversal}).
For $B$ frozen, the dominant band is the better one at every rank above 8, a consistent reversal of the low-rank pattern.
The test finds $|S|$ larger than $F$ in one of the five, at rank 32 in the replication (post hoc), although neither A $-$ B is established there.
\fi
\iftmlr\else Independently, in methods that train both factors, \citet{lee2026learningrate} see the same direction: PiSSA (dominant band) overtakes vanilla LoRA with the rank, and MiLoRA (minor band) loses its lead, within 1 to 2 points.\fi
\endblock
\block{aboveeightend}
\paragraph{With Mistral-7B, the effect lasts up to rank 64 on the minor band.}
At rank 64, $A$ frozen beats $B$ frozen on the minor band by $+4.88$ points, and we detect no difference on the dominant band (at most $3.42$ points).
\paragraph{In the eight-task mixture, the effect is small.}
A $-$ B is tested at every rank.
Its interval excludes zero on both bands up to rank 16 and on the minor band at rank 32, with gains of only $+0.73$ to $+2.19$ points.
The test of $F$ against $|S|$ runs at rank 2 only.
It finds the factor larger there, since it compares the factor with the band, not with zero: $F$ is small, but $|S|$ is smaller.
Inside the mixture, A $-$ B on the dominant band is negative on ARC-Easy, $-1.78$ points.
\Cref{app:rankexp} gives each result above rank 8 and, for each pair and band, the smallest rank at which we detect no difference.
The rank-by-factor interaction test is pending.
\endblock
\block{rankxvi}
\paragraph{At rank 16, the minor band, where MiCA freezes $B$, still shows the effect.}
At rank 16, A $-$ B on the minor band is established on both main pairs, suggestive on the format task, and on OpenBookQA with Qwen2.5-7B (post hoc) (\Cref{tab:results}).
\endblock
\block{rankxvifoot}
\iftmlr
On this pair, $A$ frozen on the minor band and the free LoRA have the same mean accuracy, $62.72\,\%$.
A $-$ B on that band therefore equals the gap between the free LoRA and $B$ frozen on the minor band in \Cref{sec:methods}.
The two intervals differ because they come from different tests.
\else
\unskip\footnote{On this pair, $A$ frozen on the minor band and the free LoRA have the same mean accuracy, $62.72\,\%$.
A $-$ B on that band therefore equals the gap between the free LoRA and $B$ frozen on the minor band in \Cref{sec:methods}.
The two intervals differ because they come from different tests.}
\fi
\endblock
\block{rankxvi2}
On the dominant band at rank 16, A $-$ B is not established on OpenBookQA with either Qwen model.
On the format task, A $-$ B on the dominant band is established, but one diverged run of $B$ frozen drives it, and its paired $t$ interval over seeds contains zero (\Cref{app:collapse}).
\endblock
\block{rankxvidetail}
The diverged run scored $0.0\,\%$, and its training loss rose far above those of the other seeds.
At the other learning rate that tied in the selection, A $-$ B on that band is $+3.58$ points (descriptive) (post hoc).
\endblock
\block{rankxxxii}
\paragraph{At rank 32 on OpenBookQA with Qwen2.5-1.5B, the replication (post hoc) shows the band effect exceeding the factor effect.}
There $|S|$ exceeds $F$, although neither A $-$ B is established (\Cref{tab:results}).
\endblock
\block{rankxxxiidetail}
In plain words, $B$ frozen on the dominant band beats $A$ frozen on the minor band.
The test rests on the difference between these two configurations, $F + S = -4.17$ points (interval $[-8.00, -0.39]$ at level $98.75\,\%$).
At rank 32, both A $-$ B differences are positive but not established (\Cref{tab:results}).
The gap between the two bands for $B$ frozen accounts for the rest.
The first experiment at rank 32 left this comparison not established.
\endblock
\block{rankxxxii2}
On the format task, the two rank-32 experiments disagree on the dominant band.
\endblock
\block{stops}
\paragraph{Above rank 8, the ranks at which the effect is not established depend on the band.}
To read the profile, we use one rule (post hoc): for each pair and band, we name the smallest tested rank at which A $-$ B is not established.
The ranks between 8 and 16 were not tested.
On OpenBookQA with Qwen2.5-1.5B, A $-$ B is first not established at rank 16 on the dominant band, and at rank 32 on the minor band.
On the format task, it is first not established at rank 32 on the minor band.
\endblock
\block{stopsno}
No rank is named for the dominant band of the format task, because the two rank-32 experiments disagree there.
\endblock
\block{stops3detail}
With 500 or 614 test items, effects of a few points at these ranks are often too small to detect.
``Not established'' therefore leaves room for a real effect.
\endblock
\block{stops3}
\Cref{app:notest} gives how large a real effect could still be when A $-$ B is not established.
The rank-by-factor interaction is post hoc, because the direction was already visible; its contrast was fixed before it was computed.
On each main pair, it is the mean of $F$ at ranks 1, 2, 4 and 8 minus the mean of $F$ at ranks 16, 32 and 64.
It is $+6.51$ $[+5.69, +7.33]$ points on OpenBookQA and $+13.08$ $[+11.63, +14.54]$ on the format task, $+8.34$ $[+7.20, +9.49]$ without rank 1 (paired $t$ over seeds), and it is positive on each band.
\endblock
\block{stops3tail}
Here $F$ is A $-$ B averaged over the two bands, and each rank counts once.
The two pairs are not pooled.
The interval of the interaction test is a bootstrap over seeds and test items, with seeds drawn independently at each rank.
That independence widens the interval, and the format task has only 614 test items.
\endblock
\block{mistral}
\paragraph{With Mistral-7B, the minor band keeps the effect up to rank 64 (post hoc).}
On OpenBookQA, A $-$ B is established on both bands at rank 32, with $F$ exceeding $|S|$ (\Cref{tab:results}).
It is also established on the minor band at rank 64 (post hoc).
\endblock
\block{mistralcontrast}
This contrasts with Qwen2.5-1.5B on the same task, where A $-$ B is not established on either band at rank 32.
\endblock
\block{mistraldetail}
Mistral-7B and Qwen2.5-1.5B ran on the same model of graphics card, so the two do not differ in hardware.
At rank 64, $A$ frozen on the dominant band chose its learning rate at the edge of the grid, on an unresolved tie.
It may be slightly under-tuned.
\endblock
\block{mistraltail}
\iftmlr
At rank 64, A $-$ B on the dominant band is not established, and we do not read it as the absence of an effect (\Cref{app:rankexp}).
\else
We therefore do not read the dominant band at rank 64 as the absence of an effect (\Cref{app:rankexp}).
\fi
\endblock
\block{llama}
\paragraph{With an 8B model at rank 8, A $-$ B is established on one band of each task.}
With Llama-3.1-8B, our largest model, the result is weaker.
At rank 8, $A$ frozen leads on the minor band of OpenBookQA ($+7.02$ $[+3.02, +10.92]$) and on the dominant band of the format task ($+4.38$ $[+2.04, +6.80]$).
On the other band of each task we detect no difference (at most $6.45$ and $4.09$ points), nor at rank 16 on OpenBookQA.
In none of the three settings, OpenBookQA at ranks 8 and 16 and the format task at rank 8, do we know whether the factor or the band matters more (\Cref{app:llama8b}).
\endblock
\block{mixture}
\paragraph{In the eight-task mixture, the effect is small: A $-$ B is $+0.73$ to $+2.19$ points where established.}
With Qwen2.5-1.5B, a single adapter is trained on all eight tasks, and we test A $-$ B on its mean accuracy.
\endblock
\block{mixturegain}
Fine-tuning gains little here: at rank 2, the base model scores $58.44\,\%$ and the free LoRA $63.93\,\%$.
\endblock
\block{mixture2}
A $-$ B is established on both bands up to rank 16, and on the minor band at rank 32 (\Cref{tab:results}).
\iftmlr Inside the mixture, A $-$ B on the dominant band is established negative on ARC-Easy (\Cref{app:rankexp}).\fi
\endblock
\block{mixture2detail}
Ranks 16 and 32 are post hoc (\Cref{sec:prespec}).
Two values are suggestive (established, with the interval end within 0.5 points of zero): $+0.73$ on the dominant band at rank 16 and $+0.84$ on the minor band at rank 32.
Most learning-rate selections in the mixture ended on unresolved ties (\Cref{tab:ties}).
\endblock
\block{insidemix}
\paragraph{Inside the mixture, freezing \texorpdfstring{$A$}{A} on the dominant band is worse on ARC-Easy.}
\iftmlr\else We ran secondary tests on each of the seven tasks above the floor, at level $99.69\,\%$.\fi
On ARC-Easy, A $-$ B on the dominant band is $-1.78$ points $[-3.37, -0.19]$ (established negative, suggestive)\iftmlr\ (\Cref{app:rankexp} gives the other tests)\fi.
\endblock
\block{insidemixobqa}
\iftmlr We ran secondary tests on each of the seven tasks above the floor, at level $99.69\,\%$.\fi
On that task, $A$ frozen is below $B$ frozen on both bands, so $F$ is negative.
The test of $F$ against $|S|$ finds $|S|$ larger, but this reflects the sign of $F$.
It does not show that the band moves accuracy more.
On OpenBookQA inside the mixture, A $-$ B is not established.
It is $+1.73$ points on the dominant band (not established, at most $5.52$) and $+3.25$ on the minor band (not established, at most $7.06$).
\endblock
\block{gsm}
\paragraph{On GSM8K, the factor also matters more than the band at rank 2.}
With Llama-3.2-1B at rank 2, trained for two epochs, $A$ frozen beats $B$ frozen by $+5.62$ points on the dominant band and $+6.57$ on the minor band, and the factor matters more than the band.
At rank 8, the gap holds on both bands ($+3.75$ and $+5.24$), but whether the factor matters more than the band is open (\Cref{tab:gsm}).
The GSM8K tests are not part of the Holm correction.
\endblock
\block{gsmtable}
\begin{table}[!htbp]
\centering
\small
\caption{\textbf{GSM8K with Llama-3.2-1B: A $-$ B is established on both bands at ranks 2 and 8, and the factor is larger than the band at rank 2.}
A $-$ B in accuracy points, with its interval (97.5\,\%), eight seeds. The five-shot base model scores $6.44\,\%$.
Ranks 16 and 32 are descriptive: no test, no status.}
\label{tab:gsm}
\begin{tabular}{rlll}
\toprule
rank & A $-$ B, dominant band & A $-$ B, minor band & verdict \\
\midrule
2 & $+5.62$ $[+3.81, +7.41]$ & $+6.57$ $[+4.89, +8.27]$ & factor \\
8 & $+3.75$ $[+2.15, +5.39]$ & $+5.24$ $[+3.20, +7.28]$ & open \\
16 & $+1.98$ (descriptive) & $+4.47$ (descriptive) & no test \\
32 & $+0.75$ (descriptive) & $+3.75$ (descriptive) & no test \\
\bottomrule
\end{tabular}
\end{table}
\endblock
\block{gsm8b}
\begin{table}[!htbp]
\centering
\small
\caption{\textbf{GSM8K with Llama-3.1-8B: A $-$ B is established on both bands at ranks 8 and 16, and the verdict is factor at both ranks.}
A $-$ B in accuracy points, with its interval (97.5\,\%), eight seeds (2 to 9), 1,319 test problems. The five-shot base model scores $55.12\,\%$.
The last column is the matched free LoRA minus $B$ frozen on the minor band.
The experiment was chosen post hoc and is outside the Holm family.}
\label{tab:gsm8b}
\begin{tabular}{rllll}
\toprule
rank & A $-$ B, dominant band & A $-$ B, minor band & verdict & free LoRA $-$ $B$ frozen \\
\midrule
8 & $+4.63$ $[+2.62, +6.70]$ & $+6.89$ $[+4.78, +8.99]$ & factor & $+7.75$ $[+5.62, +9.91]$ \\
16 & $+4.62$ $[+2.71, +6.59]$ & $+7.53$ $[+5.35, +9.67]$ & factor & $+7.74$ $[+5.73, +9.78]$ \\
\bottomrule
\end{tabular}
\end{table}

\paragraph{GSM8K with Llama-3.1-8B: scope and declarations.}
\label{app:gsm8b}
We decided to run this experiment after reading the GSM8K results with Llama-3.2-1B, so the choice is post hoc.
Its protocol and its tests were written before its runs.
Training lasts two epochs, 436 steps, with a learning rate selected for each configuration.
Mean test accuracies at rank 8 are $65.10\,\%$ for the free LoRA, $64.83\,\%$ and $64.23\,\%$ for $A$ frozen on the dominant and minor bands, and $60.20\,\%$ and $57.34\,\%$ for $B$ frozen.
At rank 16 they are $65.76\,\%$, $66.39\,\%$ and $65.55\,\%$, then $61.76\,\%$ and $58.02\,\%$.
$F$ is $+5.76$ points and $S$ is $-1.72$ at rank 8; they are $+6.08$ and $-2.29$ at rank 16.
With Llama-3.2-1B at rank 8 the verdict was open (\Cref{tab:gsm}); we compare the two models by status only.
$A$ frozen minus the free LoRA is descriptive: $-0.27$ $[-1.84, +1.30]$ and $-0.86$ $[-2.62, +0.82]$ points at rank 8, $+0.63$ $[-0.88, +2.10]$ and $-0.21$ $[-1.73, +1.34]$ at rank 16 (95\,\%).
Two rate selections ended on an unresolved tie, $A$ frozen on the dominant band at rank 8 and on the minor band at rank 16.
No run collapsed, and none was excluded.
The experiment used $353.13$ GPU-hours, above the projection of $340.7$ and under its cap of 420.
Our interventions on the job queue change no run and no test.
We lowered the number of concurrent jobs, kept our jobs off busy machines, and restarted the pipeline by hand after each stall.
Two events touched the procedure itself.
First, the rate-selection jobs were moved once after a bug in our driver script.
Second, an author checked the answer extractor by hand on three blind samples of model outputs, without consulting any model.
The second sample failed at its first round, with 111 agreements on 120: the nine disagreements were outputs that loop until the token limit.
Before any score was read, we added a rule that reads a looping output as no answer, and a new blind sample of 120 outputs then agreed on 120 of 120.
The other two samples agreed on 50 of 50 and 59 of 60 outputs, and each threshold was fixed in advance.
\endblock
\block{oclevelstab}
\begin{table}[!htbp]
\centering
\small
\caption{\textbf{A $-$ B is positive on the minor band at every level, and $F$ is about equal to $|S|$ only at L1.}
OpenBookQA with Qwen2.5-1.5B at rank 2, trained as a generation task, in accuracy points (descriptive).}
\label{tab:oclevels}
\begin{tabular}{lrrrr}
\hline
Level & A $-$ B dominant & A $-$ B minor & $F$ & $S$ \\
\hline
L1 & $+1.70$ & $+2.48$ & $+2.09$ & $-2.04$ \\
L2 & $-0.05$ & $+2.62$ & $+1.29$ & $-1.66$ \\
L3 & $+1.40$ & $+3.22$ & $+2.31$ & $-1.09$ \\
L4 & $+1.15$ & $+3.17$ & $+2.16$ & $-1.27$ \\
\hline
\end{tabular}
\end{table}
\endblock

\fi%

\paragraph{Summary.}
At ranks 1 to 8, with short training, the frozen factor matters more than the band on the two
main pairs.
The result narrows outside that range, and inside it in the settings of \Cref{sec:narrow}.

\secmark{methods}
\section{At their own settings, MiCA and PiCa show a smaller factor effect or none detected}
\label{sec:methods}

Does the factor result carry over to MiCA \citep{rudiger2026mica} and PiCa \citep{hwang2026pica}
themselves?
Their ranks and recipes start at rank 16, so we test the two methods at their own settings.
For MiCA we use three setups.
The first is our protocol, with $B$ frozen on the minor band and a rate selected per
configuration.
The second is MiCA's recipe: the PEFT library, rank 16 and the published learning rate.
The third keeps that library and rank and selects a rate for each method.

\begin{table}[!ht]
\centering
\small
\caption{Under our protocol, the matched free LoRA beats MiCA's configuration in all seven
comparisons.
On OpenBookQA, we detect no difference between the free LoRA and $B$ frozen on the dominant band.
Differences are in accuracy points.
First block: matched free LoRA minus MiCA's configuration ($B$ frozen, minor band).
Second block: the same free LoRA minus $B$ frozen on the dominant band.
``Barely'' means a lower bound within 0.5 points of zero.
The second block is post hoc, with descriptive 95\% intervals.}
\label{tab:gap}
\begin{tabular}{llrll}
\toprule
Task and model & rank & difference & interval & reading \\
\midrule
OpenBookQA, Qwen2.5-1.5B & 16 & $+5.85$ & $[+2.37, +9.38]$ & free LoRA ahead; Holm-robust \\
OpenBookQA, Qwen2.5-1.5B & 32 & $+4.25$ & $[+0.85, +7.90]$ & free LoRA ahead \\
Format task, Llama-3.2-1B & 16 & $+4.78$ & $[+1.89, +7.90]$ & free LoRA ahead \\
Format task, Llama-3.2-1B & 32 & $+3.09$ & $[+0.02, +6.03]$ & free LoRA ahead, barely \\
OpenBookQA, Qwen2.5-7B & 16 & $+6.27$ & $[+2.83, +9.67]$ & free LoRA ahead (post hoc) \\
OpenBookQA, Qwen2.5-7B & 32 & $+3.75$ & $[+0.35, +7.13]$ & free LoRA ahead, barely \\
OpenBookQA, Mistral-7B & 32 & $+3.90$ & $[+0.25, +7.70]$ & free LoRA ahead, barely \\
\midrule
\multicolumn{5}{l}{Free LoRA minus $B$ frozen on the dominant band} \\
OpenBookQA, Qwen2.5-1.5B & 16 & $+1.40$ & $[-1.68, +4.52]$ & interval includes zero \\
OpenBookQA, Qwen2.5-1.5B & 32 & $-1.62$ & $[-4.55, +1.28]$ & interval includes zero \\
OpenBookQA, Qwen2.5-1.5B & 64 & $-1.28$ & $[-4.02, +1.42]$ & interval includes zero \\
OpenBookQA, Mistral-7B & 32 & $+0.45$ & $[-2.20, +3.10]$ & interval includes zero \\
OpenBookQA, Mistral-7B & 64 & $-0.62$ & $[-3.32, +2.05]$ & interval includes zero \\
\bottomrule
\end{tabular}
\end{table}

\paragraph{The matched free LoRA beats MiCA's configuration in all seven comparisons.}
Under our protocol, we compare MiCA's configuration with the matched free LoRA on four models
(\Cref{tab:gap}, first block).
The free LoRA is ahead by $+3.09$ to $+6.27$ points in all seven, in three of them barely.
Only one row is Holm-robust, OpenBookQA with Qwen2.5-1.5B at rank 16.
The three other rows on the main pairs are established and not Holm-robust, and the three 7B rows
are outside the Holm family.
\iftmlr\else
On OpenBookQA, the free LoRA has slightly fewer trainable parameters than MiCA's configuration,
which favours MiCA's configuration.
On the format task it has slightly more.
There we do not separate the effect of size from the effect of the frozen factor (\Cref{tab:g1}).
\fi%
The two other setups give the same sign, with gaps of other sizes (\Cref{app:methodarms}).

\paragraph{Flipping MiCA's frozen factor helps on the format task only.}
Flipped MiCA freezes $A$ on the minor band and keeps MiCA's recipe.
We compare it with MiCA under two matchings, the same rank 16 and about MiCA's
trainable-parameter count.
On the format task with Llama-3.2-1B, flipped MiCA is ahead by $+3.48$ and $+3.50$ points under
the two matchings.
Both differences are established, and neither is Holm-robust (\Cref{tab:methods,tab:holm}).
On OpenBookQA with Qwen2.5-1.5B, we detect no difference (at most 5.83 points).
With Qwen2.5-7B on the eight-task mixture, the flip changes accuracy by $+0.37$ points, and we
detect no difference (at most 1.05 points, outside the Holm family).
We tested flipped MiCA in these three settings only (\Cref{app:flip7}).

\paragraph{On OpenBookQA at MiCA's ranks, the minor band costs more than the frozen factor.}
The second block of \Cref{tab:gap} separates the band from the factor on OpenBookQA.
With $B$ frozen on the dominant band, the free LoRA is no longer ahead.
The five differences range from $-1.62$ to $+1.40$ points, and every interval includes zero.
The comparison is descriptive and post hoc.
With Qwen2.5-1.5B, a separate pre-specified test finds that moving $B$ frozen from the minor band
to the dominant band gains $+4.45$ points at rank 16 and $+5.87$ at rank 32, both established.
On OpenBookQA at these ranks, the deficit of MiCA's configuration thus comes with its band.
For MiCA, the factor matters on the format task and the band on OpenBookQA.

\paragraph{PiCa leads a free LoRA of its size only at its published learning rate.}
For PiCa we ask whether its gain survives against a free LoRA of about its trainable-parameter
count.
We use PiCa's recipe at rank 256, a rank from its paper, and at rank 16.
With the published rate for both methods, three of the four settings pass the gate of
\Cref{sec:prespec}, and OpenBookQA at rank 16 does not.
PiCa leads by 6.05 points on OpenBookQA at rank 256 and on the format task at rank 16.
At that rate both leads are established; the equal values are a coincidence.
On the format task at rank 256, we detect no difference: the interval runs from PiCa ahead by
4.23 points to the free LoRA ahead by 2.70.
We then select the rates as in our protocol, on two seeds kept apart from the eight reported.
With selected rates, we detect no difference in any setting tested (\Cref{app:picatable}).
\iftmlr\else
The tests at the published rate and the tests at selected rates share no correction for multiple
tests.
\fi%
Our rule, fixed in advance, states the weaker result, so we claim no lead for PiCa.
Its lead holds only at its published rate \citep{lee2026learningrate}.
\iftmlr\else
PiCa's recipe is also short in steps: 15 on our packed OpenBookQA training set and 150 on the
format task.
At 15 steps and rank 256, the free LoRA still gains $+5.65$ points over the base model.
At rank 16 its gain is $-0.42$ points, so it does not learn the task and we read no test.
On ARC-Challenge at rank 256, its gain over the base model is $-1.03$ points, with no test.
\fi%

At rank 256 and PiCa's parameter count, PiCa and $B$ frozen on the dominant band do not differ
detectably on either main pair, so we make no claim for $A$ frozen there.

The factor result of \Cref{sec:rank} is thus a low-rank result.
At the methods' own settings, we detect it only for flipped MiCA on the format task.

\secmark{robustness}
\section{The rank-2 advantage survives most controls, not longer training on OpenBookQA}
\label{sec:robustness}

Is the advantage of $A$ frozen at rank 2 an artefact of our protocol?
We test five objections (\Cref{tab:robustdraft}).
Three leave the advantage in place, and a fourth changes no
configuration detectably.
The fifth, the longer budgets of \Cref{sec:rank}, passes on the format task and not on OpenBookQA.

\begin{table}[!htb]
\centering
\small
\caption{\textbf{The rank-2 advantage of $A$ frozen over $B$ frozen passes each test on the task where it was run, except a longer training budget on OpenBookQA.}
Results are A $-$ B in accuracy points on both bands.
``Holds'' means that A $-$ B stays established (\Cref{tab:statusterms}).
The last row instead compares each configuration with its rotated basis. Intervals and test levels: \iftmlr\Cref{tab:robustfull}\else\Cref{app:controls}\fi, and \Cref{app:x16} for the $16\times$ budget.}
\label{tab:robustdraft}
\begin{tabular}{>{\raggedright\arraybackslash}p{3.6cm}>{\raggedright\arraybackslash}p{4.4cm}>{\raggedright\arraybackslash}p{6.2cm}}
\toprule
Objection & Test & Result \\
\midrule
Rate-selection seeds & Eight new seeds, same rates & Holds on both tasks: $+9$ to $+19$ \\
$A$ frozen trains more parameters & $B$ frozen given 1.09 to 1.30$\times$ the parameters (format task) & Holds: $+3$ to $+14$ \\
The learning rate of $B$ frozen & Every rate of its grid and two above it & Holds at every rate (post hoc) \\
A short training budget & 16 times the budget (format task), 4 times (OpenBookQA) & Holds on the format task ($+10.83$, $+16.48$). On OpenBookQA, it holds only on the minor band at the best checkpoint, by $+8$; at the last step, both intervals include 0 (\Cref{sec:narrow}) \\
The basis inside the band & A random rotation of the frozen basis inside its band & Rotated and original bases do not differ detectably on the dominant band, the only one tested (at most 5 points) \\
\bottomrule
\end{tabular}
\end{table}

\paragraph{The extra parameters of $A$ frozen do not explain it.}
On OpenBookQA, we adapt only the query and output projections of attention.
Their input and output sizes are equal, so the two parameter counts are equal.
A $-$ B is then $+7.93$ points and is established.
On the format task, $B$ frozen collapsed on these modules (mean accuracy below 45\%), so we instead gave $B$ frozen 1.09 to 1.30 times the parameters of $A$ frozen.
$A$ frozen stays ahead, by $+3$ to $+14$ points.

\paragraph{The advantage does not need a spectral band.}
We freeze $A$ at its random initialisation, as in LoRA-FA \citep{zhang2023lorafa}, and compare
with $B$ frozen.
A $-$ B is then $+9.62$ to $+18.42$ points and is established.
With $A$ or $B$ frozen on random orthonormal bases, the format task gives $+11.07$ points on
eight seeds.
That second control has fixed learning rates and is outside the Holm family.

\paragraph{Changing one training element at a time leaves the advantage established, with one
exception.}
The elements are the scale $\alpha/r$, the optimiser, the step size per module, the number of updates
and the amount of data (\Cref{tab:controls}).
With one eighth of the OpenBookQA training data, we detect no difference on the dominant band (at
most 6.70 points).
GaLore \citep{zhao2024galore} restricts the optimiser state and not the weights.
At rank 2 on OpenBookQA, a GaLore projection on the input side beats one on the output side by $+9.62$ points.
At rank 32 we detect no difference (\Cref{app:robust}).

\paragraph{Two results go against parts of our reading.}
LoRA+ \citep{hayou2024loraplus} gives $B$ a larger learning rate than $A$.
A free LoRA trained that way beats $A$ frozen on the dominant band of the format task.
The result is established and not Holm-robust, and it weakens any statement that $A$ frozen
matches a free LoRA.
We also derived a predictor of A $-$ B from the one-layer least-squares analysis of
\citet{zhu2024asymmetry}.
Over 40 configurations, its correlation with the measured A $-$ B is negative, in a post hoc
analysis (\Cref{app:robust}).

\paragraph{Freezing $A$ has a cost in merging and forgetting, and none in memory or time.}
We add the adapters of OpenBookQA, ARC-Easy and ARC-Challenge to one model and score each task
against its own adapter.
On the minor band, $A$ frozen loses more than $B$ frozen, by 8.59 points at rank 2 and 4.28 at
rank 32.
Both differences are established, and most of the loss falls on ARC-Easy.
On the dominant band, we detect no difference.
On HellaSwag \citep{zellers2019hellaswag}, $A$ frozen forgets more than $B$ frozen in 4 of 16
tests, and less in none.
Memory and time per step match within 0.122 GiB and 0.007 seconds (\Cref{app:memtime}).
\iftmlr\else
\Cref{app:robust} gives the other side effects, among them intruder dimensions, and
\Cref{app:controls} the controls on the training setup.
\fi%

\secmark{discussion}
\section{Discussion: the factor result holds at low rank, and its cause stays open}
\label{sec:discussion}

\iftmlr\else
\paragraph{What the experiments support.}
At ranks 1 to 8 with short training, on the two main pairs, the frozen factor moves accuracy more
than the band.
\Cref{app:holds} lists the settings where the main result holds.
\fi%

\paragraph{Why the frozen factor matters: a hypothesis.}
We state one hypothesis: the output directions that the update needs move during training, and the input directions move
less.
A trained $B$ follows them, and a $B$ frozen before training does not.
Freezing $B$ would then cost more than freezing $A$.
One intervention and one descriptive analysis agree with the hypothesis.
In the intervention, we set how much the basis of $B$ frozen overlaps the output directions that a
free LoRA learns.
At rank 2, with a test fixed in advance, accuracy rises with the overlap on both main pairs, and
the correlation is established (\Cref{app:overlap}), as it is post hoc at ranks 8 and 16.
On OpenBookQA its rank-2 interval is skewed, and at rank 8 it is not Holm-robust.
That intervention uses an oracle and is not a method.
The descriptive analysis is post hoc and uses one free LoRA per main pair.
It measures the overlap of the leading gradient subspaces at step 0 and at the last step, rescaled
to 0 for random subspaces and 1 for identical ones.
On OpenBookQA at rank 2, the rescaled overlap is 0.415 on the input side and 0.108 on the output
side.
The output side is lower in 88\% of the 196 adapted modules, as on the format task
(\Cref{app:gradsides}).

Two results do not fit the hypothesis.
$A$ frozen at a random initialisation still beats $B$ frozen.
On OpenBookQA, unlike on the format task, the accuracy of $A$ frozen does not rise detectably with
its overlap with the learned input directions.
A direct test on the input side did not resolve the question (\Cref{app:ig}).
On the format task at rank 2, the input directions a trained adapter ends with gain $+3.85$ points
over the band (established, outside the Holm family), and those of the step-0 gradient recover
$+0.96$ of these 3.85 points (not established), with one failed control check.
The hypothesis stays open.

\paragraph{Practical reading.}
A frozen factor saves little memory or time (\Cref{sec:robustness}), and it costs accuracy.
At rank 2 on the two main pairs, $A$ frozen is 1.72 to 2.99 points below the matched free LoRA
(descriptive; one of the four intervals excludes zero).
For a comparison of frozen-subspace variants, we report or fix five things:
\begin{enumerate}[nosep]
\item The frozen factor, and not only the band.
\item The same band with the other factor frozen, which separates factor from band.
\item A learning rate selected per configuration, by a rule fixed before the runs.
\item Exact trainable counts, and a matched free LoRA of about the same count.
\item A test at the method's own ranks, since a gain at rank 2 need not hold at rank 32.
\end{enumerate}

\paragraph{Limitations.}
\begin{itemize}[nosep]
\item \emph{Scope.} We propose no new method.
We do not extend the claim beyond ranks 1 to 8, short budgets, the two main pairs and the rank-2
replications.
The verdict \emph{factor} at rank 16 on GSM8K comes from one post hoc experiment.
We ran longer training on the two main pairs only.
\item \emph{Models and tasks.} We cover three model families, and the largest model has 8B
parameters.
We have one synthetic generation task, the format task, and one reasoning task, GSM8K.
Its supervised tests use two models, and a reinforcement-learning run with a third detected no
difference (\Cref{app:pending}).
Four other generation targets, trained for 1,187 steps, give A $-$ B between $-0.05$ and $+3.22$
points, without tests.
On ARC at the best checkpoint, the advantage remains in one test of four (\Cref{app:robust}).
\item \emph{Side effects.} On sentences with two entities, absent from the format task's data,
$A$ frozen on the dominant band scores 4.92\% against 33.82\% for $B$ frozen (descriptive,
\Cref{app:transfer}).
At rank 16 on OpenBookQA, $A$ frozen forgets far more on WikiText (\Cref{tab:forgetrk}).
\item \emph{Intervals.} Our intervals reflect the data order, the initial $A$ of the free LoRA and
the test items, not the choice of learning rate, task or model.
\item \emph{Tests.} The Holm correction is post hoc.
$F$ and $S$ are descriptive, and only the cross differences are tested.
We fixed the tests in our own version history, not in an external registry.
\item \emph{Controls.} Most controls reuse the seeds and test items of the main experiments, so
they are not independent confirmations.
Most controls run at rank 2 only, and some use fewer seeds.
\end{itemize}

\paragraph{Conclusion.}
MiCA and PiCa differ in band and in frozen factor, and we crossed the two.
At ranks 1 to 8, with short training, on the two main pairs, the frozen factor matters more than
the band.
The result repeats at rank 2 with two 7B models, and on GSM8K with a 1B model and, post hoc, an
8B model.
Above rank 8, under longer training on OpenBookQA and at the methods' own settings, the effect is
smaller or not detected.
At low rank, we recommend holding the frozen factor fixed before attributing a difference
between frozen-subspace methods to the band.
\iftmlr Code, protocols and run logs are in the supplement, to be released.\else

\secmark{repro}
\section{Reproducing the results}
\label{sec:repro}
\def\wantedblock{repro}\immediate\write\secposfile{block arxiv_only repro}
\fi
\let\FloatBarrier\layoutFloatBarrier

\FloatBarrier
\secmark{references}
\bibliography{refs}
\bibliographystyle{tmlr}

\appendix
\secmark{appendix}
\iftmlr
\section{Statistical details}
\label{app:stats}
\label{app:rules}
\def\wantedblock{bootstrap}\immediate\write\secposfile{block appendix bootstrap}\block{bootstrap}
\paragraph{The bootstrap.}
Each comparison is a difference of test accuracies between two configurations.
Its uncertainty comes from the training seed and the finite test set.
We capture both with a paired bootstrap.
In each of 10,000 resamples, we draw the eight seeds and the test items with replacement.
We apply the same draw to every configuration and recompute the difference.
We report the central percentile interval of the resampled differences.
Its \emph{level} is the share of resamples it contains.
Rerunning the same configurations with the same seeds changes the result slightly.
For OpenBookQA at ranks 2, 16, 32 and 64, the eight-seed means of the reruns differ from the first runs by $-0.37$ to $+0.55$ points.
No configuration differs by more than 1 point.
\endblock
\block{levels}
\paragraph{Levels.}
Each experiment fixes in advance which comparisons are tests, and at which level.
The tests corrected together form its \emph{family}.
When the family holds $m$ tests, each is at level $1 - 0.05/m$: 97.5\,\% for two, 98.75\,\% for four, 99.375\,\% for eight.
The two tests of A $-$ B on each band at rank 2 on a main pair are an example: their interval, in \Cref{fig:design}, is at level 97.5\,\%.
Tests outside the main family are \emph{secondary}, with their own stated level.
The test of $F$ against $|S|$ has four verdicts (\Cref{sec:tests}).
\emph{Factor}: $F - S$ and $F + S$ are both established positive, each at level 97.5\,\%; both must hold, so no further correction is needed.
\emph{Band}: either is established negative, each tested at the stricter level $1 - 0.05/4$, that is 98.75\,\%.
\emph{Open}: neither holds.
\emph{No test}: the experiment has no such test.
\endblock
\block{bounds}
\paragraph{Bounds and descriptive values.}
When a comparison is not established, we write ``at most $X$ points'', where $X$ is the upper end of the interval, the end on the side of the expected effect.
For a contrast without an expected direction, $X$ is the larger absolute end.
A comparison is descriptive when its experiment did not plan it as a test, or declared in advance that it had too little statistical power.
Descriptive values are given with a 95\,\% interval when one exists.
A few experiments test a prediction fixed in advance rather than a difference.
When the interval excludes zero on the side opposite to the prediction, the prediction is \emph{contradicted}.
\endblock
\block{posthoc}
\paragraph{How each post hoc decision arose.}
This paragraph gives the order of events behind the marks \dag\ of \Cref{sec:prespec}.
The protocol of each experiment was committed to our version history before its first run.
It fixed the rules for selecting learning rates, the seeds and the tests.
In the first kind of \dag, the test was written before the runs it reads, except at rank 1 on the format task.
There, nine runs with $B$ frozen came from an earlier experiment, and their results were known before the rank-1 experiment was planned, so the comparison is not blind (\Cref{sec:rank}).
\begin{itemize}[leftmargin=*, itemsep=0pt]
\item The first experiment ran ranks 2, 16, 32 and 64. Two further experiments ran ranks 1, 4 and 8.
\item The replication at ranks 16 and 32 was decided after ranks 4 and 8 had been read.
  Its tests of A $-$ B and of $F$ against $|S|$ were added then, before it ran.
  It uses a finer grid of learning rates, because the first experiment's grid was too coarse in one case.
  Its protocol also fixed how to read the profile of A $-$ B across ranks (\Cref{sec:rank}); that rule is marked \dag.
\item Qwen2.5-7B at rank 16 was chosen after the other 7B results (ranks 2 and 32) had been read.
\item The eight-task mixture at ranks 16 and 32 was run once ranks 2 and 8 had been seen.
\item Mistral-7B at rank 64 was run after ranks 2 and 32 had been read.
\item Flipped MiCA was suggested by the rank-16 results of \Cref{sec:rank}. Its tests were fixed before it ran.
\end{itemize}
\Cref{tab:deviations} lists the other decisions that followed results already read.
\endblock
\block{minorband}
\paragraph{The minor band.}
On Qwen2.5-1.5B, for example, the minor band of the query and output projections of attention holds singular values below the
resolution of 16-bit numbers.
On the output projection, one of them is numerically zero.
We use the band as the methods define it.
\endblock
\block{edge}
\paragraph{The edge rule.}
If the best rate is at the edge of the grid, the true best rate may lie outside it.
We then add one more step in that direction (for example $5 \times 10^{-5}$ below $10^{-4}$) and select again, once.
Two experiments whose protocols allow two steps (SGD and the LoRA-FA initialisation) are the exception.
If the best rate is still at the new edge, we keep it and report it as \emph{at the edge of the grid}.
\endblock
\block{tie}
\paragraph{The tie rule.}
Two rates are \emph{tied} when their mean validation accuracies differ by 0.005 or less.
We train both on three more seeds (30, 31 and 32) and keep the one with the better mean over the five seeds.
If the lead is still 0.005 or less, we keep that rate and report an \emph{unresolved tie}: the data cannot tell the two rates apart, and the result could have been reported at the other one.
Where a tie matters for a result, we also give the result at the other tied rate.
\endblock
\block{contrastsintro}
\Cref{tab:contrasts} gives A $-$ B for every setting tested at several ranks, with the interval of its pre-specified test.
A $-$ B is the accuracy of $A$ frozen minus that of $B$ frozen on the same band.
The table is behind \Cref{fig:rank}, and \Cref{fig:forest} shows A $-$ B, $F$ and $S$ at rank 2 for each task and model in the figure.

The intervals are at level 98.75\,\% for the first experiment at ranks 32 and 64 on the two main pairs, and 97.5\,\% otherwise.
The first experiment uses seeds 2 to 9; ``first experiment'' in \Cref{tab:results,tab:contrasts} and ``first exp.'' in \Cref{tab:reversal} refer to it.
The replication supplies ranks 16 and 32 on the main pairs, with seeds 12 to 19.
The two experiments are never pooled.
\Cref{tab:contrasts}, \Cref{tab:g1} and \Cref{tab:results} use the same seeds.
\Cref{tab:results} (\Cref{app:rankexp}) follows the same rules, with eight seeds per configuration.
Its column of verdicts gives the pre-specified test of $F$ against $|S|$.

\Cref{tab:g1} compares MiCA's configuration, $B$ frozen on the minor band, with the matched free LoRA (\Cref{sec:configs}).
A positive value means the matched free LoRA is better.
The intervals are at level 97.5\,\%.
The primary tests are the main pairs at ranks 16 and 32.
The secondary tests are ranks 4 and 8, Qwen2.5-7B at ranks 2 and 32, and Mistral-7B.
For the other Qwen2.5-7B rows, the protocol gives no label.
\endblock
\block{figforest}
\begin{figure}[!htbp]
\centering
\paperfig[width=\iftmlr0.86\linewidth\else\linewidth\fi]{figures/fig2_forest_rank2.pdf}
\caption{\textbf{At rank 2, the frozen factor matters more than the band in seven of the nine settings shown; a tenth has no test.}
Each row is one task and model.
\iftmlr
Left: A $-$ B per band with its interval; filled: established, hollow: not established, pale: no test.
Right: $F$ and $S$, and the verdict of the test of $F$ against $|S|$ (``not est.'': open).
Two 7B rows, marked in the figure, are post hoc (\Cref{app:contrasts}).
\else
Left: A $-$ B, the accuracy of $A$ frozen minus that of $B$ frozen on the same band, in accuracy points, for the dominant band
(circles) and the minor band (squares); positive means freezing $A$ is better.
Bars are the intervals of the pre-specified tests, at each test's own level (97.5\,\% for most; \Cref{tab:tests} gives each).
A filled marker is established: its interval excludes zero.
A hollow marker is not established.
A pale marker is descriptive, not a test: the dominant band on ARC-Challenge at 7B, which its protocol declared too weak to test.
\dag: post hoc, decided after the 7B results on OpenBookQA were read (\Cref{sec:prespec}).
Right: the factor effect $F$ (mean A $-$ B over the two bands) and the band effect $S$ (mean gain of the minor band over the dominant
band; $S > 0$ means the minor band is better), with $95\,\%$ descriptive intervals.
The label at the right edge is the verdict of the pre-specified test of $F$ against $|S|$: ``$F > |S|$'' for the verdict factor,
``not est.'' for the verdict open, and ``no test'' when the experiment has no such test.
For the eight-task mixture the experiment computed only the one-sided bounds this test needs, not intervals for $F$ and $S$, so only
the label is shown.
Eight seeds per configuration.\fi}
\label{fig:forest}
\end{figure}
\endblock
\block{arcvii}
\paragraph{ARC-Challenge with Qwen2.5-7B, dominant band.}
The experiment judged in advance that it had too little statistical power for a test on this band, and declared the comparison
descriptive before the runs.
A $-$ B is $+2.89$ points, with a $95\,\%$ interval of $[+0.66, +5.12]$ (\Cref{fig:forest}, hollow marker: no test).
\endblock
\block{counts}
\paragraph{The parameter counts of \Cref{tab:g1} are close but not equal.}
Relative to $B$ frozen, the matched free LoRA trains $0.9921$ times the parameters on OpenBookQA with Qwen2.5-1.5B and $0.9747$ times on
Qwen2.5-7B.
On the format task it trains $1.0750$ times the parameters at rank 16 and $1.0078$ times at rank 32.
In these rows, the matched free LoRA therefore has slightly more parameters on the format task and slightly fewer on OpenBookQA; across all experiments the ratio lies between $0.9747$ and $1.1338$.
\endblock
\block{robustsum}
\paragraph{What the rest of the robustness record holds.}
The supplementary material (Section~S2) gives each of the following in full, with its intervals and counts.
\paragraph{Module controls.}
On the square modules (query and output projections), where both frozen configurations train the same number of parameters, A $-$ B is $+7.93$ points on OpenBookQA, established positive; on the format task $B$ frozen collapsed, so the test was not computed (their mean accuracy fell below the 45\,\% threshold that the protocol fixes for a collapse).
The cost of a frozen $B$ is not additive across module groups: the module-group interaction is $-6.60$ points on OpenBookQA ($[-10.00, -3.27]$, descriptive) and $-14.07$ on the format task ($[-17.45, -10.53]$).
\paragraph{Four times the updates, and the training length.}\label{app:stepcontrol}
With an effective batch of 8 instead of 32 (about 196 steps on the same tokens), A $-$ B is $+6.27$ $[+2.36, +10.32]$ on the dominant band and $+10.98$ $[+6.62, +15.50]$ on the minor band, both established (level 98.75\,\%).
With four times the training tokens, $A$ frozen on the dominant band peaks on validation at $189.8$ correct answers of 400 at step 39 and ends at $167.5$; $B$ frozen ends at $167.9$.
On the test set, A $-$ B on the dominant band is $+7.05$, $+3.95$, $+0.95$ and $-1.75$ at steps 39, 59, 98 and 196, and on the minor band $+11.12$, $+9.05$, $+6.60$ and $+3.52$ (post hoc, descriptive).
Scored at the best checkpoint, ARC keeps the effect in one test of four.
\paragraph{A stronger free LoRA, and a predictor that points the wrong way.}
A free LoRA trained with LoRA+ beats $A$ frozen on the dominant band of the format task by $3.01$ points ($[+0.06, +6.07]$, established, suggestive).
The predictor derived from \citet{zhu2024asymmetry} has a rank correlation of $-0.362$ with the observed A $-$ B over 40 configurations, with an interval below zero: the prediction is contradicted (post hoc).
\paragraph{The letter prompt.}\label{app:letter}
Scored under the letter prompt, a format the adapters were not trained for, A $-$ B is negative in 6 of the 12 comparisons of six ranks (2 to 64) and two bands (descriptive, post hoc).
The six are the two bands at rank 2, the minor band at ranks 4 and 8, the dominant band at rank 16 and the minor band at rank 64.
The base model answers $76.6\,\%$ with this prompt.
\paragraph{Forgetting and merging.}
At the selected rates, $A$ frozen forgets more of HellaSwag than $B$ frozen in 4 of 16 tests (established negative; for example $-15.34$ points on the format task at rank 2 on the dominant band) and less in none.
At rank 16 on OpenBookQA (post hoc), the matched free LoRA forgets less than $B$ frozen on the minor band ($-3.16$, $[-5.33, -0.96]$), and $A$ frozen is not established to forget more or less than $B$ frozen ($-1.89$ and $-0.34$ for $B$ minus $A$ on the two bands).
At matched accuracy (format task, dominant band), no difference in HellaSwag forgetting is detected (at most $2.32$ points) and $A$ frozen forgets less on WikiText by $78.32$ log-percent (a measure at matched accuracy and rank 2, not the percentages of \Cref{tab:forgetrk}); on the minor band the match failed ($4.12$ points).
On the minor band, $A$ frozen loses more when adapters are merged, by $8.59$ points at rank 2 and $4.28$ at rank 32: A $-$ B is $-8.59$ $[-11.21, -5.94]$ and $-4.28$ $[-6.35, -2.22]$, both established negative.
The within-stratum correlation between intruder-dimension count and forgetting, a test we call $I_1$ (unrelated to the interaction $I$ of \Cref{app:mech}), is not established on the three pairs where it can be computed ($-0.039$, $-0.061$, $-0.618$).
\endblock
\block{forgetrktab}
\begin{table}[!htbp]
\centering
\small
\caption{\textbf{Forgetting at rank 16 on OpenBookQA with Qwen2.5-1.5B (descriptive, 95\,\% intervals).} The runs are reruns, with saved adapters, of the configurations of the first rank-16 experiment, seeds 2 to 9, so their accuracies differ slightly from the reported ones ($55.75$ against $56.00$ for $A$ frozen on the dominant band); \Cref{tab:gap} uses the replication. Test accuracy (\%) and
forgetting: points for HellaSwag and TruthfulQA, increase of word perplexity on WikiText in \%; a positive value is more forgetting. HellaSwag and TruthfulQA are two benchmarks the adapters were not trained on.}
\label{tab:forgetrk}
\begin{adjustbox}{max width=\linewidth}
\setlength{\tabcolsep}{4pt}
\begin{tabular}{@{}lrrrr@{}}
\toprule
Configuration & Accuracy & HellaSwag & TruthfulQA & WikiText \\
\midrule
$B$ frozen, dominant band & $54.10$ & $+0.26$ $[-1.92, +2.42]$ & $+7.77$ $[+5.14, +10.45]$ & $+25.91$ $[+24.27, +27.51]$ \\
$B$ frozen, minor band & $50.40$ & $+2.30$ $[+0.26, +4.35]$ & $+4.94$ $[+2.48, +7.28]$ & $+33.41$ $[+31.30, +35.58]$ \\
$A$ frozen, dominant band & $55.75$ & $+2.15$ $[-0.29, +4.47]$ & $+7.07$ $[+4.36, +9.79]$ & $+131.86$ $[+121.62, +142.93]$ \\
$A$ frozen, minor band & $54.37$ & $+2.64$ $[+0.19, +5.09]$ & $+6.82$ $[+3.79, +9.85]$ & $+234.16$ $[+215.63, +254.05]$ \\
Matched free LoRA, rank 7 & $56.20$ & $-0.86$ $[-3.05, +1.20]$ & $+5.51$ $[+2.85, +8.12]$ & $+32.20$ $[+30.19, +34.29]$ \\
\bottomrule
\end{tabular}
\end{adjustbox}
\end{table}
\endblock
\block{ranksum}
\paragraph{What the record of every setting and rank holds.}
The supplementary material (Section~S3) gives each of the following in full.
\Cref{tab:results} has the numbers.
\paragraph{The four experiments.}
On the two main pairs, four experiments cover the ranks.
The first runs ranks 2, 16, 32 and 64 with seeds 2 to 9; it tests A $-$ B on each band and $F$ against $|S|$ at ranks 2, 32 and 64, and at rank 16 only $F$ against $|S|$.
Two more experiments run ranks 4 and 8, and rank 1, with the same tests.
A replication runs ranks 16 and 32 on eight new seeds (12 to 19) with a finer grid of five rates, each $\sqrt{2}$ times the previous one; its tests of A $-$ B and of $F$ against $|S|$ are post hoc, added after ranks 4 and 8 had been read.
\paragraph{Ranks 1 to 8.}
A $-$ B is established on both bands at ranks 1, 2, 4 and 8, and $F$ exceeds $|S|$ at each of these ranks (the format task at rank 1 is post hoc).
At rank 8, A $-$ B is $+7.18$ and $+7.25$ on OpenBookQA.
\paragraph{Other models and tasks, at rank 2.}
A $-$ B is established on both bands of OpenBookQA with both 7B models and with Qwen3-1.7B-Base, on the format task with Qwen2.5-7B (post hoc), and on the minor band of ARC-Challenge with Qwen2.5-7B (post hoc).
With the instruction-tuned Qwen2.5-1.5B it is established on the dominant band only ($+3.97$ on the minor band, at most $8.35$).
On the crossed pairs it is $+14.23$ and $+12.70$ (format task, Qwen2.5-1.5B) and $+8.30$ and $+5.37$ (OpenBookQA, Llama-3.2-1B): its size follows the task more than the model.
On ARC with Qwen2.5-1.5B it is established on the minor band only (at most $1.48$ and $3.63$ on the dominant band), because $B$ frozen already takes $51\,\%$ to $94\,\%$ of the free LoRA's gain over the base model ($+8.25$ on ARC-Easy, $+6.77$ on ARC-Challenge, against $+12.77$ on OpenBookQA).
\paragraph{Ranks 16 and 32.}
At rank 16, A $-$ B on the minor band is established on both main pairs (suggestive on the format task) and on OpenBookQA with Qwen2.5-7B (post hoc): $+3.90$ and $+3.40$ in the replication, $+6.27$ with Qwen2.5-7B.
On the dominant band it is not established on OpenBookQA (at most $5.42$); on the format task one diverged run of $B$ frozen drives it.
At rank 32 on OpenBookQA, the replication (post hoc) has $|S|$ above $F$: $F + S = -4.17$ ($[-8.00, -0.39]$, level $98.75\,\%$), although neither A $-$ B is established.
The two rank-32 experiments disagree on the dominant band of the format task.
The post hoc points of \Cref{fig:rank} are the format task at rank 1, the replication at ranks 16 and 32, Qwen2.5-7B at rank 16, the mixture at ranks 16 and 32 and Mistral-7B at rank 64.
\paragraph{Mistral-7B, and the eight-task mixture.}
With Mistral-7B, A $-$ B is established on both bands at rank 32 and on the minor band at rank 64 ($+4.88$); on the dominant band at rank 64 it is not established (at most $3.42$), and that configuration chose its rate at the edge of the grid, on an unresolved tie.
Three configurations at rank 32 and four at rank 64 chose the lowest rate of the grid, and the extra step below did worse.
In the mixture the base model scores $58.44\,\%$ and the free LoRA $63.93\,\%$ at rank 2; A $-$ B is established on both bands up to rank 16 and on the minor band at rank 32, with gains of $+0.73$ to $+2.19$ (the values $+0.73$ and $+0.84$ are suggestive; a suggestive result has an interval that ends within 0.5 points of zero at its own level; $+0.84$ is Holm-robust and $+0.73$ is not).
Inside it, A $-$ B on the dominant band is $-1.78$ ($[-3.37, -0.19]$) on ARC-Easy, and on OpenBookQA it is $+1.73$ (at most $5.52$) and $+3.25$ (at most $7.06$), not established.
We did not test the interaction between rank and factor.
\endblock
\block{notestsum}
Where A $-$ B is not established, the largest effect its interval allows is at most $5.42$ and $5.47$ (rank 16, dominant band, Qwen2.5-1.5B and 7B); at most $4.67$ and $6.51$ (rank 32, OpenBookQA, first experiment) and $4.40$ and $4.80$ (replication); at most $3.26$ and $4.45$ (rank 32, format task, first experiment); at most $6.05$ (rank 32, format task, minor band); at most $3.08$ and $5.97$ (rank 32, Qwen2.5-7B); at most $4.80$, $4.34$ and $3.42$ (rank 64: OpenBookQA and the format task on either band, Mistral-7B on the dominant band).
The smallest rank at which A $-$ B is not established is 16 on the dominant band and 32 on the minor band of OpenBookQA, and 32 on the minor band of the format task (a post hoc rule).
The interval can allow a real effect in all these cases: with 500 or 614 test items, effects of a few points are often too small to detect.
The supplementary material (Section~S4) lists them with the rules that name each rank.
\endblock
\block{ocsum}
\paragraph{Rank 2 on OpenBookQA, trained as a generation task.}
\label{app:oc}
The answer-only generation task is the first level (L1) of a ladder of four targets for the same OpenBookQA questions with Qwen2.5-1.5B at rank 2: the answer letter (L1), the letter and the answer text (L2), a fixed JSON object (L3), and the same object with the supporting fact (L4).
Every level trains for 1,187 optimiser steps (about 7.7 epochs of the 4,957 training questions, with one question per sequence, where the main protocol packs several rows into each sequence), not 49, and generation is scored with a parser that extracts the letter.
This differs from the main protocol, so the results are descriptive and not pre-specified tests of A $-$ B.
\Cref{tab:oclevels} gives A $-$ B per band and level, with $F$ and $S$ computed from the mean accuracies.
$F$ is about equal to $|S|$ at L1, below it at L2 and above it at L3 and L4.
A $-$ B is positive in 7 of 8 combinations of band and level, and the 95\,\% interval excludes zero in 4 (the minor band at every level).
Levels L1 and L3 were also trained for 74 and 297 steps, with tests fixed in advance (one family of 12).
At 74 steps, A $-$ B is established on both bands at both levels: on the dominant band $+3.48$ $[+0.02, +7.53]$ at L1 and $+5.95$ $[+2.47, +9.77]$ at L3, and on the minor band $+3.05$ and $+4.50$.
At 297 steps it is established on the minor band at both levels ($+4.35$ and $+3.20$), and not on the dominant band ($+2.75$ and $+2.90$, at most $5.90$ and $5.95$).
The change of A $-$ B between 74 and 1,187 steps is established at L3 on the dominant band only: A $-$ B falls by $4.55$ points $[+0.32, +9.03]$.
\endblock
\block{datasum}
\paragraph{Data, training and models in short.}
The supplementary material (Section~S6) gives each item in full.
Besides the two main pairs, the pairs are ARC-Easy and ARC-Challenge with Qwen2.5-1.5B, OpenBookQA with the instruction-tuned Qwen2.5-1.5B, with Qwen3-1.7B-Base (rank 2) and with Mistral-7B, OpenBookQA, ARC-Challenge and the format task with Qwen2.5-7B, the two main tasks with their models swapped, OpenBookQA and the format task with Llama-3.1-8B, GSM8K with Llama-3.2-1B and Llama-3.1-8B (and with Qwen2.5-1.5B-Instruct in one reinforcement-learning run), and an eight-task mixture (BoolQ, PIQA, SIQA, HellaSwag, WinoGrande, ARC-Easy, ARC-Challenge, OpenBookQA) scored by its mean accuracy.
The test sets have 500 questions (OpenBookQA), 2376 (ARC-Easy), 1172 (ARC-Challenge) and 614 items (format task).
Every run of the main design sees about 600,000 supervised tokens ($602{,}112$ on OpenBookQA and $601{,}469$ on the format task). The two counts follow one rule on different data: every token of a packed block is supervised on OpenBookQA, and only the answer, about 69 tokens per sequence, on the format task. The effective batch is 32 sequences, which gives 49 optimiser steps on OpenBookQA (about 5.8 passes over its 4,957 training rows) and 274 on the format task (3,178 rows).
The learning rate is selected on 400 validation items per selection seed (299 for ARC-Challenge with Qwen2.5-7B; 3,099 for the mixture).
Runs use AdamW with $\beta = (0.9, 0.999)$, $\epsilon = 10^{-8}$, no weight decay, gradient clipping at 1.0, base weights in bfloat16, a linear warm-up over the first 5\,\% of the steps (2 of 49, 13 of 274) and a cosine decay; sequences are packed and cut at 384 tokens.
The experiments ran on NVIDIA RTX 4000 Ada cards (all 213 runs of Mistral-7B); the format task with Qwen2.5-7B and the Llama-3.1-8B runs ran on RTX A5000 cards, so the card model is confounded with the task or the model.
Across the experiments, the matched free LoRA trains between $0.9747$ and $1.1338$ times the parameters of $B$ frozen, and $A$ frozen between $1.0500$ and $1.2777$ times those of the free LoRA.
\endblock
\block{experimentsptr}
Each experiment of Table~S1 that the text does not describe in full is described in Section~S1 of the supplementary material (the ranks 2 to 64 and the replication, the crossed pairs, the mixture, factor against band on three pairs, the oracle and activation experiments, factor stitching, core-only adapters, GaLore, square modules, random bases, LoRA-FA, the controls on the step size and the training length, training-data size, rotation inside the band, module type, output complexity, the least-squares predictor and the side effects).
\endblock
\block{mechsum}
\paragraph{The mechanism experiments in short.}
The supplementary material (Section~S5) gives each in full.
\emph{Unfreezing a factor.} The cost of a frozen factor is the accuracy gained by also training it from the same start. The interaction $I$, the cost of freezing $B$ minus the cost of freezing $A$, is established positive at rank 2 on both main pairs and bands: $+10.02$ and $+10.85$ points on OpenBookQA (dominant and minor band), $+13.21$ and $+20.64$ on the format task; $D$, the gap between the two starts once both factors are trained, is not established on any of the four (at most $4.40$).
On OpenBookQA, unfreezing $B$ on the dominant band gains $+11.40$ points and unfreezing $A$ gains $+1.38$ (descriptive).
\emph{Overlap.} A basis orthogonal to the learned directions recovers $0.182$ of the gap between $B$ frozen on the dominant band and the free LoRA (interval includes zero); a squared cosine of one quarter recovers $0.910$ on OpenBookQA and $0.865$ on the format task.
The rank correlations are $+0.902$ and $+0.753$ at rank 2, and $+0.524$ and $+0.451$ at rank 8 (post hoc).
\emph{Input side.} $A$ frozen on the learned input directions beats $A$ frozen on the dominant band by $+4.82$ points on the format task (established) and shows no difference on OpenBookQA.
With each factor frozen on its own learned directions, no difference is detected (at most $2.40$ on OpenBookQA and $5.05$ on the format task).
\emph{Core-only adapters.} With the learned directions on both sides, the core alone reaches $81.03\,\%$ against $81.37\,\%$ for the free LoRA on the format task and $54.80\,\%$ against $56.20\,\%$ on OpenBookQA; the spectral placement beats random bases on OpenBookQA by $+2.98$ (suggestive; at most $0.96$ on the format task); the bases of FCCA~\citep{ye2026fcca} are not detectably better than the spectral ones (at most $5.67$ and $7.66$).
\endblock
\block{gradsum}
The input side of the leading gradient subspace is the more reproducible across disjoint minibatches in 98 of 126 settings on OpenBookQA and 100 of 126 on the format task (the output side in 28 and 26).
Between step 0 and the last step of a free LoRA at rank 2, the overlap above chance is $0.108$ (output) and $0.415$ (input) on OpenBookQA, $0.059$ and $0.190$ on the format task; the output side is the lower in $0.88$ of 196 and $0.85$ of 112 modules.
Two halves of the batch at step 0 give $0.832$ and $0.946$ on OpenBookQA.
At rank 32 the sides are $0.199$ and $0.378$ on OpenBookQA, $0.121$ and $0.148$ on the format task.
These measurements are descriptive, rest on one run per task and do not test the hypothesis of \Cref{sec:discussion}; Section~S5 of the supplementary material gives the analysis.
\endblock
\block{methodsum}
\paragraph{MiCA and PiCa in short.}
The supplementary material (Section~S7) gives each item in full.
Four values of \Cref{tab:gap} also appear as A $-$ B in \Cref{tab:results}. $+4.78$ (format task) is a coincidence, since it comes from another experiment (rank 8). $+6.27$ (Qwen2.5-7B, rank 16) recurs because $A$ frozen on the minor band and the free LoRA have the same mean there. $+3.90$ and $+1.40$ recur too, each time in a different setting.
Other values recur across experiments: $78.32$, $54.37$, $+7.93$, $+9.62$, $+10.32$ and $+11.07$ each come from two distinct readings, and we checked each pair against its logs.
Further repeats, such as $+8.30$, $+4.82$, $+3.48$, $+4.38$ and $+8.70$, also belong to different experiments.
MiCA's configuration is our $B$ frozen on the minor band, on our seven adapted modules rather than the query and value projections of its paper.
At MiCA's recipe our code and the PEFT implementation differ by $-0.23$ points on OpenBookQA and $-0.77$ on the format task (descriptive); both run without adapter dropout.
In \Cref{tab:methods}, the second flipped row runs at rank 13 on OpenBookQA and 14 on the format task ($1.030$ and $1.006$ times MiCA's parameters); the matched free LoRA runs at rank 7 and 8 ($0.992$ and $1.075$ times).
Flipped MiCA beats MiCA on the format task, but the matched free LoRA stays above it; with tuned rates in the PEFT library the matched free LoRA leads MiCA on OpenBookQA at rank 16, and at MiCA's recipe on both tasks (descriptive).
PiCa writes its update as $U_r B$ with $U_r$ frozen on the singular vectors and $B$ trained and shared across layers; its parameter counts imply that $U_r$ spans the input side, so PiCa freezes $A$ on the dominant band.
At rank 256 it trains $5{,}898{,}240$ parameters on OpenBookQA and $6{,}029{,}312$ on the format task, in 15 and 150 optimiser steps.
At its published rate, Part A (primary) is in \Cref{tab:tests}.
With rates selected on separate seeds (Part B, secondary, level $99.375\,\%$), the matched free LoRA minus PiCa is $-1.20$ $[-3.73, +1.18]$ on OpenBookQA at rank 256, $+0.10$ $[-2.57, +2.98]$ on the format task at rank 256 and $-0.16$ $[-3.63, +3.28]$ at rank 16; PiCa minus $B$ frozen on the dominant band is $-0.42$ $[-3.52, +2.78]$, $-0.61$ $[-3.60, +2.36]$ and $+3.14$ $[-0.22, +6.53]$; none is established.
At PiCa's parameter count, on the dominant band, $A$ frozen beats $B$ frozen on the format task at rank 16 by $+6.78$ points at the published rate, but not at selected rates, so under the weaker-result rule it is not established; in every other case no difference is detected.
We read PiCa's gain at its published rate as under-training of the free LoRA, not as evidence about frozen factors.
\endblock
\block{deviationssum}
\begin{table}[!htbp]
\centering
\small
\caption{\textbf{Deviations that bear on a result in the body.} Every decision taken after related results had been read is marked post hoc\posthoc\ in the text; the departures from a protocol are stated where the result is reported. The full registry, one row per deviation, is in Table~S2 of the supplementary material.}
\label{tab:deviations}
\begin{tabular}{>{\raggedright\arraybackslash}p{3.4cm}>{\raggedright\arraybackslash}p{10.6cm}}
\toprule
Kind & Experiments \\
\midrule
Decided after related results were read; tests fixed before the runs & the replication at ranks 16 and 32 (after ranks 4 and 8); Qwen2.5-7B at rank 16, ARC-Challenge and the format task (after the other 7B results); output overlap at ranks 8 and 16 (after rank 2); scaling $\alpha$ and unfreezing a factor at rank 16 (after the rank-2 controls); GSM8K with Llama-3.1-8B (after GSM8K with Llama-3.2-1B); flipped MiCA (after the rank-16 results); forgetting at matched accuracy (after the forgetting and merging results); the eight-task mixture at ranks 16 and 32 (after ranks 2 and 8); Mistral-7B at rank 64 (after ranks 2 and 32). \\
Fixed after the results they predict or compare had been read & the least-squares predictor; every learning rate of $B$ frozen; factor against band on three pairs; forgetting at rank 16; the format task at rank 1 (nine runs of $B$ frozen existed and had been read). \\
Departures from a protocol & Mistral-7B at ranks 32 and 64 (lowest rate of the grid; one unresolved tie); the diverged run at rank 16 on the format task (kept); no adapter dropout in flipped MiCA;  square modules on the format task ($B$ frozen collapsed); a different GPU model for Qwen2.5-7B on the format task; two of eight runs of $A$ frozen on the activation-null directions collapsed ($16.4$ and $20.4\,\%$); under SGD the free LoRA collapsed on 5 of 8 seeds (mean $35.10\,\%$); the match at equal accuracy failed on the minor band ($4.12$ points); resamples drawing the same seed twice in the overlap interval; the order of the intruder-dimension analysis; the relation of flipped MiCA to LoRA-RLMO. \\
\bottomrule
\end{tabular}
\end{table}
\endblock
\block{controlsmore}
\begin{scriptsize}
\begin{longtable}{p{3.3cm}p{3.0cm}p{5.0cm}p{3.0cm}}
\caption{\textbf{Further descriptive controls.} The rows of the controls table of the paper that are not summarised there. Same conventions as that table.}
\label{tab:controlsmore}\\
\toprule
Control & Addresses & Result & Status \\
\midrule
\endfirsthead
\toprule
Control & Addresses & Result & Status \\
\midrule
\endhead
Loss on the answer\posthoc & a scoring artefact & lower loss for $A$ frozen at every rank (format task) & descriptive \\
Transfer across tasks\posthoc & a side effect on neighbouring tasks & \Cref{sec:discussion} and \Cref{app:transfer} & descriptive \\
Removing modules at evaluation\posthoc & where the gain sits & on OpenBookQA, $B$ frozen (dominant) gains mostly in the MLP down projection, $A$ frozen (dominant) in the gate projection; $A$ frozen (minor) also leans on the down projection (4 seeds) & descriptive \\
Energy at the base model\posthoc & whether the share of input or output-gradient energy held by a frozen basis orders the accuracy of the arms that freeze that factor & Spearman correlation $+0.105$ and $+0.101$ on OpenBookQA (frozen $A$, frozen $B$), $+0.219$ and $+0.098$ on the format task & not established, no claim made \\
Rank and band over all comparisons\posthoc & how A $-$ B changes with the rank & $-1.44$ points per doubling of the rank $[-1.81, -1.07]$ over $104$ comparisons (mixed model with a random intercept per experiment); the minor band is $+0.90$ $[-0.12, +1.93]$ above the dominant band & descriptive, rough intervals \\
Repeated runs at rank 16 & whether two runs of one configuration agree & on OpenBookQA, two repeated runs of $B$ frozen on the dominant band differ on $28$ of 500 test predictions, and two of $A$ frozen on $7$; four runs of one $B$-frozen configuration differ in score by up to $1.6$ points & descriptive \\
GSM8K, Llama-3.2-1B, ranks 2 to 32 & a generative reasoning task & A $-$ B on both bands at ranks 2 and 8, the factor larger at rank 2 (\Cref{tab:gsm}) & established; ranks 16 and 32 descriptive \\
Controls repeated at rank 16 & whether the controls hold at MiCA's rank & the LoRA scale, the decomposition and the overlap (\Cref{app:rankexp,app:overlap}); longer training (\Cref{app:cvr16}) & minor band mostly established \\
\bottomrule
\end{longtable}
\end{scriptsize}
\endblock
\block{testsshort}
Table~S1 of the supplementary material lists every pre-specified test that has a bootstrap interval (284 rows), with the paired $t$ interval.
\Cref{tab:tests} repeats the rows that the claims rely on, with the row numbers of Table~S1: A $-$ B at rank 2 and $F$ against $|S|$ for the two main pairs at ranks 2, 16, 32 and 64, PiCa at rank 256 and the output-overlap correlations.
A $-$ B at the other ranks, and the replication, are in \Cref{tab:contrasts}.
Rows are unmarked when primary and marked $^\circ$ when their protocol labels them neither primary nor secondary; \dag\ marks a post hoc test (\Cref{sec:prespec}).
The level is $1 - 0.05/m$ for a family of $m$ tests, and the status is the one given by the experiment's own family.
For a rule test, est.\,$+$ is the outcome the rule names and est.\,$-$ the opposite; for $F$ against $|S|$, est.\,$+$ is the verdict factor, n.e. the verdict open and est.\,$-$ the verdict band.
\endblock
\block{testsintro}
\Cref{tab:tests} lists the pre-specified tests that have a bootstrap interval, one per row, grouped by the section that uses them.
Unmarked rows are primary: their protocol calls them primary, or they form its single family and are not called secondary.
Rows marked $^\circ$ are pre-specified but neither primary nor secondary in their protocol.
The text cites several, for example A $-$ B in the replication at ranks 16 and 32.
Secondary tests are not listed; the text gives their level where it quotes them.
The tests of intruder dimensions and the post hoc tests of forgetting at rank 16 are in \Cref{app:robust}.

Each row gives the contrast, its estimate and interval, the level fixed in advance ($1 - 0.05/m$ for a family of $m$ tests) and the status in the experiment's own family.
For a rule test, est.\,$+$ is the outcome the rule names and est.\,$-$ the opposite.
For $F$ against $|S|$, est.\,$+$ is the verdict factor, n.e. the verdict open and est.\,$-$ the verdict band, whose two components are at level 98.75\,\%.
A row marked \dag\ is post hoc: its test was written before its runs, but we decided to run it after related results had been read (\Cref{sec:prespec}).
\endblock
\block{experiments}
The experiments of \Cref{tab:tests} that the text does not describe in full are listed here, in the order of the table.
\begin{description}[leftmargin=0pt, itemsep=2pt]
\item[Ranks 2 to 64, and the replication.] \Cref{sec:rank} describes them. ``49 steps'' is the standard budget on OpenBookQA: 49 optimiser steps, which is 600,000 supervised tokens and about 5.8 passes over the packed training set. ``4 times more steps'' trains on the same 600,000 tokens with batches four times smaller.
\item[Task and model crossed.] The two main pairs mix task and model.
  We run the format task on Qwen2.5-1.5B and OpenBookQA on Llama-3.2-1B at rank 2, to see whether the advantage of $A$ frozen depends on one task--model combination.
\item[Eight-task mixture, cross-task tests.] On seven of the eight tasks (one is below the floor), we ask whether the share of the free LoRA's gain that $B$ frozen keeps follows a property of the task, measured in advance.
  The first property is the base model's margin between the correct and the best wrong answer.
  The second is the share of the update in the output directions of the answer tokens.
\item[Factor against band, three pairs together\posthoc.] On the three pairs that played no part in forming the hypothesis (the format task, ARC-Easy, ARC-Challenge), we test $F$ against $|S|$ at rank 2 and at rank 64.
  The test uses the mean of each quantity over the pairs, not over the ranks.
  The hypothesis came from OpenBookQA.
  The analysis code was written after one of the three pairs had been read, so the test is post hoc.
\item[Flipped MiCA, and MiCA and LoRA in PEFT.] \Cref{sec:methods} describes them.
\item[Output overlap and oracle input factor.] \Cref{sec:mechanism} describes them. The oracle freezes $A$ on the two leading input directions of the free LoRA's mean update.
  We compare it with $A$ frozen on the dominant band and with the oracle that freezes $B$ on the two leading output directions.
\item[Unfreezing a factor.] \Cref{app:mech} describes it and defines the interaction $I$.
\item[$A$ frozen on activation bands.] We freeze $A$ on the input directions that the module's inputs use most (activation-dominant band) or least (activation-null band).
  These are the eigenvectors of the second moment of the module's inputs in the base model, with the largest or smallest eigenvalues.
  The question is whether freezing $A$ can be made as costly as freezing $B$.
\item[Factor stitching.] We ask whether the input factor that $B$ frozen learns is a usable half of a LoRA solution.
  We join the output factor learned with $A$ frozen and the input factor learned with $B$ frozen through a small trained $r \times r$ matrix.
  We compare this stitched adapter with the $A$-frozen adapter given the same matrix, with a random input projection, and with a free LoRA of rank 1.
  The stitched adapter is established negative against the $A$-frozen adapter on both pairs and both bands.
  On the format task, dominant band, the difference is $-46.64$ points $[-51.61, -41.73]$.
\item[Core-only adapters.] Both outer factors are frozen and only a $16 \times 16$ core between them is trained, as in LoRA-XS~\citep{balazy2024loraxs} and FCCA~\citep{ye2026fcca}.
  We ask which side benefits more from the directions a free LoRA learns, whether spectral bases beat random ones, and whether the bases of FCCA beat the dominant spectral bases.
\item[GaLore projection side.] GaLore~\citep{zhao2024galore} trains all weights but keeps its optimiser state in a low-rank projection of each gradient, on one side of each matrix.
  We compare projecting every matrix on its input side, on its output side, and on the side that GaLore's code picks from the matrix shape.
\item[Square modules.] \Cref{app:robust} gives the result. The minor band of these two modules holds singular values below the resolution of 16-bit numbers, so random orthonormal bases take its place.
  Then $R$, the analogue of $S$, compares a random basis with the dominant band.
\item[Random bases.] On the format task, $A$ or $B$ is frozen on 24 random orthonormal bases each.
  Rank correlations ask whether $q$ tracks accuracy, where $q$ is the share of the target update that a basis captures in a least-squares fit.
\item[LoRA-FA initialisation.] $A$ is frozen at the random initialisation of LoRA-FA~\citep{zhang2023lorafa}, the free LoRA's initial $A$.
  We compare it with $B$ frozen and with the same rows made orthonormal.
\item[$\alpha/r$ sweep, same step size, SGD, LoRA+.] \Cref{app:controls} and \Cref{app:robust} give their results.
  For the same step size, the frozen $A$ of each module is multiplied by $\sqrt{d_{\text{in}}/d_{\text{out}}}$.
  Then a sign-like AdamW step moves the update by the same amount as with $B$ frozen.
  The rescaling does not match steps where AdamW's normalisation is not saturated, nor its weight decay.
\item[Four times the training budget, and ARC at its best checkpoint.] The best checkpoint is chosen on validation accuracy among ten checkpoints. The test set is scored only there.
\item[Every learning rate of $B$ frozen\posthoc.] $A$ frozen at its selected rate is compared with $B$ frozen at every rate of $B$'s grid, including the rate that is best on the test set.
  The result is established only if every difference is.
  It answers the objection that $B$ frozen was tuned badly.
  It is post hoc because the test scores of $B$ frozen at its selected rate had been read.
\item[Training-data size.] One eighth or one half of the training examples, with the same number of optimiser steps.
\item[Rotation inside the band.] AdamW treats each coordinate separately, so it is not invariant to a rotation of the frozen basis inside its band.
  We apply a random rotation that keeps the subspace, to see whether the coordinates matter.
\item[Module type.] We freeze $A$ or $B$ separately in the attention modules and in the MLP modules, to ask where freezing $B$ costs more.
  The parameter counts differ between these settings.
  The costs are not additive (\Cref{app:robust}).
\item[Output complexity.] On the same OpenBookQA questions, the model learns four answer formats, from the answer letter alone to a JSON object with the supporting fact.
  We test whether A $-$ B grows from the simplest to the richest format.
\item[Least-squares predictor\posthoc.] \Cref{app:robust} gives its result.
  The predictor uses the input activations of each module in the base model on 512 training rows.
  Its target is the mean update of the pair's highest-rank free LoRA.
  It was fixed after the observed values of A $-$ B had been read.
\item[Side effects: forgetting, forgetting at matched accuracy\posthoc, adapter merging.] \Cref{sec:robustness} describes them. Forgetting is measured on the first 1,000 HellaSwag items.
\end{description}
\endblock
\block{robustintro}
This appendix details \Cref{sec:robustness}. As there, A $-$ B is the accuracy of $A$ frozen minus that of $B$ frozen on the same band.
\Cref{tab:robustfull} gives the five objections to A $-$ B at rank 2, with their tests and intervals.
Then come two controls (parameter count, module groups) and the results on the training budget.
The appendix ends with the other results against the advantage of $A$ frozen and the side effects: forgetting, intruder dimensions and merging.
\Cref{app:controls} lists every control in one table.
\input{tables/robustness_full_appendix}
\endblock
\block{square}
\paragraph{The extra parameters of \texorpdfstring{$A$}{A} frozen do not explain the advantage on OpenBookQA.}
At the same rank, $A$ frozen trains more parameters than $B$ frozen (\Cref{sec:budgets}).
To remove this difference, we adapt only the query and output projections of attention.
Their input and output dimensions are equal, so both frozen configurations train exactly the same number of parameters in the same shape.
On OpenBookQA, A $-$ B on these square modules is $+7.93$ points, established positive.
On the format task, the runs with $B$ frozen collapsed on these modules: their mean accuracy fell below the 45\,\% threshold that the
protocol fixes for a collapse.
The test could therefore not be computed.
This control uses its own seeds (selection on seeds 145 and 146, eight reported seeds from 150 to 157).
\endblock
\block{modulegroups}
\paragraph{The cost of a frozen \texorpdfstring{$B$}{B} is not additive across module groups.}
At rank 2 on the dominant band, we freeze $A$ or $B$ separately in the attention modules and in the MLP modules.
The design has four configurations, each with one frozen factor in attention and one in the MLP.
In attention, we compute $A$ frozen minus $B$ frozen twice: once when the MLP has $A$ frozen, and once when the MLP has $B$ frozen.
The \emph{module-group interaction} is the first of these differences minus the second.
It is $-6.60$ points on OpenBookQA (descriptive, 95\,\% interval $[-10.00, -3.27]$) and $-14.07$ on the format task ($[-17.45, -10.53]$).
The experiment tested the cost in each group and the difference of the two costs, not the module-group interaction.
When the MLP has $A$ frozen, the difference in attention is small: $-0.52$ and $-1.69$ points on OpenBookQA and the format task.
When the MLP has $B$ frozen, it is large: $+6.07$ and $+12.38$ points.
The cost of freezing $B$ in attention and the cost of freezing $B$ in the MLP therefore overlap instead of adding.
\endblock
\block{setup}
\paragraph{The training setup.}
We changed one part of the training at a time.
Each change addresses one way in which the training, rather than the frozen factor, could produce the advantage of $A$ frozen.
The parts are:
the scale of the update ($\alpha/r$, including the scale of rsLoRA~\citep{kalajdzievski2023rslora});
the optimiser (SGD instead of AdamW~\citep{loshchilov2019adamw});
the step size per module;
the number of updates on the same tokens;
the frozen basis (random instead of spectral);
and the amount of training data.
At rank 2, every comparison of $A$ frozen with $B$ frozen stays established positive except one: one eighth of the data on the dominant band of OpenBookQA, which is not established.
\Cref{app:controls} gives each control, what it addresses and its values.
Only the LoRA-FA control reuses the $B$-frozen runs of the main experiments.
The other controls train their own.
\endblock
\block{longer}
\paragraph{Longer training narrows the gap on OpenBookQA.}
\iftmlr\else With four times the training budget (four times the training tokens), A $-$ B on OpenBookQA is no longer established on the dominant band.
It is $+3.62$ points there (not established, at most $7.35$ points), against $+8.45$ on the minor band (established).
Each run is scored at its best checkpoint, chosen on validation accuracy. \fi
Against the standard number of training steps, four times the training budget changes A $-$ B on the dominant band by $-4.98$ points (descriptive, 95\,\% interval $[-8.05, -1.95]$) and on the minor band by $-1.65$ points (descriptive, $[-5.57, +2.22]$).
\iftmlr\else With four times the updates at the same number of tokens, A $-$ B stays established on both bands (\Cref{app:stepcontrol}). \fi

\paragraph{Validation curves.}
The validation values are means over seeds 2 to 9, in correct answers out of 400.
$A$ frozen on the dominant band peaks at $189.8$ at step 39 and ends at $167.5$ at step 196.
$B$ frozen peaks at $171.0$ at step 98 and ends at $167.9$.
The selected checkpoints are early for $A$ frozen on the dominant band (median step 39), later for $B$ frozen (137) and for the free LoRA ($127.5$).

\paragraph{Test scores at fixed steps.}
We scored the saved checkpoints of the same runs on the test set (post hoc, descriptive; nothing was retrained).
A $-$ B is:
\begin{center}\small
\begin{tabular}{@{}lrrrr@{}}
\toprule
Step & 39 & 59 & 98 & 196 \\
\midrule
Dominant band & $+7.05$ $[+3.63, +10.53]$ & $+3.95$ $[+0.45, +7.42]$ & $+0.95$ $[-2.60, +4.50]$ & $-1.75$ $[-5.22, +1.88]$ \\
Minor band & $+11.12$ $[+7.53, +14.77]$ & $+9.05$ $[+5.35, +12.80]$ & $+6.60$ $[+2.73, +10.55]$ & $+3.52$ $[-0.23, +7.38]$ \\
\bottomrule
\end{tabular}
\end{center}
\noindent From the runs at the standard 49 steps to step 196, A $-$ B changes by $-10.35$ $[-14.13, -6.50]$ on the dominant band and $-6.58$ $[-10.52, -2.58]$ on the minor band.
From step 196 to each run's best validation checkpoint it changes by $+5.38$ $[+2.42, +8.42]$ and $+4.93$ $[+1.87, +7.92]$.
The dominant-band configurations use the same learning rates at 49 steps and in these runs.
The minor-band configurations do not, so on that band the change mixes the training budget and the rate.
These runs do not separate the training budget from the position in the schedule, and they use eight seeds.

At the last step of four times the training budget, we do not detect an advantage of $A$ frozen on either band.
The best checkpoint of $A$ frozen on the dominant band is also early.
This result goes against the thesis on that band.
The control changes the number of training steps and the scoring at once.
At the standard 49 optimiser steps (\Cref{app:data}), $B$ frozen stays between $40.85$\,\% and $46.05$\,\% across its whole rate grid, barely above the base model's $42.20$\,\%.

\paragraph{The format task.}
Its base model scores $0$\,\%.
Four times the training budget means 1,096 steps instead of 274.
A $-$ B at the last step is $+11.18$ $[+7.52, +14.80]$ on the dominant band and $+15.96$ $[+12.26, +19.71]$ on the minor band (both established, level 98.75\,\%).
Against 274 steps, A $-$ B changes by $-0.94$ $[-5.47, +3.73]$ and $-4.44$ $[-9.18, +0.13]$ (not established).
\endblock
\block{stepcontrol}
\paragraph{Four times the updates leaves A $-$ B established on both bands.}
\label{app:stepcontrol}
The step control trains OpenBookQA with Qwen2.5-1.5B at rank 2 with an effective batch of 8 instead of 32.
It makes about 196 optimiser steps instead of 49 on the same 600,000 supervised tokens.
A $-$ B is $+6.27$ points on the dominant band (established, $[+2.36, +10.32]$) and $+10.98$ on the minor band (established, $[+6.62, +15.50]$), both at level 98.75\,\%.
The configuration means are $47.23\,\%$ ($B$ frozen, dominant band), $44.92\,\%$ ($B$ frozen, minor band), $53.50\,\%$ ($A$ frozen, dominant band) and $55.90\,\%$ ($A$ frozen, minor band).
The matched free LoRA reaches $54.80\,\%$, and the base model $42.20\,\%$.
From these means (descriptive, no interval), $F = +8.625$, $S = +0.045$, $S_A = +2.40$ and $S_B = -2.31$.
With four times the training budget instead (four times the tokens), $B$ frozen reaches $48.62\,\%$ on the dominant band and $44.02\,\%$ on the minor band at its best checkpoint.
So four times the updates leaves A $-$ B established on both bands, whereas four times the training budget does not establish it on the dominant band.
What remains on OpenBookQA is a dependence on the protocol (\Cref{app:oc}).
\endblock
\block{arcbest}
\paragraph{ARC scored at the best checkpoint.}
When the ARC runs are scored at their best checkpoint instead of the last one, A $-$ B is established in only one of the four tests
(two tasks, two bands).
That test is the minor band of ARC-Easy, and it is suggestive.
\endblock
\block{loraplus}
\paragraph{A stronger free LoRA beats \texorpdfstring{$A$}{A} frozen on the format task.}
LoRA+~\citep{hayou2024loraplus} trains $B$ with a larger learning rate than $A$.
We tune a free LoRA this way over 24 combinations of the ratio and the rate.
The frozen configurations get 6 rates, which favours the free LoRA.
This LoRA+ free LoRA does not differ detectably from the free LoRA we use elsewhere (descriptive).
$A$ frozen on the dominant band is below it by $3.01$ points $[+0.06, +6.07]$ on the format task (established negative, suggestive).
This does not change the comparison of $A$ frozen with $B$ frozen.
\endblock
\block{letterrule}
\paragraph{The letter prompt, rank by rank.}
\label{app:letter}
These results\posthoc\ are descriptive.
The adapters are trained with the likelihood rule on OpenBookQA with Qwen2.5-1.5B.
They are scored under the letter prompt: the answer letter on a prompt with lettered options, a format they were not trained for.
A $-$ B is the number of correct items of $A$ frozen minus those of $B$ frozen, pooled over 8 seeds (4,000 items per configuration).
At rank 2, the dominant band gives $-621$ and the minor band $-1{,}496$.
At rank 4, they give $+1{,}686$ and $-1{,}156$, and at rank 8, $+1{,}404$ and $-851$.
At rank 16, they give $-196$ and $+1{,}167$, and at rank 32, $+823$ and $+315$.
At rank 64, they give $+1{,}108$ and $-107$.
The six negative comparisons are the two bands at rank 2 and the minor band at ranks 4 and 8.
They also include the dominant band at rank 16 and the minor band at rank 64.
The base model answers $76.6\,\%$ with this prompt.
\endblock
\block{predictor}
\paragraph{A predictor we derive from the analysis of Zhu et al. does not order our results.}
\citet{zhu2024asymmetry} analyse a single linear layer fitted by least squares.
From their setting we derive, for each module, the loss that remains when $A$, or $B$, is frozen on a given basis.
This gives, for each configuration, a predicted gap: how much better freezing $A$ should be than freezing $B$ (\Cref{app:experiments}).
The predictor is ours, not theirs.
We fixed the prediction that the predicted gaps and the observed A $-$ B would rise together over 40 configurations: the two main pairs at ranks 2, 16, 32 and 64, and ARC-Easy and ARC-Challenge at ranks 2 to 64, each on both bands.
We wrote it after the observed values had been read, so the test is post hoc\posthoc.
The rank correlation is $-0.362$, with an interval below zero.
The prediction is \emph{contradicted}: it is established in the opposite direction.
The 40 configurations span several ranks, and A $-$ B changes with the rank, so the correlation mixes differences within and between ranks.
The configurations are also not independent, and we did not fix a test within each rank.
This least-squares predictor does not explain the asymmetry we observe in trained language models.
\endblock
\block{forget}
\paragraph{At the selected learning rates, freezing \texorpdfstring{$A$}{A} forgets more of HellaSwag in 4 tests of 16.}
\emph{Forgetting} is the HellaSwag~\citep{zellers2019hellaswag} accuracy of the base model minus that of the adapted model, in points.
A larger value means more forgetting.
We compare forgetting of $B$ frozen minus forgetting of $A$ frozen, so a negative value means that $A$ frozen forgets more.
The 16 tests cover four task--model pairs, ranks 2 and 32, and two bands.
Of these, 4 are established negative, for example $-15.34$ points on the format task at rank 2 on the dominant band.
None is established positive.
The same experiment also compares MiCA's configuration with the matched free LoRA at rank 32.
On the format task, the matched free LoRA forgets less than $B$ frozen on the minor band, by $2.93$ points (established negative, suggestive).
On the three other pairs, no difference is detected.
\endblock
\block{forgetrk}
\paragraph{At rank 16 on OpenBookQA, \texorpdfstring{$A$}{A} frozen is not established to forget more or less than \texorpdfstring{$B$}{B} frozen\posthoc.}
The experiment above tests forgetting at ranks 2 and 32.
Rank 16 is the rank of the methods we compare with (\Cref{sec:methods}), and it was shown only in a descriptive map.
We therefore reanalyse the per-item records of the 40 rank-16 adapters of OpenBookQA with Qwen2.5-1.5B and
of the base model ($58.30\,\%$ on HellaSwag).
No new evaluation is run.
The tests were fixed after the rank-16 values had been seen in that map.
Every statement here is therefore post hoc\posthoc, and holds at the selected learning rates only.
Forgetting and its sign are defined as above.
The bootstrap intervals are at level $97.5\,\%$, two tests per family.
The forgetting of $B$ frozen on the dominant band minus that of $B$ frozen on the minor band is $-2.04$\posthoc\ ($[-4.24, +0.08]$), not
established.
The forgetting of the matched free LoRA (rank 7) minus that of $B$ frozen on the minor band is $-3.16$\posthoc\
($[-5.33, -0.96]$), established negative: the matched free LoRA forgets less.
The forgetting of $B$ frozen minus that of $A$ frozen is $-1.89$\posthoc\ ($[-4.16, +0.39]$) on the dominant band and $-0.34$\posthoc\ ($[-2.82, +2.21]$) on the minor
band, both not established.
The statuses at rank 32 are those of the experiment above and are not recomputed.
\Cref{tab:forgetrk} gives the descriptive map of the three benchmarks at rank 16.

\endblock
\block{matched}
\paragraph{At matched task accuracy, on the one pair and band where the match held, no extra forgetting of \texorpdfstring{$A$}{A} frozen is detected.}
$A$ frozen also learns its task better, and forgetting may grow with how much a model learns.
To separate the two, we lower the learning rate of $A$ frozen until its validation accuracy matches that of $B$ frozen (format task,
dominant band, 16 seeds).
At this rate, no difference in HellaSwag forgetting is detected (at most $2.32$ points).
We also measure forgetting as the increase of word perplexity on 62 WikiText documents~\citep{merity2017wikitext}, in
\emph{log-percent}: $100 \ln(\text{perplexity of the adapted model} / \text{perplexity of the base model})$.
On WikiText, $A$ frozen now forgets less than $B$ frozen, by $78.32$ log-percent (established).
On the minor band the match failed: $A$ frozen stayed $4.12$ points more accurate than $B$ frozen.
Its lower forgetting there ($-3.81$ points of HellaSwag and $-52.56$ log-percent of WikiText) is therefore not a result at equal accuracy.
The estimate at zero accuracy gap is $-5.57$ points of HellaSwag (descriptive).
For WikiText on this pair and band, the extra forgetting of $A$ frozen at the selected learning rates is compatible with an effect of its larger learning rate.
We also compared equal learning rates, in the selection runs of the same pair and band (two seeds, descriptive).
There, $A$ frozen forgets more than $B$ frozen at each of the six rates of the grid.
These results are compatible with the extra forgetting of $A$ frozen coming from how much more it learns, at least on this pair.
A smaller effect of the frozen factor itself cannot be excluded.
\endblock
\block{intruders}
\paragraph{Whether intruder dimensions track forgetting inside a configuration is not established.}
\citet{shuttleworth2024illusion} define an \emph{intruder dimension} as a leading singular vector of the fine-tuned weight whose similarity with every singular vector of the pretrained weight stays below a threshold.
They report that the number of intruder dimensions goes with forgetting, across epochs and learning rates.
We follow their definition on the 480 adapters whose forgetting is measured (OpenBookQA 160, the format task 160, ARC-Easy 80, ARC-Challenge 80; seeds 2 to 9).
We compare each of the 10 leading left singular vectors of the adapted weight with all the left singular vectors of the base weight.
A vector is an intruder when its largest absolute cosine is below $0.5$.
The input side is counted the same way with the right singular vectors, and counts are summed over the modules.

Pooled over configurations and ranks, the count would mostly measure the frozen factor and its selected rate.
The pre-specified test $I_1$ is therefore computed within strata (not the interaction $I$ of \Cref{app:mech}).
A stratum is one configuration at one rank with its eight seeds.
$I_1$ is the mean over the strata of the Spearman correlation between the output-side count and the HellaSwag forgetting.
A stratum where either variable is constant is left out.
Its bootstrap interval, over seeds and items, is at level $98.75\,\%$ (four pairs).

\begin{center}\small
\begin{tabular}{@{}lrrr@{}}
\toprule
 & Strata used & $I_1$ & Interval \\
\midrule
OpenBookQA & 10 of 20 & $-0.039$ & $[-0.439, +0.369]$ \\
Format task & 13 of 20 & $-0.061$ & $[-0.322, +0.347]$ \\
ARC-Challenge & 1 of 10 & $-0.618$ & $[-1.000, +0.850]$ \\
\bottomrule
\end{tabular}
\end{center}
\noindent $I_1$ is not established on the three pairs where it can be computed.
On ARC-Easy, no stratum has both a count and a forgetting that vary across seeds, so $I_1$ cannot be computed.
With eight seeds per stratum and little seed-to-seed variation, $I_1$ could resolve only a strong association.
A test that is not established does not show that there is no association.
The input-side twin of $I_1$ is also not established: $-0.130$ over 10 strata on OpenBookQA ($[-0.368, +0.173]$) and $+0.018$ over 13 on the format task ($[-0.202, +0.317]$).
With the threshold at $0.3$ and at $0.7$ instead of $0.5$, the output-side values are $-0.01$ and $+0.17$ on OpenBookQA and $-0.03$ and $+0.13$ on the format task (point values).

Over all the adapters of a pair, which mixes configurations and ranks, we also report two correlations (descriptive, 95\,\% intervals).
They are associations across configurations, not evidence that intruder dimensions explain the forgetting of the frozen factors.

\begin{center}\small
\begin{tabular}{@{}lrr@{}}
\toprule
 & Spearman, count and forgetting & Partial, controlling for update norm and accuracy \\
\midrule
OpenBookQA & $-0.021$ $[-0.244, +0.183]$ & $-0.348$ $[-0.436, -0.145]$ \\
Format task & $+0.320$ $[+0.152, +0.427]$ & $+0.195$ $[+0.032, +0.318]$ \\
ARC-Challenge & $+0.118$ $[-0.357, +0.443]$ & $-0.049$ $[-0.490, +0.401]$ \\
\bottomrule
\end{tabular}
\end{center}

The geometry fixes some counts.
Below rank 5, a singular vector of the adapted weight that lies inside the frozen band cannot be an intruder at the threshold $0.5$, so those counts are not read as findings.
At ranks 16 to 64 the minor-band adapters have no intruder on either side on OpenBookQA and the format task (mean count $0.00$, as for the null draws), so their strata do not enter $I_1$.
At rank 16 on OpenBookQA, the dominant-band adapters with $B$ frozen have on average $60.38$ output-side intruders (null draw $12.12$) and forget $+0.26$ points.
With $A$ frozen they have $49.62$ ($15.88$) and forget $+2.15$ points.
The matched free LoRA has $8.38$ ($0.00$) and forgets $-0.86$ points (descriptive).
\endblock
\block{merge}
\paragraph{On the minor band, \texorpdfstring{$A$}{A} frozen loses more when adapters are merged.}
We \emph{merge} adapters.
The adapters trained on OpenBookQA, ARC-Easy and ARC-Challenge are added to the same model.
For each task, we measure the accuracy of the merged model minus that of the task's own adapter.
On the minor band, $A$ frozen loses more than $B$ frozen when merged, by $8.59$ points at rank 2 and $4.28$ at rank 32: A $-$ B is $-8.59$ $[-11.21, -5.94]$ and $-4.28$ $[-6.35, -2.22]$,
both established negative.
On the dominant band, no difference is detected (at most $6.32$ and $4.93$ points).
\Cref{sec:discussion} discusses these side effects together with transfer to other tasks.
\endblock
\block{controlsintro}
\Cref{tab:controls} lists every control of \Cref{sec:robustness}, with the explanation it addresses, its result and its status.
\Cref{app:experiments} names each experiment, and Section~S1 of the supplementary material describes it.
\endblock
\block{controlslist}
\paragraph{Controls on the training setup.}
Each of the following controls changes one part of the training and asks whether the advantage of $A$ frozen survives.
\begin{itemize}
\item \textbf{The scale of the update.} LoRA multiplies the update by $\alpha/r$, and rsLoRA~\citep{kalajdzievski2023rslora} proposes $\alpha/\sqrt{r}$ instead. If the
  advantage came from the effective step size, it would change with $\alpha$. We vary $\alpha/r$ from $1/4$ to 4, rsLoRA's scale
  included, and select the learning rate again each time. A $-$ B is established positive in all six tests on OpenBookQA, from
  $+6.65$ to $+10.55$ points.
\item \textbf{The optimiser.} We train with AdamW~\citep{loshchilov2019adamw}, which rescales each coordinate of the gradient and
  could treat the two factors differently. With SGD instead of AdamW, A $-$ B is $+7.93$ and $+8.47$ points on the two bands of OpenBookQA, both established.
  Under SGD the free LoRA collapsed on 5 of 8 seeds (test scores $19.6$ to $32.8\,\%$, mean $35.10\,\%$), so it is not a reference there.
\item \textbf{The step size per module.} With AdamW, each trained coordinate moves by about the learning rate at each step.
  A step therefore moves $\Delta W$ by different amounts when $A$ or $B$ is the trained factor. We rescale the frozen factor so that one sign-like step moves
  $\Delta W$ by the same amount in both configurations. The rescaling does not cover AdamW's weight decay. With this rescaling, A $-$ B is established positive in all four tests, from $+9.70$ to $+18.34$
  points.
\item \textbf{The number of updates.} With four times the updates on the same data (batch 8 instead of 32), A $-$ B is $+6.27$ and
  $+10.98$ points on OpenBookQA, both established.
\item \textbf{The frozen basis.} Freezing $A$ at LoRA-FA's random initialisation, instead of a spectral band, keeps the advantage over $B$
  frozen on each spectral band. It holds in all four tests (from $+9.62$ to $+18.42$ points). These tests share the $B$-frozen runs of the main experiments.
  With 24 random orthonormal bases for each factor on the format task, A $-$ B is
  $+11.07$ points $[+9.35, +12.84]$ over eight seeds (established; outside the Holm family).
  Seeds 2 and 3 ran on an earlier version of the code than seeds 4 to 9; on seeds 4 to 9 alone the difference is $+11.00$ $[+9.21, +12.83]$ (descriptive).
  The first test of this control, on two seeds, gave $+11.28$ $[+9.14, +13.37]$.
  Between bases, accuracy has a standard deviation of $1.34$ points for $B$ frozen and $0.65$ for $A$ frozen, so the choice of random basis moves accuracy much less than the choice of factor.
  The 288 runs took $46.2$ GPU-hours, above the first estimate, and we raised the compute cap of this control.
\item \textbf{The amount of data.} With one eighth or one half of the training data and the same number of steps, A $-$ B is
  established positive in 7 of 8 tests.
\end{itemize}
\endblock
\block{controlstable}
\begin{scriptsize}
\begin{longtable}{p{3.3cm}p{3.0cm}p{5.0cm}p{3.0cm}}
\caption{\textbf{Controls.} Each row is one control, on the two main pairs at rank 2 unless stated.
``Addresses'' is the alternative explanation, or the side effect, the control examines.
``A $-$ B'' is the accuracy of $A$ frozen minus that of $B$ frozen on the same band, in accuracy points.
Status as defined in \Cref{tab:statusterms}: established positive or negative, or not established, written ``at most $X$'' with
$X$ the largest effect its interval allows; ``contradicted'' means a prediction fixed in advance is established in the opposite
direction.
\dag: post hoc (\Cref{sec:prespec}); a \dag\ row marked ``descriptive'' has no test.
\Cref{app:experiments} names each control, and Section~S1 of the supplementary material describes it.
Eight seeds per configuration unless stated.}
\label{tab:controls}\\
\toprule
Control & Addresses & Result & Status \\
\midrule
\endfirsthead
\multicolumn{4}{l}{\textit{\Cref{tab:controls}, continued}} \\
\toprule
Control & Addresses & Result & Status \\
\midrule
\endhead
Square modules only (query and output projections) & the extra parameters of $A$ frozen & A $-$ B $+7.93$ on OpenBookQA; format task not computed ($B$ frozen collapsed) & established on OpenBookQA \\
$\alpha/r$ from $1/4$ to 4, rsLoRA's scale~\citep{kalajdzievski2023rslora} included & the scale of the update & A $-$ B from $+6.65$ to $+10.55$ (OpenBookQA) & 6 of 6 established \\
SGD instead of AdamW & AdamW's rescaling & A $-$ B $+7.93$ and $+8.47$ (OpenBookQA); the free LoRA collapsed on 5 of 8 seeds, so it is void as a reference & 2 of 2 established \\
Same step size per module & a larger step for $A$ frozen & A $-$ B $+9.70$ to $+18.34$ & 4 of 4 established \\
4$\times$ updates on the same tokens (batch 8) & a too short optimisation & A $-$ B $+6.27$ and $+10.98$ (OpenBookQA) & 2 of 2 established \\
$A$ frozen at LoRA-FA's random init. & the need for a spectral $A$ & A $-$ B $+9.62$ to $+18.42$ & 4 of 4 established; they share the main $B$-frozen runs \\
24 random bases per factor, 8 seeds & the spectral subspace & A $-$ B $+11.07$ (format task) & established \\
1/8 and 1/2 of the data & the amount of data & 7 of 8 tests established & 1 not established (at most $6.70$) \\
4$\times$ training budget, best checkpoint & under-training of $B$ frozen & OpenBookQA: minor band $+8.45$; dominant band $+3.62$; at the last step $+3.52$ and $-1.75$ & best checkpoint: minor established, dominant not (at most $7.35$); last step: both intervals include 0 (descriptive) \\
ARC, best checkpoint & scoring the last step & 1 of 4 tests established & three not established \\
LoRA+ free LoRA & an under-tuned reference & $A$ frozen, dominant band, minus LoRA+: $-3.01$ (format task) & established negative, suggestive \\
Only a $16 \times 16$ core trained, both factors frozen & which side's basis limits the adapter, LoRA-XS's placement & learned output basis minus learned input basis: $+7.31$ (format task), $+2.70$ (OpenBookQA); spectral minus random: $+2.98$ (OpenBookQA) & format established, OpenBookQA not (at most $7.38$); spectral beats random on OpenBookQA only \\
Predictor derived from Zhu et al.\posthoc & the asymmetry as a least-squares effect & rank correlation $-0.362$ over 40 configurations & contradicted \\
Full fine-tuning & the ceiling & above $A$ frozen by $+4.58$ (OpenBookQA), $+5.54$ (format task) & descriptive \\
Forgetting of HellaSwag & a side effect outside the task & $B$ frozen minus $A$ frozen: negative, $A$ forgets more & 4 of 16 established negative \\
Forgetting at matched task accuracy\posthoc & the effect of learning more & $A$ minus $B$ frozen, HellaSwag: $-0.25$ (format task, dominant band, 16 seeds) & not established (at most $2.32$) \\
Merging three adapters & a side effect of merging & $A$ minus $B$ frozen, minor band: $-8.59$, $-4.28$ (ranks 2, 32) & established negative; dominant band not established \\
Summed likelihoods; the letter prompt\posthoc & the length normalisation of scores & with summed likelihoods, A $-$ B keeps its sign in 11 of 12 settings (not rank 64, dominant band); under the letter prompt, a change of format, it is negative in 6 of 12 & descriptive \\
Eight further controls (loss on the answer, transfer across tasks, removing modules, energy at the base model, rank and band over all comparisons, repeated runs, GSM8K, controls repeated at rank 16) & each addresses a scoring artefact, a side effect or the ceiling & Table~S3 of the supplementary material & descriptive \\
\bottomrule
\end{longtable}
\end{scriptsize}
\endblock
\block{rankexp}
On the two main pairs, four experiments cover the ranks of \Cref{sec:rank}.
All but the last use the grid of learning rates of \Cref{sec:rates}.
\begin{itemize}
\item \emph{Ranks 2, 16, 32 and 64}, seeds 2 to 9. At ranks 2, 32 and 64 this experiment tests A $-$ B on each band and $F$
  against $|S|$. At rank 16 it tests only $F$ against $|S|$. Ranks 32 and 64 share one family of four tests, at level 98.75\,\%.
\item \emph{Ranks 4 and 8}, and \emph{rank 1}: two more experiments, seeds 2 to 9, with the same tests.
\item \emph{A replication at ranks 16 and 32}, on eight new seeds (12 to 19, with the learning rate chosen on seeds 10 and 11) and a
  finer grid. The grid has five rates, each $\sqrt{2}$ times the previous one, centred on the rate that the first experiment selected for
  the same configuration. Its protocol tests the free LoRA against MiCA's configuration (\Cref{sec:methods}). Its tests of
  A $-$ B on each band and of $F$ against $|S|$ are post hoc\posthoc. We added them after reading ranks 4 and 8.
  They were written down before the replication ran.
\end{itemize}
\endblock
\block{mistralrates}
\paragraph{The learning rates of Mistral-7B at ranks 32 and 64.}
Five configurations are selected at each rank: the four frozen ones and the free LoRA.
At rank 32, three of them chose the lowest rate of the grid.
At rank 64, four did.
For each of these, the extra step below the grid that the edge rule allows was tried and did worse.
Their selected rate is therefore not at the edge of what was tried.
The fifth at rank 64, $A$ frozen on the dominant band, chose that extra step, on an unresolved tie.
When the best rate is the lowest one tried, a still lower rate might be better, so this configuration may be slightly under-tuned.
The result on the dominant band at rank 64 involves this configuration.
It is not established, and we do not read it as the absence of an effect.
\endblock
\block{posthocpoints}
\paragraph{The post hoc points of \Cref{fig:rank}.}
The post hoc points of \Cref{fig:rank}, listed in its caption, are:
\begin{itemize}
\item the format task at rank 1 (not blind);
\item the replication's A $-$ B at ranks 16 and 32 (tests added after ranks 4 and 8 were read);
\item Qwen2.5-7B at rank 16 (decided after ranks 2 and 32);
\item ranks 16 and 32 of the mixture (decided after ranks 2 and 8);
\item rank 64 of Mistral-7B (decided after ranks 2 and 32).
\end{itemize}
The tests of these points were written down before their runs.
\endblock
\block{notest}
A $-$ B is not established in these cases (\Cref{sec:rank}); the number in brackets is the largest effect the interval allows,
in points:
\begin{itemize}
\item rank 16\posthoc, dominant band: OpenBookQA with Qwen2.5-1.5B (at most $5.42$) and with Qwen2.5-7B (at most $5.47$);
\item rank 32: OpenBookQA with Qwen2.5-1.5B on both bands, in both experiments (at most $4.40$ and $4.80$ in the
  replication\posthoc).
  Also the format task on the minor band in the replication (at most $6.05$) and on both bands in the first experiment.
  Also Qwen2.5-7B on both bands (at most $3.08$ and $5.97$);
\item rank 64: both main pairs on both bands (at most $4.80$ on OpenBookQA and $4.34$ on the format task), and Mistral-7B on the
  dominant band (at most $3.42$).
\end{itemize}
\endblock
\block{transfer}
These results\posthoc\ are descriptive: they were analysed after the main experiments, without a test (\Cref{sec:prespec}).
We train adapters on OpenBookQA with Qwen2.5-1.5B at rank 2 and evaluate them on ARC-Easy, where the base model scores $70.92\,\%$.
The free LoRA reaches $67.59\,\%$.
$A$ frozen reaches $66.06\,\%$ on the dominant band and $61.91\,\%$ on the minor band.
$B$ frozen reaches $71.91\,\%$ and $69.90\,\%$ on the two bands.
From ARC-Easy to OpenBookQA, $A$ frozen reaches $49.48\,\%$ and $49.78\,\%$ (dominant and minor band), against $48.45\,\%$ and
$45.20\,\%$ for $B$ frozen.
From OpenBookQA to ARC-Challenge, the base model scores $43.94\,\%$.
On the dominant band, $A$ frozen reaches $46.81\,\%$ and $B$ frozen $45.56\,\%$.
On the minor band, $A$ frozen reaches $43.48\,\%$ and $B$ frozen $43.14\,\%$.
With the summed log-likelihood instead of the length-normalised rule, the ordering on ARC-Easy is the same: $A$ frozen reaches $68.91\,\%$ and $64.70\,\%$, $B$ frozen $73.47\,\%$ and $71.90\,\%$.
\paragraph{New sentences for the format task.}
On 283 new CoNLL-2003 sentences of the task's own kind, A $-$ B is $+9.63$ $[+5.74, +13.65]$ on the dominant band and $+15.90$ $[+12.06, +19.83]$ on the minor band (descriptive).
These values are close to those of the main test set.
On 1,717 new sentences with exactly two entities, a kind absent from the task's data, the tests are not computed, because the mean of $B$ frozen falls below the $45\,\%$ threshold of the protocol.
There $A$ frozen on the dominant band scores lower still, $4.92\,\%$ against $33.82\,\%$ for $B$ frozen (descriptive): the advantage does not transfer to this kind of sentence.
\endblock
\block{deviationsintro}
Every experiment reports its departures from its protocol.
\Cref{tab:deviations} lists those that bear on a result in the body, and Table~S2 of the supplementary material gives each one.
\Cref{tab:ties} counts, over the selected configurations of each experiment, the learning-rate selections that ended on an unresolved tie or at the edge of the grid.
A selection is at the edge when its rate is the smallest or the largest tried, after the extra step that the edge rule allows (\Cref{app:rules}): one step, or two in two experiments.
Two of these ties are between rates with equal validation counts; the tie rule leaves them unresolved, and floating-point rounding chose the rate kept. One is in the first experiment on OpenBookQA, the other in the first experiment on the format task.
The selected rates are unchanged.
Ties and edge rates are routine: the rules of \Cref{sec:rates} handle them, and each is declared in its experiment's report.
We count them to show how often a reported value could have come from a neighbouring, equally good rate.
\endblock
\block{deviations}
\begin{small}
\begin{longtable}{p{3.2cm}p{6.3cm}p{4.6cm}}
\caption{\textbf{Deviations that bear on a result in the body.} The first group of rows are decisions taken after related results had been read. Each is marked post hoc\posthoc. The other rows are departures from the protocol, with what the paper does about each.}
\label{tab:deviations}\\
\toprule
Experiment & What happened & What the paper does \\
\midrule
\endfirsthead
\multicolumn{3}{l}{\textit{\Cref{tab:deviations}, continued}} \\
\toprule
Experiment & What happened & What the paper does \\
\midrule
\endhead
\multicolumn{3}{@{}l}{\textit{Decided after related results had been read. The tests were fixed before the runs. Marked post hoc\posthoc.}} \\
Replication at ranks 16 and 32 & Its tests of A $-$ B and of $F$ against $|S|$ were added after ranks 4 and 8 had been read. & Marked post hoc\posthoc. \\
Qwen2.5-7B, rank 16; Qwen2.5-7B, ARC-Challenge and format task & Decided after the other 7B results had been read. & Marked post hoc\posthoc. \\
Output overlap, rank 8 & Decided after rank 2 had been read. & Marked post hoc\posthoc. \\
Flipped MiCA & Decided after the rank-16 results of the replication had been read. & Marked post hoc\posthoc. \\
Forgetting at matched accuracy & Decided after the forgetting and merging results had been read. & Marked post hoc\posthoc. \\
Eight-task mixture, ranks 16 and 32 & Decided after ranks 2 and 8 had been read. & Both ranks marked post hoc\posthoc. \\
Mistral-7B, rank 64 & Decided after ranks 2 and 32 had been read. & Marked post hoc\posthoc. \\
\midrule
\multicolumn{3}{@{}l}{\textit{Fixed after the results they predict or compare had been read.}} \\
Least-squares predictor; every learning rate of $B$ frozen; factor against band on three pairs & Fixed after the results they predict or compare had been read. & Marked post hoc\posthoc. \\
Rank 1, format task & Nine of the runs with $B$ frozen at rank 1 existed, and had been read in an earlier experiment, before the rank-1 protocol was written. & Marked post hoc\posthoc\ (\Cref{sec:rank}, \Cref{fig:rank}). \\
Forgetting at rank 16 & The rank-16 forgetting values had been shown in a descriptive map before the tests were fixed. The records of the earlier evaluation were read again and nothing was evaluated. & Marked post hoc\posthoc\ (\Cref{app:robust}). \\
\midrule
\multicolumn{3}{@{}l}{\textit{Departures from the protocol.}} \\
Mistral-7B, ranks 32 and 64 & Most selected rates were the lowest of the grid, and the extra step below it was tried and did worse. At rank 64, $A$ frozen on the dominant band chose that extra step, on an unresolved tie. & Stated in \iftmlr\Cref{app:rankexp}\else\Cref{sec:rank}\fi; the result on the dominant band at rank 64 is not read as the absence of an effect. \\
Format task, rank 16, replication & One run with $B$ frozen on the dominant band diverged and scored 0. It was kept, as the protocol requires. It drives A $-$ B on that band and the comparison of the two bands with $B$ frozen. & Reported with a post hoc\posthoc\ check at the other tied rate and a rerun of the same seed (\Cref{sec:rank}, \Cref{app:collapse}). \\
Flipped MiCA & No adapter dropout in any configuration, where MiCA's recipe uses dropout. & Stated in \Cref{app:methodarms}. \\
Flipped MiCA, its relation to LoRA-RLMO & The protocol requires the paper to name the relation of the flipped configuration to LoRA-RLMO. & Named in \Cref{sec:background,sec:configs}: LoRA-RLMO starts $A$ on the minor right singular vectors with $B = 0$, the starting point of flipped MiCA. \\
Random frozen bases & Seeds 2 and 3 ran on an earlier version of the code than seeds 4 to 9. & Stated in \Cref{app:controls} and \Cref{tab:controls}. \\
Square modules, format task & $B$ frozen collapsed on these modules, so the tests were not computed. & The control is stated for OpenBookQA only. \\
Qwen2.5-7B, format task & This task ran on a different GPU model from the other 7B tasks, so the GPU and the task are confounded. & No difference between tasks is tested. \\
A frozen on activation bands & Two of the eight runs of $A$ frozen on the activation-null directions collapsed (test scores $16.4$ and $20.4\,\%$). The established comparison partly rests on them. & Kept in the appendix with this reservation (\Cref{app:activation}). \\
SGD & Under SGD, the free LoRA collapsed on 5 of 8 seeds at its selected rate (mean $35.10\,\%$), so the free-LoRA reference is void. & Only A $-$ B is used from this experiment; the comparison with the free LoRA under SGD is not used. \\
Forgetting at matched accuracy, minor band & $A$ frozen at its lowered rate was still $4.12$ points more accurate than $B$ frozen, outside the margin fixed in advance, so the match did not hold. & Its results are not read as results at equal accuracy: the side effect at equal accuracy is not established on that band (\Cref{sec:robustness}). \\
Output overlap, OpenBookQA, rank 2 & The bootstrap interval of the rank correlation is affected by resamples in which the same seed is drawn twice. & Stated in \Cref{app:overlap}. \\
Intruder dimensions and forgetting & The text asks that the analysis not run while other training is running, and it did. Its null draws come from one stream per adapter, a child of one seed sequence, because the adapters are separate jobs, where the text names a single stream. & Stated in \Cref{app:robust}; the null counts are a descriptive reference, and no test uses them. \\
\bottomrule
\end{longtable}
\end{small}
\endblock
\block{collapse}
With $B$ frozen on the dominant band at rank 16 on the format task with Llama-3.2-1B, seed 14 of the eight reported seeds scored
$0.0\,\%$.
From step 50 its training loss rose above those of the other seeds; it ended at $1.65417$, against $0.01852$ to $0.02473$ for the
seven other seeds.
The selected rate is at the edge of the grid and tied with the next rate.
Every reported run is kept, so the reported A $-$ B on this band is $+15.17$ points, established.
The bootstrap interval of this value, $[+3.64, +36.50]$, excludes zero. The paired $t$ interval over the eight seeds, $[-11.08, +41.41]$, contains zero.
The paired $t$ interval is computed on the eight per-seed differences alone, so it ignores the noise of the test items.
It includes zero because the one diverged seed spreads these differences widely.
Two checks measure the effect of this run.
Both are post hoc\posthoc: they were decided after the divergence was seen, and written down before they were run.
At the other tied rate, A $-$ B is $+3.58$ points ($95\,\%$ interval $[+1.30, +5.90]$).
$B$ frozen on the dominant band minus $B$ frozen on the minor band (the opposite sign of the band effect of $B$ frozen in $S$) is $-0.55$ ($[-3.20, +2.12]$)
instead of the reported $-12.13$.
The reported $-12.13$ is established negative, suggestive (bootstrap interval $[-40.36, -0.08]$).
Its paired $t$ interval over the eight seeds, $[-43.56, +19.29]$, contains zero.
Seed 14 rerun at the selected rate scored $75.6\,\%$ and did not diverge again.
The verdict factor of the replication at this rank also depends on this run.
Its cross difference $F + S$ uses $B$ frozen on the dominant band, and its paired $t$ interval over seeds, $[-10.45, +41.51]$, contains zero.
\endblock
\block{overlapdetail}
\paragraph{How the five levels are built.}
For each module, we draw a Gaussian matrix, remove its component along $U_L$, and make it orthonormal with a QR decomposition: this gives $U_\perp$.
$U_\perp$ is drawn once per module and serves every level and every seed, so the results are conditional on this draw.
The five levels are $c^2 = 0$, $1/4$, $1/2$, $3/4$ and $1$.
Level 0 is a random basis orthogonal to $U_L$, and level 1 is $U_L$ itself.
We committed the bases before any run, and checked that their columns are orthonormal and make the stated angle with $U_L$.
At rank 8\posthoc, the construction is the same, with eight learned directions.
The input-side experiment freezes $A$ on $V(c) = c\,V_L + \sqrt{1 - c^2}\,V_\perp$ in the same way, where $V_L$ holds the leading right singular vectors of the same mean update.
\paragraph{The oracle declaration.}
$U_L$ and $V_L$ come from free LoRAs trained on the same training data as the new runs, and these LoRAs saw no test item.
The levels therefore use training-set information that a frozen-factor method would not have.
They are an oracle construction, not a method.
On the input side, level 1 bounds what a placement of the frozen $A$ can reach.
\paragraph{The share of the gap recovered.}
At each level, the share of the gap recovered is $P(c) = (a_c - a_B) / (a_F - a_B)$.
Here $a_c$ is the accuracy of $B$ frozen at level $c$, $a_B$ that of $B$ frozen on the dominant band, and $a_F$ that of the matched free LoRA, all at the same rank.
\Cref{tab:overlaplevels} gives it at each level.
At level 0 it is between $0.055$ and $0.182$, and each of its four $95\,\%$ intervals includes zero.
From $c^2 = 1/4$ on, it is between $0.865$ and $1.166$, so a quarter of overlap already recovers most of the gap.
\begin{table}[h]
\centering
\small
\caption{\textbf{A quarter of overlap with the learned output directions recovers most of the gap of $B$ frozen.}
The share $P(c)$ of the gap between $B$ frozen on the dominant band and the matched free LoRA that $B$ frozen recovers at each overlap level $c^2$ (descriptive; 1 is the whole gap).
Last row: the gap itself, in accuracy points. \dag: post hoc.}
\label{tab:overlaplevels}
\begin{tabular}{lrrrr}
\toprule
& \multicolumn{2}{c}{rank 2} & \multicolumn{2}{c}{rank 8\posthoc} \\
\cmidrule(lr){2-3}\cmidrule(lr){4-5}
$c^2$ & OpenBookQA & format task & OpenBookQA & format task \\
\midrule
0 & $+0.182$ & $+0.078$ & $+0.154$ & $+0.055$ \\
1/4 & $+0.910$ & $+0.865$ & $+1.043$ & $+1.000$ \\
1/2 & $+0.988$ & $+0.885$ & $+1.063$ & $+1.166$ \\
3/4 & $+1.068$ & $+0.988$ & $+1.071$ & $+0.908$ \\
1 & $+1.133$ & $+0.972$ & $+1.071$ & $+1.035$ \\
\midrule
gap (points) & $+10.32$ & $+15.11$ & $+6.32$ & $+8.20$ \\
\bottomrule
\end{tabular}
\end{table}
\paragraph{A caveat on the intervals of the rank correlation.}
The interval of each rank correlation is the percentile interval that the protocol fixed.
In a resample, a seed drawn twice gives tied points, which lowers the variance of the ranks.
On OpenBookQA at rank 2, only $20.7\,\%$ of the resampled correlations are at least as far from zero as the estimate, $+0.902$, so the interval $[+0.289, +0.963]$ lies mostly below it.
The same share is $31.0\,\%$ on the format task at rank 2, and $49.4\,\%$ and $50.6\,\%$ at rank 8.
\endblock
\block{activation}
Freezing $A$ on input directions the task's inputs barely use makes it worse than $B$ frozen on the format task, with two of eight runs collapsed\posthoc.
\paragraph{The activation bands.}
For each adapted module, we take the uncentred second moment of its inputs in the base model, on 512 training sequences of the task.
The \emph{activation-dominant band} holds the eigenvectors of its $r$ largest eigenvalues, and the \emph{activation-null band} those of its $r$ smallest.
At rank 2, $A$ is frozen on one of these bands and $B$ trains from zero, with as many trainable parameters as the other configurations with $A$ frozen.
LoRA-Null~\citep{tang2026loranull} also freezes $A$ near a null space of the activations, but of pre-training data and at rank 128.
\paragraph{Three tests per pair.}
The protocol fixed three tests on each pair, at level $1 - 0.05/3$.
The first compares the activation-dominant band with the activation-null band, both with $A$ frozen.
The second compares $B$ frozen on the dominant band with $A$ frozen on the activation-null band: it asks whether the asymmetry reverses.
The third compares the activation-null band with the base model, to check that this configuration learns.
\paragraph{Format task.}
All three tests are established.
$B$ frozen on the dominant band beats $A$ frozen on the activation-null band by $+15.09$ points $[+2.12, +32.42]$.
Two of the eight seeds of $A$ frozen on the activation-null band collapse, to $16.4\,\%$ and $20.4\,\%$.
With them, the paired $t$ interval over seeds, $[-6.43, +36.60]$, contains zero.
The rate selected for this configuration is at the edge of its grid.
A follow-up decided after the results tried larger rates on seeds 0 and 1 and selected the same rate, so the test is unchanged.
\paragraph{OpenBookQA.}
The activation-dominant band beats the activation-null band by $+8.43$ points, and $A$ frozen on the activation-null band still learns, $+3.88$ points above the base model.
We detect no reversal (at most $1.73$ points), but this test is underpowered.
An addendum written before any run says why: $B$ frozen on the dominant band is only $2.45$ points above the base model on this pair, and the smallest difference the test could detect is about $5.6$ points.
So the OpenBookQA arm says nothing about a reversal.
\endblock
\block{ratetiesdetail}
\paragraph{Llama-3.1-8B: scope and declarations.}
\label{app:llama8b}
The runs cover OpenBookQA at ranks 8 and 16 and the format task at rank 8, with seeds 2 to 9, and their tests are outside the Holm family.
The base model scores $46.80\,\%$ on OpenBookQA and $0\,\%$ on the format task.
We reduced the scope to these ranks after a timing run exceeded its bound, so ranks 2 and 32 were not run, and the choice of rank 8 came after that timing run.
One machine was excluded by hand, and the pair of timing runs of the format task was rerun.
The files used to resume one interrupted run were moved before the reading.
At rank 8 on OpenBookQA, the matched free LoRA beats $B$ frozen on the minor band by $+8.30$ points $[+4.10, +12.45]$, established.
These runs used a different GPU model from most of the 7B runs (all but the format task with Qwen2.5-7B), so we report their statuses beside the 7B statuses and compare no effect sizes across model sizes.

\paragraph{Results at the other tied rate.}
\label{app:othertied}
In the eight-task mixture at ranks 2, 16 and 32, two learning rates tied at the selection (\Cref{tab:tiedrates}).
We reran the other tied rate, in 64 runs; this reading is descriptive and post hoc, and it changes no status.
A $-$ B stays positive in all six cases (three ranks, two bands), and no interval at the other rate contains zero.
At rank 32 on the dominant band, the selected rate gives $+0.22$ $[-0.24, +0.68]$, not established, and the other rate gives $+0.78$ $[+0.30, +1.28]$, so the reading we kept is the more cautious one.
One reference check of this rerun failed at ranks 16 and 32, by about one point, so a gap of that size between the two rates may come from the code version and not from the rate.

\begin{table}[h!]
\centering\small
\caption{\textbf{In the eight-task mixture, the other tied learning rate never changes the sign of A $-$ B.}
A $-$ B in accuracy points with its interval, at the selected rates and at the other rate that tied in the selection (descriptive and post hoc).}
\label{tab:tiedrates}
\begin{tabular}{rlll}
\toprule
Rank & Band & Selected rates & Other tied rate \\
\midrule
2 & dominant & $+1.17$ $[+0.51, +1.82]$ & $+1.60$ $[+0.99, +2.21]$ \\
 & minor & $+1.95$ $[+1.27, +2.64]$ & $+2.61$ $[+1.87, +3.35]$ \\
16 & dominant & $+0.73$ $[+0.20, +1.26]$ & $+0.61$ $[+0.03, +1.20]$ \\
 & minor & $+1.98$ $[+1.45, +2.51]$ & $+1.33$ $[+0.81, +1.85]$ \\
32 & dominant & $+0.22$ $[-0.24, +0.68]$ & $+0.78$ $[+0.30, +1.28]$ \\
 & minor & $+0.84$ $[+0.32, +1.35]$ & $+1.32$ $[+0.81, +1.82]$ \\
\bottomrule
\end{tabular}
\end{table}

In the first OpenBookQA experiment at rank 16, the two best rates of $A$ frozen on the dominant band had equal validation counts, 1,006 of 2,000 each, and floating-point rounding chose $2 \times 10^{-3}$ over $10^{-3}$.
We reran this configuration at $10^{-3}$ on the eight reported seeds (post hoc, descriptive).
A $-$ B on the dominant band, which this experiment did not test, is then $+1.68$ $[-1.10, +4.50]$ instead of $+1.53$ $[-1.67, +4.70]$ recomputed from the reported runs, $F$ is $+2.85$ instead of $+2.78$, and $S$ is $-3.32$ instead of $-3.25$; the reported results keep the rate that was chosen.
\endblock
\block{mechintro}
As in \Cref{sec:mechanism}, A $-$ B is the accuracy of $A$ frozen minus that of $B$ frozen on the same band.
\endblock
\block{share}
\paragraph{The overlap experiment, share of the gap.}
On OpenBookQA at rank 2, a basis orthogonal to the learned directions recovers $0.182$ of the gap between $B$ frozen on the dominant
band and the free LoRA.
Its $95\,\%$ interval includes zero.
A squared cosine of one quarter recovers $0.910$ on OpenBookQA and $0.865$ on the format task at rank 2 (descriptive).
The rank correlations at rank 2 are $+0.902$ on OpenBookQA and $+0.753$ on the format task.
At rank 8\posthoc, they are $+0.524$ and $+0.451$.
\endblock
\block{unfreeze}
\paragraph{The advantage of \texorpdfstring{$A$}{A} frozen is about as large as the cost of keeping \texorpdfstring{$B$}{B} frozen.}
A $-$ B mixes two effects: the cost of keeping a factor frozen during training, and the effect of which factor starts on the band.
To separate them, we also train each frozen configuration with its frozen factor unfrozen.
On each band, this gives four accuracies.
The accuracy $a$ is for $A$ frozen.
The accuracy $a'$ is for the same start ($A$ on the band, $B$ at zero) with both factors trained: this is LoRA-RLPO on the dominant band and LoRA-RLMO on the minor band~\citep{rlvrortho2026}.
The accuracy $b$ is for $B$ frozen.
The accuracy $b'$ is for the same start as $B$ frozen with both factors trained.
Then
\begin{align*}
\text{A} - \text{B} = a - b &= I + D, \\
I &= (b' - b) - (a' - a), \\
D &= a' - b'.
\end{align*}
$I$, which we call the \emph{interaction}, measures how much more it costs to keep $B$ frozen than to keep $A$ frozen.
The term $b' - b$ is the gain from unfreezing $B$, and $a' - a$ the gain from unfreezing $A$.
$D$ measures whether the starting point still matters once both factors are trained.
On both main pairs and both bands, $I$ is established positive: $+10.02$ and $+10.85$ points on OpenBookQA (dominant and minor band),
$+13.21$ and $+20.64$ on the format task.
$D$ is not established on any of the four (at most $4.40$ points).
The starting point therefore adds no detectable effect, but an effect of up to that size cannot be excluded.
On OpenBookQA, for example, unfreezing $B$ on the dominant band gains $+11.40$ points, while unfreezing $A$ gains $+1.38$ (descriptive).
So $I$ alone is about as large as A $-$ B: the cost of keeping $B$ frozen accounts for about as much as the whole of A $-$ B.
\endblock
\block{inputside}
\paragraph{The learned input directions help on one pair only.}
We ran the same overlap experiment on the input side, freezing $A$ at a controlled overlap with the learned input directions.

Freezing $A$ on the learned input directions themselves beats $A$ frozen on the dominant band by $+4.82$ points on the format task
(established), but no difference is detected on OpenBookQA.
If a frozen $A$ worked because good input directions are easy to find, the learned input directions should help on both pairs.
They help on one only, so we do not make that claim.
\endblock
\block{bothlearned}
\paragraph{When each factor is frozen on its own learned directions, no asymmetry is detected.}
The same experiment compares $A$ frozen on the learned input directions with $B$ frozen on the learned output directions $U_L$.
Each arm freezes one factor and trains the other.
No difference is detected on either pair (at most $2.40$ points on OpenBookQA and $5.05$ on the format task).
So, given the directions a free LoRA learns, we detect no difference between freezing either factor, within these bounds.
The cost of a frozen $B$ in the $2 \times 2$ design is compatible with bases that miss those directions.
\endblock
\block{coreintro}
This appendix gives the core-only experiment: its question and main test, then its other comparisons.
\endblock
\block{coreother}
With the learned directions on both sides, the core alone comes close to the free LoRA: $81.03\,\%$ against $81.37\,\%$ on the format
task and $54.80\,\%$ against $56.20\,\%$ on OpenBookQA (descriptive).
The spectral placement of LoRA-XS beats random bases on OpenBookQA, by $+2.98$ points (suggestive); on the format task no difference
is detected (at most $0.96$ points).
Bases computed before training by FCCA~\citep{ye2026fcca} are not detectably better than the spectral ones on either pair (at most $5.67$ and $7.66$
points).
\endblock
\block{gradsides}
A descriptive analysis, without training, compares how reproducible the leading subspace of each weight gradient is across disjoint
minibatches.
It does so on the input side and on the output side, per module type, rank and checkpoint.
The input side is the more reproducible in 98 of these 126 settings on OpenBookQA and in 100 of 126 on the format task.
It is the less reproducible in the others (28 and 26 settings).
The input side is therefore not the more reproducible in every setting, and the analysis does not test whether the difference explains the asymmetry.
\Cref{tab:gradobqa,tab:gradformat} give the values behind these counts.
The two trained checkpoints are the base model merged with the matched free LoRA of the rank-2 and of the rank-64 experiments (seed 2).

\begin{table}[h!]
\centering\scriptsize
\caption{\textbf{OpenBookQA, Qwen2.5-1.5B: agreement of the leading gradient subspaces across disjoint minibatches, input side / output side.}
Normalised agreement of the top-$r$ subspaces, mean over the ten pairs of minibatches and the sampled layers; 1 is identical subspaces. Rows: module type. Columns: $r$.}
\label{tab:gradobqa}
\begin{tabular}{lcccccc}
\toprule
Projection & $r=1$ & $r=2$ & $r=4$ & $r=8$ & $r=16$ & $r=32$ \\
\midrule
\multicolumn{7}{l}{\emph{base model}} \\
query & 0.868 / 0.705 & 0.943 / 0.777 & 0.816 / 0.735 & 0.686 / 0.691 & 0.541 / 0.652 & 0.465 / 0.558 \\
key & 0.780 / 0.726 & 0.747 / 0.841 & 0.717 / 0.801 & 0.664 / 0.782 & 0.581 / 0.664 & 0.462 / 0.551 \\
value & 0.987 / 0.894 & 0.989 / 0.841 & 0.966 / 0.800 & 0.946 / 0.729 & 0.769 / 0.524 & 0.585 / 0.370 \\
output & 0.985 / 0.888 & 0.988 / 0.872 & 0.966 / 0.808 & 0.903 / 0.699 & 0.792 / 0.509 & 0.668 / 0.343 \\
gate & 0.989 / 0.821 & 0.992 / 0.802 & 0.981 / 0.782 & 0.948 / 0.719 & 0.850 / 0.572 & 0.726 / 0.395 \\
up & 0.965 / 0.835 & 0.991 / 0.810 & 0.986 / 0.828 & 0.951 / 0.767 & 0.850 / 0.602 & 0.717 / 0.432 \\
down & 0.978 / 0.892 & 0.979 / 0.892 & 0.962 / 0.868 & 0.922 / 0.804 & 0.811 / 0.675 & 0.645 / 0.497 \\
\addlinespace
\multicolumn{7}{l}{\emph{after the free LoRA matched at rank 2}} \\
query & 0.821 / 0.082 & 0.531 / 0.255 & 0.371 / 0.428 & 0.301 / 0.558 & 0.319 / 0.527 & 0.341 / 0.445 \\
key & 0.280 / 0.510 & 0.230 / 0.727 & 0.327 / 0.587 & 0.375 / 0.492 & 0.345 / 0.463 & 0.313 / 0.424 \\
value & 0.906 / 0.002 & 0.970 / 0.004 & 0.875 / 0.011 & 0.703 / 0.028 & 0.555 / 0.069 & 0.434 / 0.126 \\
output & 0.985 / 0.002 & 0.929 / 0.004 & 0.819 / 0.009 & 0.712 / 0.022 & 0.640 / 0.034 & 0.581 / 0.050 \\
gate & 0.994 / 0.001 & 0.965 / 0.003 & 0.866 / 0.005 & 0.757 / 0.009 & 0.682 / 0.019 & 0.603 / 0.029 \\
up & 0.993 / 0.002 & 0.916 / 0.004 & 0.844 / 0.015 & 0.723 / 0.022 & 0.648 / 0.038 & 0.565 / 0.059 \\
down & 0.931 / 0.004 & 0.877 / 0.009 & 0.716 / 0.026 & 0.589 / 0.063 & 0.482 / 0.085 & 0.396 / 0.114 \\
\addlinespace
\multicolumn{7}{l}{\emph{after the free LoRA matched at rank 64}} \\
query & 0.776 / 0.134 & 0.505 / 0.304 & 0.372 / 0.441 & 0.284 / 0.577 & 0.306 / 0.530 & 0.329 / 0.442 \\
key & 0.285 / 0.531 & 0.259 / 0.752 & 0.341 / 0.610 & 0.371 / 0.508 & 0.325 / 0.467 & 0.294 / 0.419 \\
value & 0.928 / 0.020 & 0.963 / 0.044 & 0.872 / 0.041 & 0.707 / 0.043 & 0.545 / 0.066 & 0.418 / 0.111 \\
output & 0.976 / 0.062 & 0.917 / 0.040 & 0.815 / 0.030 & 0.721 / 0.029 & 0.631 / 0.030 & 0.574 / 0.038 \\
gate & 0.993 / 0.006 & 0.957 / 0.006 & 0.871 / 0.012 & 0.791 / 0.010 & 0.710 / 0.015 & 0.621 / 0.021 \\
up & 0.992 / 0.045 & 0.940 / 0.030 & 0.844 / 0.027 & 0.753 / 0.025 & 0.656 / 0.035 & 0.564 / 0.055 \\
down & 0.901 / 0.094 & 0.788 / 0.073 & 0.717 / 0.068 & 0.590 / 0.075 & 0.475 / 0.080 & 0.385 / 0.092 \\
\bottomrule
\end{tabular}
\end{table}

\begin{table}[h!]
\centering\scriptsize
\caption{\textbf{Format task, Llama-3.2-1B: agreement of the leading gradient subspaces across disjoint minibatches, input side / output side.}
Normalised agreement of the top-$r$ subspaces, mean over the ten pairs of minibatches and the sampled layers; 1 is identical subspaces. Rows: module type. Columns: $r$.}
\label{tab:gradformat}
\begin{tabular}{lcccccc}
\toprule
Projection & $r=1$ & $r=2$ & $r=4$ & $r=8$ & $r=16$ & $r=32$ \\
\midrule
\multicolumn{7}{l}{\emph{base model}} \\
query & 0.988 / 0.977 & 0.959 / 0.961 & 0.971 / 0.971 & 0.944 / 0.943 & 0.945 / 0.941 & 0.927 / 0.919 \\
key & 0.940 / 0.950 & 0.941 / 0.956 & 0.932 / 0.946 & 0.938 / 0.954 & 0.941 / 0.943 & 0.912 / 0.922 \\
value & 0.998 / 0.972 & 0.997 / 0.970 & 0.994 / 0.968 & 0.973 / 0.948 & 0.985 / 0.958 & 0.958 / 0.925 \\
output & 0.994 / 0.977 & 0.978 / 0.960 & 0.985 / 0.966 & 0.983 / 0.958 & 0.982 / 0.954 & 0.953 / 0.917 \\
gate & 0.980 / 0.964 & 0.982 / 0.969 & 0.995 / 0.983 & 0.971 / 0.957 & 0.984 / 0.965 & 0.971 / 0.940 \\
up & 0.980 / 0.965 & 0.985 / 0.971 & 0.992 / 0.978 & 0.977 / 0.961 & 0.979 / 0.960 & 0.964 / 0.937 \\
down & 0.997 / 0.974 & 0.994 / 0.978 & 0.989 / 0.977 & 0.986 / 0.974 & 0.976 / 0.960 & 0.969 / 0.947 \\
\addlinespace
\multicolumn{7}{l}{\emph{after the free LoRA matched at rank 2}} \\
query & 0.262 / 0.240 & 0.292 / 0.235 & 0.282 / 0.255 & 0.279 / 0.327 & 0.275 / 0.398 & 0.293 / 0.432 \\
key & 0.239 / 0.444 & 0.228 / 0.425 & 0.213 / 0.476 & 0.237 / 0.516 & 0.308 / 0.471 & 0.336 / 0.434 \\
value & 0.774 / 0.028 & 0.699 / 0.028 & 0.601 / 0.033 & 0.454 / 0.078 & 0.384 / 0.127 & 0.360 / 0.177 \\
output & 0.481 / 0.013 & 0.479 / 0.018 & 0.415 / 0.031 & 0.384 / 0.058 & 0.371 / 0.082 & 0.368 / 0.108 \\
gate & 0.416 / 0.013 & 0.283 / 0.016 & 0.288 / 0.028 & 0.259 / 0.050 & 0.255 / 0.068 & 0.253 / 0.092 \\
up & 0.431 / 0.025 & 0.341 / 0.035 & 0.298 / 0.052 & 0.260 / 0.072 & 0.250 / 0.091 & 0.254 / 0.107 \\
down & 0.589 / 0.018 & 0.397 / 0.043 & 0.309 / 0.067 & 0.267 / 0.078 & 0.258 / 0.096 & 0.254 / 0.118 \\
\addlinespace
\multicolumn{7}{l}{\emph{after the free LoRA matched at rank 64}} \\
query & 0.264 / 0.181 & 0.249 / 0.208 & 0.237 / 0.239 & 0.227 / 0.309 & 0.234 / 0.377 & 0.262 / 0.409 \\
key & 0.104 / 0.515 & 0.213 / 0.431 & 0.187 / 0.441 & 0.207 / 0.496 & 0.276 / 0.460 & 0.306 / 0.418 \\
value & 0.653 / 0.018 & 0.648 / 0.025 & 0.493 / 0.042 & 0.401 / 0.070 & 0.356 / 0.109 & 0.339 / 0.153 \\
output & 0.359 / 0.005 & 0.318 / 0.020 & 0.335 / 0.023 & 0.335 / 0.039 & 0.337 / 0.062 & 0.349 / 0.092 \\
gate & 0.307 / 0.004 & 0.206 / 0.009 & 0.199 / 0.019 & 0.192 / 0.032 & 0.199 / 0.056 & 0.223 / 0.081 \\
up & 0.316 / 0.005 & 0.215 / 0.009 & 0.200 / 0.022 & 0.197 / 0.038 & 0.206 / 0.062 & 0.229 / 0.084 \\
down & 0.405 / 0.005 & 0.307 / 0.011 & 0.223 / 0.035 & 0.206 / 0.052 & 0.208 / 0.071 & 0.222 / 0.098 \\
\bottomrule
\end{tabular}
\end{table}

\endblock
\block{gradientdrift}
\paragraph{How far the leading gradient subspace moves during training.}
We also compared the leading subspace of each module's weight gradient at step 0 with the one at later steps.
The analysis is descriptive\posthoc and uses one free LoRA per task, a fixed batch of 64 sequences and all modules.
The output side is the side that $B$ would have to span, the input side the one that $A$ would have to span.
We report the overlap of the two subspaces above what chance gives.
At rank 2, between step 0 and the last step, it is $0.108$ on the output side and $0.415$ on the input side on OpenBookQA, and $0.059$ and $0.190$ on the format task.
The output side is the lower one in $0.88$ of the 196 modules of OpenBookQA and in $0.85$ of the 112 modules of the format task.
For comparison, two halves of the batch at step 0 give $0.832$ and $0.946$ on OpenBookQA.
The difference between the sides gets smaller with the rank: at rank 32 the two sides are $0.199$ and $0.378$ on OpenBookQA, and $0.121$ and $0.148$ on the format task.
The measurement concerns the gradient of a free LoRA, not the learned weights, and it rests on one run per task.
It therefore does not test the hypothesis of \Cref{sec:discussion}.
\endblock
\block{mica}
\paragraph{MiCA.}
Our $B$ frozen on the minor band is MiCA's configuration.
At MiCA's recipe, our code and the PEFT implementation of MiCA differ by
$-0.23$ points on OpenBookQA and $-0.77$ on the format task (our code minus PEFT, descriptive; \Cref{tab:methods}).
MiCA's paper adapts the query and value projections only.
We adapt our seven modules, so that MiCA and our configuration differ only by their recipe, and the matched free LoRA stays matched.
\endblock
\block{micacounts}
\paragraph{Ranks and counts at MiCA's recipe.}
In \Cref{tab:methods}, the second flipped row runs at rank 13 on OpenBookQA and 14 on the format task.
Its count of trainable parameters is $1.030$ and $1.006$ times MiCA's.
The matched free LoRA runs at rank 7 and 8, with $0.992$ and $1.075$ times.
\endblock
\block{picaside}
\paragraph{Which factor PiCa freezes.}
PiCa writes the update as $U_r B$.
Here $U_r$ holds $r$ leading singular vectors of the weight and is frozen, and $B$ is trained from zero and shared across the layers of a group of modules.
PiCa writes the layer as $Y = X(W_0 + U_r B)$, with inputs $X \in \mathbb{R}^{L \times m}$, $U_r \in \mathbb{R}^{m \times r}$ and $B \in \mathbb{R}^{r \times n}$.
The inputs multiply the weight from the left, so the frozen $U_r$ spans the input side and the trained $B$ has the output dimension.
PiCa's own counts of trainable parameters agree: they are $r$ times the sum of the output dimensions of the adapted modules.
In our notation, PiCa freezes $A$ on the dominant band.
\endblock
\block{picabudget}
\paragraph{PiCa's trainable parameters and training steps.}
With its trained factor shared across layers, PiCa at rank 256 trains $5,898,240$ parameters on OpenBookQA and $6,029,312$ on the
format task.
Its recipe (per-device batch 8 with 8 accumulation steps, 3 epochs) gives 15 optimiser steps on OpenBookQA and 150 on the format task, because our training sets are small.
\endblock
\block{picaparts}
\paragraph{PiCa at its recipe, both parts.}
The experiment of \Cref{sec:methods} runs every configuration at PiCa's published rate (Part A, primary).
It runs them again with rates selected on separate seeds (Part B, secondary, each at the same level $99.375\,\%$).
Part A is in \Cref{tab:tests}.
In Part B, the matched free LoRA minus PiCa is $-1.20$ $[-3.73, +1.18]$ on OpenBookQA at rank 256, $+0.10$ $[-2.57, +2.98]$ on the
format task at rank 256 and $-0.16$ $[-3.63, +3.28]$ at rank 16.
PiCa minus $B$ frozen on the dominant band at PiCa's parameter count is $-0.42$
$[-3.52, +2.78]$, $-0.61$ $[-3.60, +2.36]$ and $+3.14$ $[-0.22, +6.53]$ in the same three settings.
None is established.
The sharing, the data and the evaluation differ from PiCa's own experiments.
\endblock
\block{picatable}
\begin{table}[t]
\centering
\footnotesize
\caption{\textbf{PiCa against the matched free LoRA in every setting run: the free LoRA minus PiCa, at PiCa's published rate (Part A) and at rates selected on separate seeds (Part B).}
Points, with the paired bootstrap interval at level $99.375\,\%$; the floor column is the free LoRA minus the base model.
Part A gives the free LoRA PiCa's own rate and 15 optimiser steps on OpenBookQA, 150 on the format task.
A test is read only if the floor is met.
On ARC-Challenge the protocol has no test (descriptive floors at ranks 256 and 16: $-1.03$ and $+1.75$).}
\label{tab:picatable}
\begin{tabular}{llrlrll}
\toprule
Part & pair & rank & floor & free LoRA $-$ PiCa & interval & status \\
\midrule
A & OpenBookQA & 256 & $+5.65$ & $-6.05$ & $[-9.87, -2.47]$ & PiCa better, established \\
A & OpenBookQA & 16 & $-0.42$ & --- & --- & floor not met, test not read \\
A & format task & 256 & met & $-0.79$ & $[-4.23, +2.70]$ & not established \\
A & format task & 16 & met & $-6.05$ & $[-9.83, -2.26]$ & PiCa better, established \\
B & OpenBookQA & 256 & $+9.55$ & $-1.20$ & $[-3.73, +1.18]$ & not established \\
B & OpenBookQA & 16 & $+10.97$ & --- & --- & not read (Part A floor failed) \\
B & format task & 256 & met & $+0.10$ & $[-2.57, +2.98]$ & not established \\
B & format task & 16 & met & $-0.16$ & $[-3.63, +3.28]$ & not established \\
\bottomrule
\end{tabular}
\end{table}
Under the protocol's rule that the weaker of the two parts is stated, no setting is established.
\label{app:picatable}
\endblock
\block{rankxvimore}
\paragraph{Rank 16: the LoRA scale, and a decomposition with new seeds.}
We repeated the rank-16 design of the main pairs at three settings of the LoRA scale $\alpha/r$: $1/4$, $\sqrt{2}$ and 4.
On the minor band, A $-$ B is established at all three settings on OpenBookQA ($+5.17$, $+4.40$ and $+4.07$ points).
On the format task it is established at $\sqrt{2}$ and 4 ($+5.39$ and $+5.01$), and not at $1/4$ (at most $4.97$).
On the dominant band, it is established only on the format task at $\sqrt{2}$ ($+4.82$); elsewhere we detect no difference (at most $12.43$ points).
The decomposition reruns the four frozen configurations on eight new seeds.
Next to each, it trains the same configuration with its frozen factor trained too, from the same start.
The cost of freezing a factor is the accuracy gained by training it.
$I$, the cost of freezing $B$ minus the cost of freezing $A$, is the part of A $-$ B that comes from the frozen factor.
On the minor band of OpenBookQA, $I$ is $+4.38$ points $[+0.17, +8.72]$, established (suggestive): at the same rank and start, freezing $B$ costs more than freezing $A$.
On the dominant band of OpenBookQA and on both bands of the format task, we detect no difference (at most $6.66$, $4.00$ and $7.30$ points).
There we do not attribute the advantage of $A$ frozen to the frozen factor.
On these new seeds, A $-$ B on the minor band is $+4.38$ on OpenBookQA, equal to $I$ because the two unfrozen controls tie at $55.77\,\%$, and $+3.77$ on the format task (descriptive).
\endblock
\block{sixteen}
\paragraph{Sixteen times the training budget on the format task.}\label{app:x16}
We trained the rank-2 design for 4,384 steps, sixteen times 274, on six seeds.
At the last step, A $-$ B is $+10.83$ $[+6.49, +14.98]$ on the dominant band and $+16.48$ $[+12.16, +20.81]$ on the minor band, both established.
The gap persists, but it shrinks on the minor band.
Along this stretched schedule, A $-$ B on that band falls from $+24.73$ at step 274 to $+16.48$, an established change of $-8.25$ $[-14.36, -2.39]$.
On the dominant band, the change ($-10.67$) is not established.
The values at step 274 come from the stretched schedule, not from separate runs of 274 steps.
$B$ frozen reaches a training loss of $0.001$, and $A$ frozen $0.000$.
Yet $B$ frozen scores $10.83$ and $16.48$ points below $A$ frozen on the test set, so the gap is a gap in generalisation.
The selected rates of $B$ frozen lie at the edge of the tested grid.
\endblock
\block{budgetfig}
\paragraph{The figure of accuracy against the training budget.}\label{app:budgetfig}
\Cref{fig:budget} is at rank 2, on both main pairs.
On the format task, A $-$ B is $+12.11$ and $+20.40$ points (dominant, then minor band) at 274 steps, $+11.18$ and $+15.96$ at 1,096 steps, and $+10.83$ and $+16.48$ at 4,384 steps, all established.
On OpenBookQA, A $-$ B is $+8.60$ and $+10.10$ at 49 steps.
At step 196 of the run with four times the budget, it is $-1.75$ $[-5.22, +1.88]$ on the dominant band and $+3.52$ $[-0.23, +7.38]$ on the minor band (descriptive 95\,\% intervals, post hoc).
The points are means over seeds 2 to 9 (2 to 7 at 4,384 steps), and the bars, drawn at the end of each schedule, span the seeds; they are not confidence intervals.
Each schedule is a separate run with its own selected learning rate, so the budget and the rate change together.
The small marks are scores at intermediate steps of a longer schedule; they are not the same as shorter runs.
Every point is the score at that step, never at the best validation checkpoint.
The figure does not show OpenBookQA beyond 196 steps, or the letter format, which uses another scoring rule.
\endblock
\block{letterformat}
\paragraph{The design trained and scored with a letter prompt.}
On OpenBookQA with Qwen2.5-1.5B at rank 2, we trained and scored the design with a lettered prompt, in two arms with their own tests.
The base model answers $76.60\,\%$ in this format.
At 49 steps, A $-$ B is $+5.95$ $[+0.42, +15.78]$ on the dominant band and $+2.85$ $[+0.32, +5.42]$ on the minor band, both established.
The dominant-band value rests on one run of $B$ frozen at $46.6\,\%$, and its paired $t$ interval over seeds, $[-5.02, +16.92]$, contains zero.
At 893 steps, A $-$ B is $+0.37$ (at most $2.65$) and $+2.42$ (at most $5.63$), not established.
The test of $F$ against $|S|$ is open in both arms.
The pre-specified reading is that, in the letter format, the difference between freezing $A$ and freezing $B$ is not shown at rank 2 on this pair.
The negative values of the letter prompt above come from adapters trained in the likelihood format and scored in the letter format; we do not know why the gap reverses there, and adapters trained in the letter format do not show it.
\endblock
\block{further}
\paragraph{Reinforcement learning on GSM8K.}
We trained Qwen2.5-1.5B-Instruct with GRPO (group relative policy optimisation), a reinforcement-learning method, on GSM8K, on the dominant band.
At rank 2, the free LoRA does not gain 2 points over the base model, so the test is not read.
At rank 32, A $-$ B is not established.
\paragraph{Where the output directions of a frozen $B$ come from.}
At rank 2, we froze $B$ on output directions from seven sources and compared each with $B$ frozen on the dominant band.
Directions learned by a free LoRA on the same task help on both pairs ($+8.70$ and $+13.76$ points, established), a replication of the overlap analysis.
The learned directions here come from one separate source run per pair, so these values differ from those of \Cref{tab:overlaplevels}.
On the format task, directions learned with shuffled targets also help ($+10.32$), and directions read from the weights alone hurt ($-7.94$), both established.
The other comparisons are not established.
\endblock
\block{pairs}
\paragraph{Pairs of task and model.}
Besides the two main pairs, the other pairs are:
\begin{itemize}
\item other tasks: ARC-Easy and ARC-Challenge with Qwen2.5-1.5B;
\item an instruction-tuned model: OpenBookQA with Qwen2.5-1.5B-Instruct;
\item a more recent base model: OpenBookQA with Qwen3-1.7B-Base, at rank 2;
\item larger models: OpenBookQA, ARC-Challenge and the format task with Qwen2.5-7B, and OpenBookQA with
  Mistral-7B;
\item several tasks at once: an \emph{eight-task mixture}, in which a single adapter is trained on BoolQ, PIQA, SIQA, HellaSwag,
  WinoGrande, ARC-Easy, ARC-Challenge and OpenBookQA together.
  It is scored by its mean accuracy over the eight test sets~\citep{clark2019boolq,bisk2020piqa,sap2019siqa,zellers2019hellaswag,sakaguchi2020winogrande,clark2018arc,mihaylov2018openbookqa}.
  GSM8K is the dataset of \citet{cobbe2021gsm8k}.
\end{itemize}
\endblock
\block{budget}
\paragraph{Training steps and tokens.}
Every run under our protocol sees the same number of supervised tokens for a task, about 600,000 (602,112 on OpenBookQA and 601,469
on the format task, where only the answer tokens count, about 69 per sequence).
With an effective batch of 32 sequences, this token count fixes the number of optimiser steps.
OpenBookQA uses 49 optimiser steps (600,000 supervised tokens, about 5.8 passes over the packed training set), and the format task 274.
The exceptions are the controls that change the training length (\Cref{sec:robustness}) and the runs at MiCA's recipe
(\Cref{sec:methods}), which follow that recipe.
OpenBookQA has 4,957 training rows and the format task 3,178.
The learning rate is selected on 400 validation items per selection seed (299 on ARC-Challenge with Qwen2.5-7B); the eight-task
mixture is selected on its 3,099 validation items.
\paragraph{Optimiser and schedule.}
Every run of the main protocol uses AdamW with $\beta = (0.9, 0.999)$, $\epsilon = 10^{-8}$ and no weight decay.
It clips gradients at a global norm of 1.0 and keeps the base weights in bfloat16.
The learning rate rises linearly during the first 5\,\% of the optimiser steps (2 of 49 on OpenBookQA, 13 of 274 on the format task) and then follows a cosine decay to zero.
Sequences are packed and cut at 384 tokens, with a per-device batch of 8 and 4 accumulation steps, which makes the effective batch of 32 sequences.
Only the answer is supervised on the format task; on OpenBookQA every token of the packed blocks is supervised.
Runs at MiCA's and PiCa's recipes use their own schedules (\Cref{sec:methods}).

\endblock
\block{testsets}
\paragraph{Test sets.}
OpenBookQA has 500 test questions, ARC-Easy 2376 and ARC-Challenge 1172, and the format task 614 test items.
The mixture is scored on the test sets of its eight tasks.
Validation and test sets are disjoint, and the training data of the mixture was checked for overlap with its test sets.
\endblock
\block{models}
\paragraph{Models and precision.}
The base models are used in bfloat16 and are never updated, except in the full fine-tuning and GaLore controls.
Elsewhere only the adapter factors are trained.
All runs of an experiment use the same GPU model, recorded for every run.
The experiments ran on NVIDIA RTX 4000 Ada cards, the card model fixed for the whole series.
Runs that landed on another card model were set aside and run again, and no score of a set-aside run was read.
The 213 runs of Mistral-7B all record this card model.
The one exception in our results is the format task with Qwen2.5-7B, which ran on NVIDIA RTX A5000 cards.
For that model, the card model and the task are confounded (\Cref{app:deviations}).
\endblock
\block{reprodetail}
\paragraph{Software, and what each run records.}
The code requires Python 3.10 or later.
Its requirements file pins PyTorch 2.14.1 (CUDA 13.0 build) and Transformers 5.20.1.
Job environments were rebuilt on each compute node, and the list was not pinned during the earlier experiments, so not every run used exactly these versions.
Every run records its Python, PyTorch, Transformers, PEFT and Datasets versions and its GPU model.
Runs load each model and tokenizer from the default branch of the Hugging Face Hub.
We did not total the compute over all experiments; two costs are given where they matter, 46.2 GPU-hours for the random-bases control and 353.13 for GSM8K with Llama-3.1-8B.

\paragraph{How the format task is built.}
The sentences come from the train, validation and test splits of CoNLL-2003, with its four entity types (person, organisation, location, miscellaneous).
We keep the sentences of 3 to 60 tokens that contain at least three entities, and shuffle each split once with seed 0.
The training file has 3,178 sentences and the test file 614.
The prompt is ``Extract the entities from the text below and answer in the alien format.'', then a line ``Text:'' with the sentence and a line ``Answer:''.
The target is one line of JSON between the two tag lines \texttt{<<alien>>} and \texttt{<</alien>>}, with six keys in a fixed order: \texttt{qzx\_pers}, \texttt{qzx\_org}, \texttt{qzx\_loc}, \texttt{qzx\_misc}, \texttt{qzx\_n} and \texttt{qzx\_sig}.
The first four keys list the entities of each type.
Person names are rewritten as ``LAST, First'', and every list is sorted alphabetically without regard to case.
\texttt{qzx\_n} is the number of entities, spelled in English words, and \texttt{qzx\_sig} is the string of their initials, in upper case and in the order of the lists.
The target is derived from the gold annotations by these rules, with no model in the loop.
Generation is greedy.
An output is correct when it obeys every rule of the format: the two tags, a single line of valid JSON, the six keys in order, sorted lists of strings, a count that matches the number of listed entities and a signature that matches their initials.
This score checks the format of the output and its internal consistency; it does not check that the listed entities are the gold ones.
\endblock
\block{params}
\paragraph{Trainable parameters.}
\Cref{sec:budgets} gives the counts on Qwen2.5-1.5B at rank 2, and \Cref{tab:methods} those at MiCA's recipe.
Across our experiments, the matched free LoRA trains between $0.9747$ and $1.1338$ times the parameters of $B$ frozen.
$A$ frozen, in turn, trains between $1.0500$ and $1.2777$ times the parameters of the free LoRA.
We give both ratios where a comparison with the free LoRA matters.
At rank 1 no matched free LoRA exists.
A free LoRA of rank 1 already trains about twice the parameters of $B$ frozen, so comparisons with it are descriptive only.
Every run's own count is checked against the count computed from the model's shapes.
\endblock
\block{latecvr}
\paragraph{Longer training at rank 16.}
\label{app:cvr16}
We trained the four configurations at rank 16 for four times the budget: 196 steps on OpenBookQA with Qwen2.5-1.5B and 1,094 on the format task (four times the supervised tokens, against $4 \times 274 = 1{,}096$ steps at rank 2) with Llama-3.2-1B, eight seeds, each run scored at its validation-best checkpoint.
On the format task, A $-$ B is $+4.13$ points $[+1.00, +7.25]$ on the dominant band and $+4.11$ $[+1.34, +6.88]$ on the minor band, both established.
On OpenBookQA, it is $+0.80$ $[-2.85, +4.52]$ and $+2.85$ $[-0.65, +6.32]$, not established.
\endblock
\block{lateflip}
\paragraph{Flipped MiCA with a 7B model.}
\label{app:flip7}
With Qwen2.5-7B on eight commonsense tasks at rank 16, in MiCA's recipe, flipped MiCA minus MiCA is $+0.37$ points on the macro-average $[-0.31, +1.05]$: we detect no difference, at most $1.05$ points.
The comparison is at equal rank, where flipped MiCA trains more parameters.
Mean accuracies are $71.38\,\%$ for MiCA, $71.75\,\%$ for flipped MiCA and $73.41\,\%$ for a LoRA of the same rank.
Flipped MiCA minus that LoRA is $-1.66$ points $[-2.17, -1.13]$, and MiCA minus that LoRA $-2.03$ $[-2.58, -1.49]$ (descriptive).
By task, the flip is established in favour of flipped MiCA on BoolQ and against it on HellaSwag and ARC-Easy; the other tasks are not resolved.
Rank 32 was not run, for lack of memory.
\endblock
\block{lateig}
\paragraph{The test of the hypothesis on the input side.}
\label{app:ig}
The hypothesis says that input directions barely move during training.
If so, $A$ frozen on the input directions of the step-0 gradient should do about as well as $A$ frozen on the input directions that a trained adapter ends with.
We tested this on the format task with Llama-3.2-1B at rank 2, with eight seeds and tests fixed before the runs, at the $95\,\%$ level.
The control is $A$ frozen on the dominant band.
The learned directions beat the control by $+3.85$ points $[+1.40, +6.35]$, established: there is room to gain.
The step-0 directions beat it by $+0.96$ points $[-1.28, +3.22]$, not established.
The strong form of the prediction, that the step-0 directions recover more than half of the room, is not rejected either: the step-0 directions minus the midpoint of the control and the learned directions is $-0.97$ points $[-3.00, +1.03]$, not established.
The share of the room recovered is $+0.249$ $[-0.540, +0.836]$ (descriptive).
The outcome fixed in advance for this pattern is ``unresolved''.

One check failed, and we declare it.
The control was rebuilt for this experiment, and its first check compares it with the original run of $A$ frozen on the dominant band: on its seed-2 run, the two gave the same parsed answer on 530 of the 614 test items, where our rule asks for $95\,\%$, with accuracies $+0.33$ points apart.
As the protocol prescribes, the tests keep the control; with the original run in its place, the learned directions gain $+4.82$ points instead of $+3.85$, and the step-0 directions $+1.93$ points $[-0.33, +4.19]$ (descriptive).
On OpenBookQA, where no test was planned, the same gain is $+1.33$ points $[-1.12, +3.70]$ (descriptive).
\endblock

\def\wantedblock{ratestability}\immediate\write\secposfile{block sec3_protocol ratestability}
\def\wantedblock{posthoctable}\immediate\write\secposfile{block sec3_protocol posthoctable}
\def\wantedblock{levels}\immediate\write\secposfile{block appendix levels}
\def\wantedblock{holm}\immediate\write\secposfile{block sec3_protocol holm}
\def\wantedblock{bounds}\immediate\write\secposfile{block appendix bounds}
\def\wantedblock{testsdetail}\immediate\write\secposfile{block sec3_protocol testsdetail}
\def\wantedblock{fsvalues}\immediate\write\secposfile{block sec3_protocol fsvalues}
\def\wantedblock{fssub}\immediate\write\secposfile{block sec3_protocol fssub}
\def\wantedblock{fsalgebra}\immediate\write\secposfile{block sec3_protocol fsalgebra}
\def\wantedblock{fsweak}\immediate\write\secposfile{block sec3_protocol fsweak}
\def\wantedblock{prespecfixed}\immediate\write\secposfile{block sec3_protocol prespecfixed}
\def\wantedblock{prespeckinds}\immediate\write\secposfile{block sec3_protocol prespeckinds}
\paragraph{The configurations.}
\def\wantedblock{configsfixed}\immediate\write\secposfile{block sec3_protocol configsfixed}
\def\wantedblock{minorband}\immediate\write\secposfile{block appendix minorband}
\paragraph{Selecting the learning rate.}
Each candidate rate is trained with seeds 0 and 1 and scored on a validation set disjoint from the test set, and the best mean is kept.
\def\wantedblock{ratesgrid}\immediate\write\secposfile{block sec3_protocol ratesgrid}
Two further rules handle a best rate at the edge of the grid and a near-tie.
\def\wantedblock{edge}\immediate\write\secposfile{block appendix edge}
\def\wantedblock{tie}\immediate\write\secposfile{block appendix tie}
\paragraph{Seeds.}
\def\wantedblock{seeds}\immediate\write\secposfile{block sec3_protocol seeds}

\section{Every contrast, with its interval}
\label{app:contrasts}
\def\wantedblock{contrastsintro}\immediate\write\secposfile{block appendix contrastsintro}
\def\wantedblock{contrasts}\immediate\write\secposfile{block tables/tab_contrasts contrasts}\block{contrasts}
\begin{small}
\begin{longtable}{p{3.3cm}rp{4.6cm}p{4.6cm}}
\caption{\textbf{A frozen $-$ B frozen, with its interval, at each rank tested on several ranks.} Each cell is in accuracy points, with its status (est.\,+: established positive; n.e.: not established; desc.: descriptive, a 95\,\% interval and no test). \Cref{tab:results} gives the rank-2 results of the other pairs. \dag: post hoc (\Cref{sec:prespec}).}\label{tab:contrasts}\\
\toprule
Task and model & rank & dominant band & minor band \\
\midrule
\endfirsthead
\toprule
Task and model & rank & dominant band & minor band \\
\midrule
\endhead
OpenBookQA, Qwen2.5-1.5B & 1 & +8.70 [+4.70, +12.73] est.\,+ & +8.40 [+4.35, +12.42] est.\,+ \\
 & 2 & +8.60 [+4.80, +12.55] est.\,+ & +10.10 [+6.25, +13.95] est.\,+ \\
 & 4 & +9.10 [+4.55, +14.08] est.\,+ & +8.42 [+4.60, +12.05] est.\,+ \\
 & 8 & +7.18 [+3.70, +10.75] est.\,+ & +7.25 [+3.70, +10.73] est.\,+ \\
 & 16 & replication: +1.88 [$-$1.55, +5.42] n.e.\dag{} & replication: +3.90 [+0.50, +7.38] est.\,+\dag{} \\
 & 32 & first experiment: +1.20 [$-$2.25, +4.67] n.e. \newline replication: +1.15 [$-$2.02, +4.40] n.e.\dag{} & first experiment: +2.62 [$-$1.30, +6.51] n.e. \newline replication: +1.70 [$-$1.40, +4.80] n.e.\dag{} \\
 & 64 & +1.38 [$-$1.82, +4.62] n.e. & +1.00 [$-$2.68, +4.80] n.e. \\
\midrule
format task, Llama-3.2-1B & 1 & +29.93 [+25.12, +34.28] est.\,+\dag{} & +28.64 [+24.37, +32.88] est.\,+\dag{} \\
 & 2 & +12.11 [+8.31, +16.06] est.\,+ & +20.40 [+16.96, +23.82] est.\,+ \\
 & 4 & +8.06 [+4.48, +11.64] est.\,+ & +9.83 [+5.90, +13.76] est.\,+ \\
 & 8 & +6.68 [+3.48, +9.85] est.\,+ & +4.78 [+1.69, +7.92] est.\,+ \\
 & 16 & first experiment: +3.42 [+0.86, +6.03] desc. \newline replication: +15.17 [+3.64, +36.50] est.\,+\dag{} (one diverged run, seed 14; \Cref{app:collapse}) & first experiment: +4.32 [+1.63, +6.86] desc. \newline replication: +3.40 [+0.33, +6.37] est.\,+\dag{} \\
 & 32 & first experiment: +0.24 [$-$2.85, +3.26] n.e. \newline replication: +2.59 [+0.04, +5.23] est.\,+\dag{} & first experiment: +1.40 [$-$1.61, +4.45] n.e. \newline replication: +2.34 [$-$1.14, +6.05] n.e.\dag{} \\
 & 64 & +1.38 [$-$1.71, +4.34] n.e. & +1.08 [$-$1.93, +3.97] n.e. \\
\midrule
OpenBookQA, Qwen2.5-7B & 2 & +11.47 [+7.52, +15.35] est.\,+ & +13.15 [+9.15, +17.10] est.\,+ \\
 & 16 & +2.10 [$-$1.28, +5.47] n.e.\dag{} & +6.27 [+2.57, +10.10] est.\,+\dag{} \\
 & 32 & $-$0.17 [$-$3.38, +3.08] n.e. & +2.23 [$-$1.68, +5.97] n.e. \\
\midrule
OpenBookQA, Mistral-7B & 2 & +12.62 [+8.90, +16.50] est.\,+ & +11.02 [+7.05, +15.05] est.\,+ \\
 & 32 & +4.38 [+1.30, +7.55] est.\,+ & +8.43 [+5.15, +11.90] est.\,+ \\
 & 64 & +0.60 [$-$2.25, +3.42] n.e.\dag{} & +4.88 [+1.70, +8.10] est.\,+\dag{} \\
\midrule
eight-task mixture, Qwen2.5-1.5B & 2 & +1.17 [+0.51, +1.82] est.\,+ & +1.95 [+1.27, +2.64] est.\,+ \\
 & 8 & +1.72 [+1.14, +2.30] est.\,+ & +2.19 [+1.62, +2.77] est.\,+ \\
 & 16 & +0.73 [+0.20, +1.26] est.\,+\dag{} & +1.98 [+1.45, +2.51] est.\,+\dag{} \\
 & 32 & +0.22 [$-$0.24, +0.68] n.e.\dag{} & +0.84 [+0.32, +1.35] est.\,+\dag{} \\
\bottomrule
\end{longtable}
\end{small}
\endblock

\block{g1}
\begin{small}
\begin{longtable}{p{4.6cm}rp{4.4cm}l}
\caption{\textbf{The matched free LoRA beats MiCA's configuration in each row.} Each row is the matched free LoRA minus $B$ frozen on the minor band, in accuracy points, with its interval and status (est.\,+: established positive; n.e.: not established). \iftmlr Table~S1 of the supplementary material\else\Cref{tab:tests}\fi\ gives the level and the protocol label of each test. \dag: post hoc (\Cref{sec:prespec}).}\label{tab:g1}\\
\toprule
Task and model & rank & difference [interval] & status \\
\midrule
\endfirsthead
\toprule
Task and model & rank & difference [interval] & status \\
\midrule
\endhead
OpenBookQA, Qwen2.5-1.5B & 4 & +9.45 [+5.43, +13.50] & est.\,+ \\
OpenBookQA, Qwen2.5-1.5B & 8 & +9.02 [+5.05, +13.08] & est.\,+ \\
OpenBookQA, Qwen2.5-1.5B & 16 & +5.85 [+2.37, +9.38] & est.\,+ \\
OpenBookQA, Qwen2.5-1.5B & 32 & +4.25 [+0.85, +7.90] & est.\,+ \\
format task, Llama-3.2-1B & 4 & +11.05 [+7.33, +14.96] & est.\,+ \\
format task, Llama-3.2-1B & 8 & +7.96 [+5.11, +10.91] & est.\,+ \\
format task, Llama-3.2-1B & 16 & +4.78 [+1.89, +7.90] & est.\,+ \\
format task, Llama-3.2-1B & 32 & +3.09 [+0.02, +6.03] & est.\,+ \\
OpenBookQA, Qwen2.5-7B & 2 & +14.70 [+10.58, +18.90] & est.\,+ \\
OpenBookQA, Qwen2.5-7B & 16\dag & +6.27 [+2.83, +9.67] & est.\,+ \\
OpenBookQA, Qwen2.5-7B & 32 & +3.75 [+0.35, +7.13] & est.\,+ \\
ARC-Challenge, Qwen2.5-7B & 2\dag & +3.73 [+1.28, +6.25] & est.\,+ \\
format task, Qwen2.5-7B & 2\dag & +11.03 [+8.08, +13.95] & est.\,+ \\
OpenBookQA, Mistral-7B & 2 & +11.80 [+7.77, +16.00] & est.\,+ \\
OpenBookQA, Mistral-7B & 32 & +3.90 [+0.25, +7.70] & est.\,+ \\
\bottomrule
\end{longtable}
\end{small}
\endblock

\def\wantedblock{arcvii}\immediate\write\secposfile{block appendix arcvii}
\def\wantedblock{g1}\immediate\write\secposfile{block tables/tab_contrasts g1}
\def\wantedblock{counts}\immediate\write\secposfile{block appendix counts}

\else
\section{Statistical details}
\label{app:stats}
\def\wantedblock{bootstrap}\immediate\write\secposfile{block appendix bootstrap}
\def\wantedblock{ratestability}\immediate\write\secposfile{block sec3_protocol ratestability}
\def\wantedblock{levels}\immediate\write\secposfile{block appendix levels}
\def\wantedblock{holm}\immediate\write\secposfile{block sec3_protocol holm}
\def\wantedblock{bounds}\immediate\write\secposfile{block appendix bounds}
\def\wantedblock{testsdetail}\immediate\write\secposfile{block sec3_protocol testsdetail}
\paragraph{The factor effect and the band effect in the OpenBookQA example.}
\def\wantedblock{fsvalues}\immediate\write\secposfile{block sec3_protocol fsvalues}
\def\wantedblock{fssub}\immediate\write\secposfile{block sec3_protocol fssub}
\def\wantedblock{fsalgebra}\immediate\write\secposfile{block sec3_protocol fsalgebra}
\def\wantedblock{fsweak}\immediate\write\secposfile{block sec3_protocol fsweak}
\def\wantedblock{prespecfixed}\immediate\write\secposfile{block sec3_protocol prespecfixed}
\def\wantedblock{prespeckinds}\immediate\write\secposfile{block sec3_protocol prespeckinds}
\def\wantedblock{posthoctable}\immediate\write\secposfile{block sec3_protocol posthoctable}
\def\wantedblock{minorband}\immediate\write\secposfile{block appendix minorband}
\paragraph{The rules for the learning rate.}
\label{app:rules}
\def\wantedblock{ratessum}\immediate\write\secposfile{block sec3_protocol ratessum}
\def\wantedblock{edge}\immediate\write\secposfile{block appendix edge}
\def\wantedblock{tie}\immediate\write\secposfile{block appendix tie}

\section{Every contrast, with its interval}
\label{app:contrasts}
\def\wantedblock{contrasts}\immediate\write\secposfile{block tables/tab_contrasts contrasts}
\def\wantedblock{arcvii}\immediate\write\secposfile{block appendix arcvii}
\fi

\section{Every pre-specified test}
\label{app:tests}
\def\wantedblock{testsshort}\immediate\write\secposfile{block appendix testsshort}
\begin{scriptsize}
\begin{longtable}{@{}rp{5.6cm}p{3.6cm}ll@{}}
\caption{\textbf{The tests the claims rely on: estimate, interval, level and status.} Every one of the 284 pre-specified tests, with the paired $t$ interval, is in Table~S1 of the supplementary material; the numbers of the rows are those of that table. Estimates are in accuracy points. est.\,+ / est.\,$-$: established positive / negative; n.e.: not established. Rows marked $^\circ$ are pre-specified tests that their protocol does not label primary or secondary. \dag: post hoc (\Cref{sec:prespec}).}\label{tab:tests}\\
\toprule
\# & contrast (task, model) & estimate [interval] & level & status \\
\endfirsthead
\toprule
\# & contrast (task, model) & estimate [interval] & level & status \\
\endhead
\midrule
\multicolumn{5}{@{}l}{\textit{OpenBookQA, ranks 2 to 64}} \\
1 & A frozen $-$ B frozen, dominant band, rank 2, 49 steps (OpenBookQA, Qwen2.5-1.5B) & +8.60 [+4.80, +12.55] & 0.975 & est.\,+ \\
2 & A frozen $-$ B frozen, minor band, rank 2, 49 steps (OpenBookQA, Qwen2.5-1.5B) & +10.10 [+6.25, +13.95] & 0.975 & est.\,+ \\
7 & F against $|S|$, rank 32, 49 steps (OpenBookQA, Qwen2.5-1.5B) & $F-S$ +6.95 [+3.65, +10.35] \newline $F+S$ $-$3.12 [$-$6.92, +0.57] & 0.975 & n.e. \\
8 & F against $|S|$, rank 64, 49 steps (OpenBookQA, Qwen2.5-1.5B) & $F-S$ +5.45 [+2.20, +8.75] \newline $F+S$ $-$3.08 [$-$6.38, +0.15] & 0.975 & n.e. \\
13 & F against $|S|$, rank 2, 49 steps (OpenBookQA, Qwen2.5-1.5B) & $F-S$ +10.70 [+6.50, +14.95] \newline $F+S$ +8.00 [+4.55, +11.43] & 0.975 & est.\,+ \\
14 & F against $|S|$, rank 16, 49 steps (OpenBookQA, Qwen2.5-1.5B) & $F-S$ +6.03 [+2.50, +9.60] \newline $F+S$ $-$0.47 [$-$4.10, +3.17] & 0.975 & n.e. \\
\multicolumn{5}{@{}l}{\textit{format task, ranks 2 to 64}} \\
15 & A frozen $-$ B frozen, dominant band, rank 2 (format task, Llama-3.2-1B) & +12.11 [+8.31, +16.06] & 0.975 & est.\,+ \\
16 & A frozen $-$ B frozen, minor band, rank 2 (format task, Llama-3.2-1B) & +20.40 [+16.96, +23.82] & 0.975 & est.\,+ \\
21 & F against $|S|$, rank 2 (format task, Llama-3.2-1B) & $F-S$ +19.16 [+15.51, +22.94] \newline $F+S$ +13.36 [+9.93, +16.90] & 0.975 & est.\,+ \\
22 & F against $|S|$, rank 16 (format task, Llama-3.2-1B) & $F-S$ +4.11 [+1.40, +7.02] \newline $F+S$ +3.62 [+0.73, +6.47] & 0.975 & est.\,+ \\
23 & F against $|S|$, rank 32 (format task, Llama-3.2-1B) & $F-S$ +1.57 [$-$1.06, +4.17] \newline $F+S$ +0.08 [$-$2.54, +2.65] & 0.975 & n.e. \\
24 & F against $|S|$, rank 64 (format task, Llama-3.2-1B) & $F-S$ +1.81 [$-$0.57, +4.26] \newline $F+S$ +0.65 [$-$1.81, +3.16] & 0.975 & n.e. \\
\multicolumn{5}{@{}l}{\textit{PiCa at its recipe}} \\
132 & budget-matched LoRA $-$ PiCa at its recipe, rank 256 (OpenBookQA, Qwen2.5-1.5B) & $-$6.05 [$-$9.87, $-$2.47] & 0.99375 & est.\,$-$ \\
133 & PiCa at its recipe $-$ B frozen on the dominant band at PiCa's budget, rank 256 (OpenBookQA, Qwen2.5-1.5B) & +0.53 [$-$2.30, +3.35] & 0.99375 & n.e. \\
134 & budget-matched LoRA $-$ PiCa at its recipe, rank 256 (format task, Llama-3.2-1B) & $-$0.79 [$-$4.23, +2.70] & 0.99375 & n.e. \\
135 & PiCa at its recipe $-$ B frozen on the dominant band at PiCa's budget, rank 256 (format task, Llama-3.2-1B) & +0.57 [$-$2.32, +3.29] & 0.99375 & n.e. \\
\multicolumn{5}{@{}l}{\textit{output overlap, rank 2}} \\
138 & rank correlation of overlap c$^2$ and accuracy, B frozen over 5 levels $\times$ 8 seeds (rank 2) (OpenBookQA, Qwen2.5-1.5B) & +0.902 [+0.289, +0.963] & 0.975 & est.\,+ \\
139 & rank correlation of overlap c$^2$ and accuracy, B frozen over 5 levels $\times$ 8 seeds (rank 2) (format task, Llama-3.2-1B) & +0.753 [+0.426, +0.889] & 0.975 & est.\,+ \\
\multicolumn{5}{@{}l}{\textit{output overlap, rank 8\dag}} \\
140 & rank correlation of overlap c$^2$ and accuracy, B frozen over 5 levels $\times$ 8 seeds (rank 8) (OpenBookQA, Qwen2.5-1.5B) & +0.524 [+0.052, +0.914] & 0.975 & est.\,+ \\
141 & rank correlation of overlap c$^2$ and accuracy, B frozen over 5 levels $\times$ 8 seeds (rank 8) (format task, Llama-3.2-1B) & +0.451 [+0.201, +0.711] & 0.975 & est.\,+ \\
\bottomrule
\end{longtable}
\end{scriptsize}

\subsection{What each experiment tests}
\label{app:experiments}
\def\wantedblock{experimentsptr}\immediate\write\secposfile{block appendix experimentsptr}

\iftmlr
\section{Robustness and side effects in detail}
\label{app:robust}
\def\wantedblock{robustintro}\immediate\write\secposfile{block appendix robustintro}
\def\wantedblock{controls}\immediate\write\secposfile{block sec8_robustness controls}\block{claim}
A gap of this size deserves suspicion, so \Cref{tab:robustdraft} tests five objections to it at rank 2.
Three of them leave the advantage in place, by $+3$ to $+19$ points, and a rotation of the frozen basis changes no configuration detectably.
The remaining one, a longer budget, passes on the format task and not on OpenBookQA (\Cref{fig:budget}).
\endblock
\block{twoagainstshort}
\paragraph{Two results go against parts of our reading, without changing the measured A $-$ B.}
The first tests the reference: a free LoRA trained with LoRA+~\citep{hayou2024loraplus}, which gives $B$ a larger learning rate than $A$, beats $A$ frozen on the dominant band of the format task (established, not Holm-robust).
This result weakens any statement that $A$ frozen matches a free LoRA.
The second tests whether the least-squares analysis of one linear layer by \citet{zhu2024asymmetry} predicts where A $-$ B is large: a predictor we derived from it is negatively correlated with A $-$ B over 40 configurations\iftmlr\else\ (\Cref{app:robust})\fi.
\endblock
\block{twoagainst}
\paragraph{Two results go against parts of our reading, without changing the measured A $-$ B.}
First, a free LoRA trained with LoRA+~\citep{hayou2024loraplus}, which gives $B$ a larger learning rate than $A$, beats $A$ frozen on the dominant band of the format task by $3.01$ points.
\iftmlr This weakens any statement that $A$ frozen matches a free LoRA. \fi
Second, a predictor that we derived from the least-squares analysis of one linear layer by \citet{zhu2024asymmetry} has a rank correlation of $-0.362$ with the measured A $-$ B over 40 configurations: it points the wrong way.
\iftmlr\else Both are detailed in \Cref{app:otherrobust}.\fi
\endblock
\block{robusttable}
\begin{table}[!htb]
\centering
\small
\caption{\textbf{The rank-2 advantage of $A$ frozen over $B$ frozen passes each test on the task where it was run, except a longer training budget on OpenBookQA.}
Results are A $-$ B in accuracy points on both bands.
``Holds'' means that A $-$ B stays established (\Cref{tab:statusterms}).
The last row instead compares each configuration with its rotated basis. Intervals and test levels: \iftmlr\Cref{tab:robustfull}\else\Cref{app:controls}\fi, and \Cref{app:x16} for the $16\times$ budget.}
\label{tab:robustdraft}
\begin{tabular}{>{\raggedright\arraybackslash}p{3.6cm}>{\raggedright\arraybackslash}p{4.4cm}>{\raggedright\arraybackslash}p{6.2cm}}
\toprule
Objection & Test & Result \\
\midrule
Rate-selection seeds & Eight new seeds, same rates & Holds on both tasks: $+9$ to $+19$ \\
$A$ frozen trains more parameters & $B$ frozen given 1.09 to 1.30$\times$ the parameters (format task) & Holds: $+3$ to $+14$ \\
The learning rate of $B$ frozen & Every rate of its grid and two above it & Holds at every rate (post hoc) \\
A short training budget & 16 times the budget (format task), 4 times (OpenBookQA) & Holds on the format task ($+10.83$, $+16.48$). On OpenBookQA, it holds only on the minor band at the best checkpoint, by $+8$; at the last step, both intervals include 0 (\Cref{sec:narrow}) \\
The basis inside the band & A random rotation of the frozen basis inside its band & Rotated and original bases do not differ detectably on the dominant band, the only one tested (at most 5 points) \\
\bottomrule
\end{tabular}
\end{table}

\endblock
\block{controlsshort}
\paragraph{The advantage also passes the controls on the training setup, except one.}
Could the training setup, and not the frozen factor, produce the advantage?
We changed one part of the training at a time, and A $-$ B stays established in every one of these controls except one: with one eighth of the training data, on the dominant band of OpenBookQA, we detect no difference (at most $6.70$ points).
GaLore~\citep{zhao2024galore} restricts the optimiser state, not the weights, to a low-rank projection; there too, at rank 2 on OpenBookQA, the input side is the cheaper side to restrict\iftmlr\else\ (\Cref{app:robust})\fi.
\endblock
\block{controls}
\paragraph{The advantage also passes the controls on the training setup.}
We changed the training setup one part at a time (\Cref{tab:controls}).
The advantage remains in all these controls except one: with one eighth of the training data, on the dominant band of OpenBookQA, we detect no difference (at most $6.70$ points).
GaLore~\citep{zhao2024galore} trains all weights but keeps its optimiser state in a low-rank projection.
A projection on the input side beats one on the output side by $+9.62$ points at rank 2 on OpenBookQA; at rank 32 we detect no difference.
\endblock
\block{against}
\paragraph{Results against the advantage of $A$ frozen.}
\label{par:floor}
The objection is that $B$ frozen needs more training time.
We test it with four times the training budget; each cell is A $-$ B in points.
\begin{center}\small
\begin{tabular}{@{}p{4.3cm}p{3cm}p{2.9cm}p{2.9cm}@{}}
\toprule
 & Scoring & Dominant band & Minor band \\
\midrule
OpenBookQA & best validation checkpoint & $+3.62$ & $+8.45$ \\
OpenBookQA (post hoc, descriptive) & last step & $-1.75$ $[-5.22, +1.88]$ & $+3.52$ $[-0.23, +7.38]$ \\
Format task (1,096 steps instead of 274) & last step & $+11.18$ & $+15.96$ \\
\bottomrule
\end{tabular}
\end{center}
\noindent At the best checkpoint, the dominant band is not established (at most $7.35$).
On OpenBookQA, both last-step intervals include 0.
With four times the updates (batch 8), A $-$ B stays established on both bands (\iftmlr below\else\Cref{app:stepcontrol}\fi).
The batch-8 runs answer only part of the objection: they use 196 steps, far fewer than the 1,187 steps of the generation protocol.
\iftmlr\Cref{app:oc} gives\else\Cref{app:robust} gives the results on ARC and the letter prompt, and \Cref{app:oc}\fi\ the answer-only generation protocol, where $B$ frozen learns and A $-$ B shrinks.
\endblock
\block{length}
\paragraph{Training length: the evidence on both sides.}
The gap shrinks with longer training in two places.
On the generation ladder, A $-$ B falls to between $+1.15$ and $+3.22$ on the two JSON targets after 1,187 steps (descriptive; \Cref{app:oc}).
On OpenBookQA with four times the tokens, $A$ frozen on the dominant band ends level with $B$ frozen on validation, after an earlier peak.
On the test set, A $-$ B on the dominant band falls from $+7.05$ at step 39 to $-1.75$ at step 196, and on the minor band from $+11.12$ to $+3.52$ (descriptive, post hoc; the intermediate steps and the validation values are given below).
The gap persists at MiCA's own recipe of 3,180 steps on the format task.
There, flipped MiCA (post hoc) beats MiCA by $+3.48$ points at rank 16, about as much as A $-$ B on the minor band at rank 16 in the replication ($+3.40$; descriptive comparison across experiments).
\endblock
\block{againstmore}
On ARC, the best checkpoint keeps the effect in one test of four.
Under the letter prompt (a transfer), A $-$ B is negative in 6 of 12 comparisons, six ranks from 2 to 64 and two bands (descriptive, post hoc; \Cref{app:letter}).
With 1,187 steps (not 49) on four generation targets, A $-$ B ranges from $-0.05$ to $+3.22$ (descriptive; \Cref{app:oc}).
\endblock
\block{costsmerge}
\paragraph{Freezing $A$ has side effects outside its task.}
We add the adapters of OpenBookQA, ARC-Easy and ARC-Challenge to the same model and score each task.
Against each task's own adapter, $A$ frozen on the minor band loses more accuracy than $B$ frozen, by $8.59$ points at rank 2 and $4.28$ at rank 32.
On the dominant band, we detect no difference.
Most of the loss falls on ARC-Easy.
$A$ frozen also forgets more: at the selected learning rates, it forgets more of HellaSwag~\citep{zellers2019hellaswag} than $B$ frozen in 4 of 16 tests (four task--model pairs, ranks 2 and 32, two bands), and less in none.
\endblock
\block{costs}
\paragraph{Forgetting.}
We measure forgetting as the drop in HellaSwag accuracy from the base model, in points, and as the rise of WikiText~\citep{merity2017wikitext} perplexity in \emph{log-percent}, $100 \ln(\text{adapted perplexity} / \text{base perplexity})$.
Intruder dimensions~\citep{shuttleworth2024illusion} are leading singular vectors of the adapted weight that are far from every singular vector of the base weight.
We do not detect that their count tracks forgetting\iftmlr\else\ (\Cref{app:robust})\fi.
Adapters trained on OpenBookQA lose accuracy on ARC-Easy in every configuration except $B$ frozen on the dominant band, and most with $A$ frozen on the minor band (\Cref{app:transfer}).
$A$ frozen also learns its task better, so we lowered its learning rate to match the accuracy of $B$ frozen.
\iftmlr\else\Cref{app:robust} gives these results and the forgetting at rank 16.\fi
\endblock
\block{checkslimits}
\endblock

\def\wantedblock{against}\immediate\write\secposfile{block sec8_robustness against}
\def\wantedblock{length}\immediate\write\secposfile{block sec8_robustness length}
\def\wantedblock{sixteen}\immediate\write\secposfile{block appendix sixteen}
\def\wantedblock{budgetfig}\immediate\write\secposfile{block appendix budgetfig}
\def\wantedblock{costs}\immediate\write\secposfile{block sec8_robustness costs}
\def\wantedblock{robustsum}\immediate\write\secposfile{block appendix robustsum}
\def\wantedblock{letterformat}\immediate\write\secposfile{block appendix letterformat}
\def\wantedblock{forgetrktab}\immediate\write\secposfile{block appendix forgetrktab}
\section{Controls}
\label{app:controls}
\def\wantedblock{controlsintro}\immediate\write\secposfile{block appendix controlsintro}
\def\wantedblock{controlstable}\immediate\write\secposfile{block appendix controlstable}

\section{Every setting and rank}
\label{app:rankexp}
\begin{table}[!t]
\centering
\footnotesize
\caption{\textbf{A $-$ B by setting and rank: A $-$ B is established on both bands of both main pairs up to rank 8, and mostly not above.} A $-$ B is the accuracy of $A$ frozen minus that of $B$ frozen on the same band, in points. Bold: established, a test whose interval excludes zero (\Cref{tab:contrasts,tab:tests}). Verdict: the test of $F$ against $|S|$ (\Cref{sec:tests}): factor, band, open (neither larger) or no test. The notes under the table explain the note column.}\label{tab:results}
\begin{tabular}{@{}llrrl>{\raggedright\arraybackslash}p{2.9cm}@{}}
\toprule
& & \multicolumn{2}{c}{A $-$ B} & & \\
\cmidrule(lr){3-4}
Task, model & rank & dominant & minor & verdict & note \\
\midrule
\multicolumn{6}{@{}l}{\textit{Main pair: OpenBookQA with Qwen2.5-1.5B}} \\
 & 1 & \textbf{+8.70} & \textbf{+8.40} & factor &  \\
 & 2 & \textbf{+8.60} & \textbf{+10.10} & factor &  \\
 & 4 & \textbf{+9.10} & \textbf{+8.42} & factor &  \\
 & 8 & \textbf{+7.18} & \textbf{+7.25} & factor &  \\
 & 16 & +1.88 & \textbf{+3.90} & open & replication, after the results \\
 & 32 & +1.20 & +2.62 & open & first experiment \\
 & 32 & +1.15 & +1.70 & band & replication, after the results \\
 & 64 & +1.38 & +1.00 & open &  \\
\midrule
\multicolumn{6}{@{}l}{\textit{Main pair: format task with Llama-3.2-1B}} \\
 & 1 & \textbf{+29.93} & \textbf{+28.64} & factor & after the results \\
 & 2 & \textbf{+12.11} & \textbf{+20.40} & factor &  \\
 & 4 & \textbf{+8.06} & \textbf{+9.83} & factor &  \\
 & 8 & \textbf{+6.68} & \textbf{+4.78} & factor &  \\
 & 16 & +3.42 & +4.32 & factor & first experiment, A $-$ B not tested \\
 & 16 & \textbf{+15.17} & \textbf{+3.40} & factor & replication, after the results, one diverged run \\
 & 32 & +0.24 & +1.40 & open & first experiment \\
 & 32 & \textbf{+2.59} & +2.34 & open & replication, after the results \\
 & 64 & +1.38 & +1.08 & open &  \\
 & 2 & \multicolumn{2}{l}{\textbf{+11.07}, one contrast over both bands} & no test & random bases instead of a band \\
\midrule
\multicolumn{6}{@{}l}{\textit{7B models: OpenBookQA}} \\
OpenBookQA, Qwen2.5-7B & 2 & \textbf{+11.47} & \textbf{+13.15} & factor &  \\
 & 16 & +2.10 & \textbf{+6.27} & open & after the results \\
 & 32 & $-$0.17 & +2.23 & open &  \\
OpenBookQA, Mistral-7B & 2 & \textbf{+12.62} & \textbf{+11.02} & factor &  \\
 & 32 & \textbf{+4.38} & \textbf{+8.43} & factor &  \\
 & 64 & +0.60 & \textbf{+4.88} & open & after the results \\
\midrule
\multicolumn{6}{@{}l}{\textit{Eight-task mixture with Qwen2.5-1.5B}} \\
 & 2 & \textbf{+1.17} & \textbf{+1.95} & factor &  \\
 & 8 & \textbf{+1.72} & \textbf{+2.19} & no test &  \\
 & 16 & \textbf{+0.73} & \textbf{+1.98} & no test & after the results \\
 & 32 & +0.22 & \textbf{+0.84} & no test & after the results \\
\midrule
\multicolumn{6}{@{}l}{\textit{Other tasks and models, rank 2}} \\
ARC-Easy, Qwen2.5-1.5B & 2 & +0.31 & \textbf{+1.63} & open &  \\
ARC-Challenge, Qwen2.5-1.5B & 2 & +0.94 & \textbf{+3.18} & open &  \\
OpenBookQA, Qwen2.5-1.5B-Instruct & 2 & \textbf{+8.65} & +3.97 & factor &  \\
format task, Qwen2.5-7B & 2 & \textbf{+11.56} & \textbf{+11.07} & factor & after the results \\
ARC-Challenge, Qwen2.5-7B & 2 & +2.89 & \textbf{+4.66} & no test & A $-$ B not tested on the dominant band, after the results \\
OpenBookQA, Qwen3-1.7B-Base & 2 & \textbf{+8.52} & \textbf{+9.60} & factor & \\
format task, Qwen2.5-1.5B & 2 & \textbf{+14.23} & \textbf{+12.70} & factor & task and model crossed \\
OpenBookQA, Llama-3.2-1B & 2 & \textbf{+8.30} & \textbf{+5.37} & factor & task and model crossed \\
\midrule
OpenBookQA, Llama-3.1-8B & 8 & +2.95 & \textbf{+7.02} & open & \\
 & 16 & +0.60 & +2.45 & open & \\
format task, Llama-3.1-8B & 8 & \textbf{+4.38} & +1.69 & open & \\
GSM8K, Llama-3.2-1B & 2 to 32 & \multicolumn{3}{l}{in \Cref{tab:gsm}} & \\
\bottomrule
\end{tabular}
\par\smallskip\begin{minipage}{\linewidth}\raggedright
\emph{First experiment}, \emph{replication}: the two runs of ranks 16 and 32 on the main pairs; the replication has new seeds and a finer grid of rates. On OpenBookQA, the first experiment at rank 16 is reported in \Cref{tab:reversal} only.\\
\emph{After the results}: see \Cref{sec:prespec}. \emph{One diverged run}: it inflates the dominant-band value (\Cref{app:collapse}).\\
\emph{Random bases}: 24 random orthonormal bases per factor instead of a band, seeds 2 to 9, fixed learning rates, one contrast over both bands.
\end{minipage}
\end{table}

\begin{table}[!htb]
\centering
\small
\caption{\textbf{On OpenBookQA with Qwen2.5-1.5B above rank 8, the observed band effect $|S|$ is larger than the factor effect $F$ in all five comparisons; one of the five tests finds it.}
$F$ and $S$ are observed values in accuracy points (\Cref{sec:tests}), and only the verdict is a test.
$S$ is negative: the dominant band is the better one.
Verdict and note: as in \Cref{tab:results}.}
\label{tab:reversal}
\begin{tabular}{@{}rlrrll@{}}
\toprule
rank & run & $F$ & $S$ & verdict & note \\
\midrule
16 & first exp. & $+2.78$ & $-3.25$ & open & \\
16 & replication & $+2.89$ & $-3.44$ & open & after the results \\
32 & first exp. & $+1.91$ & $-5.04$ & open & \\
32 & replication & $+1.43$ & $-5.60$ & band & after the results \\
64 & & $+1.19$ & $-4.26$ & open & \\
\bottomrule
\end{tabular}
\end{table}

\def\wantedblock{ranksum}\immediate\write\secposfile{block appendix ranksum}
\def\wantedblock{ocsum}\immediate\write\secposfile{block appendix ocsum}
\def\wantedblock{oclevelstab}\immediate\write\secposfile{block sec4_results oclevelstab}
\def\wantedblock{gsmtable}\immediate\write\secposfile{block sec4_results gsmtable}
\def\wantedblock{gsm8b}\immediate\write\secposfile{block sec4_results gsm8b}
\def\wantedblock{rankxvimore}\immediate\write\secposfile{block appendix rankxvimore}
\def\wantedblock{latecvr}\immediate\write\secposfile{block appendix latecvr}

\section{Where A \texorpdfstring{$-$}{-} B is not established at ranks 16 to 64}
\label{app:notest}
\def\wantedblock{notestsum}\immediate\write\secposfile{block appendix notestsum}

\section{Transfer to a neighbouring task}
\label{app:transfer}
\def\wantedblock{transfer}\immediate\write\secposfile{block appendix transfer}

\section{How the OpenBookQA result depends on the protocol}
\label{app:supported}
\def\wantedblock{resultsprot}\immediate\write\secposfile{block sec1_intro resultsprot}\block{figdesign}
\begin{figure}[t]
\centering
\paperfig[width=\linewidth]{figures/fig1_design.pdf}
\caption{\textbf{Freezing $A$ instead of $B$ raises accuracy on both bands at ranks 1, 2, 4 and 8, on both main pairs.}
\iftmlr Left: a LoRA layer; a frozen factor holds singular vectors of the pretrained weight $W_0$.
\else Left: in a LoRA layer, the input factor $A$ reads the input and the output factor $B$ writes the output.
A frozen $A$ is set to rows of $V^\top$ and a frozen $B$ to columns of $U$, on the dominant or the minor band of the pretrained weight
$W_0 = U \operatorname{diag}(\sigma) V^\top$.
\fi
Middle: the four frozen configurations at rank 2 on the format task with Llama-3.2-1B, with their test accuracy (\%, mean of eight seeds); the base model scores $0\,\%$ and the free LoRA $80.07\,\%$.
Right: A $-$ B in points against the rank, with the interval of each test, then GSM8K (Llama-3.2-1B) and the mixture (Qwen2.5-1.5B) at rank 8.
In every figure, circles are the dominant band and squares the minor band; filled markers are established.
Rank 1 of the format task was added after related results were seen.}
\label{fig:design}
\end{figure}
\endblock
\block{opening}
Frozen-subspace variants of LoRA~\citep{hu2021lora} freeze one of the two adapter factors on a \emph{band}, a set of $r$ singular directions of the pretrained weight (\Cref{sec:configs}).
MiCA~\citep{rudiger2026mica} freezes the output factor $B$ on the minor band (the smallest singular values), and PiCa~\citep{hwang2026pica}, in our notation, freezes the input factor $A$ on the dominant band (the largest).
Both are motivated, and usually compared, by their band.

But the band is not the only thing that separates them: they also freeze different factors.
Take the format task of \Cref{fig:design} at rank 2: under our protocol, the two configurations these methods freeze are 19 points apart ($77.08\,\%$ against $57.92\,\%$).
Changing only the band moves $A$ frozen by 1 point and $B$ frozen by 7; changing only the factor moves accuracy by 12 to 20 points.
A reader who credits the 19 points to the band is mostly measuring the factor.
Our question is therefore: \emph{at low rank, does the frozen factor or the band matter more?}
The published gains leave it open, because each method changes both choices at once.

We answer with a $2 \times 2$ design: $A$ or $B$ frozen, on the dominant or the minor band (\Cref{fig:design}).
We compare the four configurations with each other and with a \emph{free LoRA}, whose two factors both train; the \emph{matched free LoRA} has about the same number of trainable parameters as $B$ frozen (\Cref{sec:budgets}).
Throughout, A $-$ B is the accuracy of $A$ frozen minus that of $B$ frozen, on the same band.
We ran Qwen, Llama and Mistral models (\Cref{sec:tasks}) on three multiple-choice tasks, an eight-task mixture, GSM8K~\citep{cobbe2021gsm8k} and a \emph{format task}, at ranks 1 to 64, and PiCa at rank 16 and at rank 256, a rank used in its paper (\Cref{sec:methods}).
The format task is to rewrite the named entities of a sentence into a fixed one-line JSON structure that the base models never produce.
Our main results come from two \emph{main pairs} of a task and a model: OpenBookQA with
Qwen2.5-1.5B, and the format task with Llama-3.2-1B.
On them, every configuration has the same short training budget (GSM8K trains for two full epochs) and a learning rate chosen by rules written before the runs (\Cref{sec:budgets,sec:rates}).
The format task is our primary evidence.
We formed the hypothesis on OpenBookQA, where $B$ frozen barely learns at rank 2, so we read that pair with more caution (\Crefresults).
\endblock
\block{statuses}
\paragraph{How to read the results.}
All differences are in accuracy points.
A difference is \emph{established} when its paired bootstrap interval over seeds and test items excludes zero (\Cref{app:stats} gives the levels).
Otherwise we write ``we detect no difference (at most $X$ points)'', where $X$ is the largest effect in the expected direction that the interval still allows.
\endblock
\block{results}
\paragraph{What holds.}
At ranks 1 to 8 on both main pairs, freezing $A$ instead of $B$ gains accuracy on both bands, and gains more than changing the band.
After a Holm correction, the gain survives in 15 of its 16 tests, and its lead over the band survives at ranks 1, 2 and 4 (\Cref{sec:prespec}).
\iftmlr\else That is, $A$ frozen on either band beats $B$ frozen on the other band.
\fi
The gain is largest on the format task, whose base model scores $0\,\%$: $+12.11$ and $+20.40$ points at rank 2.
At rank 8, the default rank of the PEFT library, A $-$ B is still about $+7$ points on both bands of OpenBookQA, and $+6.68$ and $+4.78$ on the format task (\Cref{fig:design}).
On the format task, it persists with sixteen times the training budget at rank 2 and with four times at rank 16, so on this task slower training does not explain it (\Cref{tab:robustdraft,app:cvr16}).
At rank 2, it repeats on OpenBookQA with Qwen2.5-7B and Mistral-7B, and, outside the correction, on GSM8K with Llama-3.2-1B, trained for two full epochs (\Cref{sec:rank2}).
\paragraph{Where the advantage fades.}
On OpenBookQA, the gap closes with four times the training budget, mostly because $A$ frozen declines\iftmlr\ (\Cref{fig:budget})\fi.
\endblock
\block{resultsprot}
At the standard 49 steps, $B$ frozen stays near the base model on OpenBookQA (\Cref{sec:narrow}).
With four times the updates, $A$ frozen still beats $B$ frozen on both bands.
With four times the training budget, we detect no difference on the dominant band at the best checkpoint.
At the last step of those runs, in an analysis decided after seeing the results, the intervals of both bands include zero (\Cref{par:floor}).
When the same questions are trained as an answer-only generation task, we observe that A $-$ B is small on both bands, although its interval excludes zero on the minor band (\Cref{app:oc}).
Trained and scored with a letter prompt, A $-$ B is established on both bands at 49 steps and on neither band at 893 steps (\Cref{tab:oqaprot}).
The OpenBookQA result depends on the protocol.
\endblock
\block{resultsrest}
On ARC and with an 8B model it holds on one band only, in the eight-task mixture it is small, and above rank 8 it holds mostly on the minor band.
At MiCA's own rank 16, flipping its frozen factor wins on the format task (not Holm-robust), is inconclusive on OpenBookQA and changes almost nothing with a 7B model.
PiCa beats the matched free LoRA only at its published learning rate (\Cref{sec:methods}).
Freezing $A$ also has a side effect: when adapters are merged on the minor band, $A$ frozen loses more accuracy on its own task than $B$ frozen (\Cref{sec:robustness}).
\endblock
\block{resultsdetail}
\iftmlr\else\paragraph{Further results.}\fi
At learning rates we select per configuration, we detect no difference between PiCa and the matched free LoRA.
\iftmlr
At PiCa's published learning rate, PiCa beats the matched free LoRA in two of the three settings where the free LoRA learns the task (\Cref{sec:methods}).
\else
At PiCa's published learning rate, PiCa beats the matched free LoRA in two of the three settings where the free LoRA clears the floor, with a recipe that gives the free LoRA 15 or 150 steps (\Cref{sec:methods}).
\fi
Freezing $A$ also forgets more than freezing $B$ in some tests (\Cref{sec:robustness}).
Three of the analyses decided after seeing the results stand out.
A replication at rank 32 finds the band more important on OpenBookQA.
With Mistral-7B, the effect holds on the minor band up to rank 64.
At matched task accuracy, $A$ frozen forgets less on WikiText than $B$ frozen.
Our results are compatible with a frozen $B$ working when its subspace overlaps the output directions that a free LoRA learns
(\Cref{sec:mechanism}).
\endblock
\block{contrib}
\paragraph{Contributions.}
Our contributions are threefold, with one open hypothesis:
(1) \textbf{Controlled comparison:} a $2 \times 2$ audit that separates the two choices of MiCA and PiCa, with tests written before the runs (\Cref{sec:prespec} names the exceptions).
(2) \textbf{Factor against band:} a measure of the two effects at ranks 1 to 64 on the main pairs, and at three ranks on two 7B models.
(3) \textbf{Published methods:} a comparison of MiCA's and PiCa's configurations with the matched free LoRA, at selected and at published learning rates.
We also propose \textbf{a hypothesis that one test could not resolve:} the output directions the update needs move during training more than its input directions, so a $B$ frozen before training would cost more than a frozen $A$ (\Cref{sec:discussion}).

\paragraph{Scope.}
Why low rank?
Rank 8 is the default of the PEFT library~\citep{mangrulkar2022peft}, and VeRA~\citep{kopiczko2024vera} and LoRA-XS~\citep{balazy2024loraxs} aim at even fewer trainable parameters.
Low rank is also where a frozen factor constrains the update most, so its cost is measured most cleanly there.
The claim covers ranks 1 to 8, short training budgets and the two main pairs, with replications at rank 2 on two 7B models and on GSM8K.
Beyond, we report what we measured and extend the claim no further (\Cref{sec:discussion} lists the limits).
The paper is an audit of existing methods and proposes no new one.
\endblock
\block{guide}
\endblock

\def\wantedblock{oqatable}\immediate\write\secposfile{block sec4_results oqatable}

\section{The settings where the main result holds}
\label{app:holds}
\def\wantedblock{supporteddetail}\immediate\write\secposfile{block sec9_discussion supporteddetail}\block{opening}
\iftmlr\else This section says what \Cref{sec:rank2,sec:methods,sec:robustness} change for someone comparing frozen-subspace variants of LoRA.\fi
\endblock
\block{supported}
\iftmlr\paragraph{What the results change for someone who compares variants.}
\else\paragraph{A comparison of MiCA and PiCa changes two things at once.}
\fi
\iftmlr\else\Cref{sec:rank2} measures the effect of the frozen factor and that of the band separately.\fi
\endblock
\block{supporteddetail}
The advantage holds on both main pairs, OpenBookQA with Qwen2.5-1.5B and the format task with Llama-3.2-1B (\Cref{sec:rank2}).
It is established on both bands at ranks 1 to 8 (the format task at rank 1, post hoc).
On OpenBookQA, this is at the standard 49 steps (\Cref{app:data}), where $B$ frozen stays near the base model.
The gap is small under the answer-only protocol (descriptive; \Cref{app:oc}) but not with four times the updates (batch 8); with four times the tokens it is not resolved on the dominant band (\Cref{par:floor}).
At rank 2 the same holds on OpenBookQA with the instruction-tuned Qwen2.5-1.5B, on the dominant band only (the minor band is not established; \Cref{sec:rank2}).
It also holds on two 7B models: on OpenBookQA and the format task (post hoc) with Qwen2.5-7B, and on OpenBookQA with Mistral-7B (\Cref{sec:rank2}).
The effect survives the training changes made one part at a time in \Cref{tab:controls}, with one exception: with one eighth of the data, the dominant band of OpenBookQA is not established.
\iftmlr\else These controls are not independent confirmations, because they reuse the seeds and test questions of the main experiments. \fi
On the format task, the effect also holds with random frozen bases instead of spectral bands (24 bases per factor, eight seeds).
On ARC with Qwen2.5-1.5B it holds on the minor band only, and the effect is small.
\endblock
\block{methods}
\paragraph{A hypothesis for why freezing $A$ costs little.}
Our best explanation so far is that the output directions the update needs move during training, and its input directions move less.
A trained $B$ can follow the output directions and a $B$ frozen before training cannot, so freezing $B$ would cost more than freezing $A$.
Two descriptive analyses bear on this hypothesis without testing it.
In the overlap analysis, at ranks 2, 8 and 16, $B$ frozen does better the more its basis overlaps the output directions that a free LoRA learns on the same data.
This analysis is an oracle, not a method.
The gradient analysis, decided after seeing the results, compares the leading gradient subspaces at step 0 and at the last step, in one free LoRA per main pair.
Their overlap above chance is rescaled so that it is $0$ for random subspaces and $1$ for identical ones.
At rank 2 on OpenBookQA, it is $0.415$ on the input side and $0.108$ on the output side, so the output side moved more.
The output side has the lower overlap in $88\,\%$ of the $196$ adapted modules, as on the format task.

One result does not fit this hypothesis: $A$ frozen at a random initialisation, not chosen from the input directions of the update, still beats $B$ frozen (\Cref{sec:background}).
On OpenBookQA, we also detect no rise in the accuracy of $A$ frozen with its overlap with the learned input directions, unlike on the format task.

We then tested the input side of the hypothesis on the format task at rank 2, and the test could not resolve it (\Cref{app:ig}).
Freezing $A$ on the input directions that a trained adapter ends with beats the band by $+3.85$ points.
Freezing it on the input directions of the step-0 gradient recovers $+0.96$ of these $3.85$ points, which is not established.
One control check of this experiment failed\iftmlr\ (\Cref{app:overlap,app:mech,app:gradsides})\else\ (\Cref{app:ig})\fi.
\endblock
\block{open}
\iftmlr\else
\paragraph{What the results leave open.}
We do not claim the advantage above the ranks we tested (see the limitations below).
At the higher ranks tested (16 and above, which include the rank 16 MiCA uses for Llama-2), the effect is established on some bands only.
At PiCa's own rank 256 and recipe, the advantage of $A$ frozen is not detected.
\fi
\endblock
\block{opendetail}
At rank 32 the band matters more on OpenBookQA with Qwen2.5-1.5B (post hoc).
With Mistral-7B, the minor band keeps the effect up to rank 64 (post hoc).
In a LoRA-XS-style design (both outer factors frozen, a $16 \times 16$ core trained), the output basis limits the adapter on the format task with Llama-3.2-1B at rank 16 (one test; not established on OpenBookQA).
\endblock
\block{costsdetail}
In descriptive transfer results (post hoc), adapters trained on OpenBookQA lose accuracy on ARC-Easy for every configuration except $B$
frozen on the dominant band.
The loss is largest with $A$ frozen on the minor band.
In the other directions, $A$ frozen transfers as well or better (\Cref{app:transfer}).
\endblock
\block{costs}
\iftmlr\else\paragraph{Side effects on other tasks.}\fi
\endblock
\block{costsend}
Together with the forgetting and merging results of \Cref{sec:robustness}, these transfer results suggest a check.
A user who freezes $A$ for its accuracy on the target task should look at what it does to the other tasks they care about.
\endblock
\block{practical}
\iftmlr\else\paragraph{What we report for each comparison.}\fi
We report or fix five things:
\begin{enumerate}[nosep]
\item the frozen factor, not only the band;
\item the same band with the other factor frozen, which separates the factor from the band;
\item a learning rate selected for each configuration separately, by a rule fixed before the runs;
\item the exact trainable counts, with a matched free LoRA of about the same count, and their ratio;
\item a test at the method's own ranks: a gain at rank 2 need not hold at rank 32 (\Cref{sec:rank}).
\end{enumerate}
A frozen factor needs no gradient and no optimiser state.
Its price is accuracy against the matched free LoRA.
At rank 2, we observe that $A$ frozen pays $1.72$ to $2.99$ points across both bands of the two main pairs (\Cref{fig:design,fig:budget}; the smallest gap is $54.97\,\%$ against $53.25\,\%$ on OpenBookQA).
$B$ frozen, below $A$ frozen there, pays more.
\iftmlr\else $A$ frozen also forgets more of HellaSwag in 4 of 16 tests (\Cref{app:robust}).\fi
The two frozen configurations cost the same memory and time per step, within $0.122$ GiB and $0.007$ seconds (\Cref{app:memtime}), so the choice rests on accuracy and side effects (\Cref{sec:robustness}).
\endblock
\block{practicaldetail}
The price is lower accuracy on some neighbouring tasks and more forgetting of HellaSwag at the selected learning rates.
At matched task accuracy on the format task (dominant band, rank 2), however, $A$ frozen forgets less on WikiText (\iftmlr\Cref{app:robust}\else\Cref{app:otherrobust}\fi); at its selected learning rate at rank 16 on OpenBookQA it forgets far more (\Cref{tab:forgetrk}).
\endblock
\block{limitations}
\paragraph{Limitations.}
\begin{itemize}[nosep]
\item \emph{Training budget.} Training is short on the main pairs (\Cref{sec:budgets}); GSM8K trains for two full epochs.
On OpenBookQA, the gap closes with four times the training budget, mostly because $A$ frozen declines; we have not tested why.
With a letter prompt, the gap holds at 49 steps and not at 893, and it reverses when adapters trained under the main protocol are scored with that prompt, which we have not explained (\Cref{sec:rank2}).
On four generation targets trained for 1,187 steps, A $-$ B lies between $-0.05$ and $+3.22$ points, without tests (\Cref{app:robust}).
Longer training was run on the two main pairs only.
\item \emph{Ranks.} At rank 8, the comparison with the band and A $-$ B on the minor band of the format task are not Holm-robust.
At rank 16 and above, A $-$ B is established on some bands only, and on OpenBookQA with Qwen2.5-1.5B the observed band effect is larger than the factor effect (\Cref{sec:rank}).
At MiCA's ranks on OpenBookQA, its band matters, and flipping its frozen factor is inconclusive; with Qwen2.5-7B, the flip changes at most $1.05$ points (\Cref{sec:methods}).
\item \emph{Where the advantage is small.} It is small in the eight-task mixture and on ARC, where it holds on the minor band only (\Crefresults).
Scored at the best checkpoint, it remains in one of four tests on ARC (\Cref{app:robust}).
$F$ is not found above $|S|$ on ARC, with the letter prompt or on GSM8K at rank 8.
\item \emph{Published recipes.} PiCa beats the matched free LoRA only at its published learning rate, in two of three settings, and at PiCa's rank 256 and recipe we detect no advantage of $A$ frozen (\Cref{sec:methods}).
\item \emph{Side effects.} On the minor band, merging costs $A$ frozen more than $B$ frozen, and it forgets more of HellaSwag in 4 of 16 tests (\Cref{sec:robustness}).
\item \emph{Models and tasks.} We used three model families, with an 8B model as the largest, one synthetic generation task, the format task, and one reasoning task, GSM8K, with one model.
With the 8B model, A $-$ B is established on one band of each task only.
Flipped MiCA is tested with Qwen2.5-1.5B and Llama-3.2-1B on the two main tasks, and with Qwen2.5-7B on the eight-task mixture only.
\item \emph{Statistics.} A comparison where we detect no difference may still hide a real effect of a few points.
Our intervals reflect the data order, the free LoRA's initial $A$ and the test items, not the choice of learning rate, task or model.
The controls reuse the seeds and test items of the main experiments, so they do not confirm the result independently, and most run at rank 2 only.
Some controls use fewer seeds, and some rate selections ended on ties or at the grid edge (\Cref{tab:ties}).
Our tests were fixed in our own version history, not in an external registry.
\end{itemize}
\endblock
\block{conclusion}
\paragraph{Conclusion.}
Frozen-subspace variants of LoRA differ in the factor they freeze, not only in the band.
On our main pairs, with short training budgets, freezing $A$ instead of $B$ raises accuracy at ranks 1 to 8, and the Holm correction keeps this gain in 15 of 16 tests.
Which factor is frozen also matters more than which band at each of these ranks, and the correction keeps that at ranks 1, 2 and 4, not at rank 8.
The gain repeats at rank 2 with two 7B models and, outside the correction, on GSM8K; with longer training it persists on the format task and closes at the last step on OpenBookQA.
At low rank, a comparison of such methods should fix the frozen factor before crediting a difference to the band.
\iftmlr\paragraph{Code and data.}
Code, protocols and the logs of every run are in the supplement and will be released.\fi
\endblock
\block{memtime}
We measured one training step of $A$ frozen, of $B$ frozen and of a LoRA with both factors trained at the same rank, on the dominant band, with sequences of 384 tokens and a batch of 2 (\Cref{tab:memtime}).
Each value is the median of three repetitions on one 20\,GB GPU, in bfloat16 with gradient checkpointing; the measure is descriptive, without a test.
At a given rank, $A$ frozen and $B$ frozen differ by at most $0.063$ GiB of peak memory with Qwen2.5-1.5B and $0.122$ GiB with Qwen2.5-7B, and by at most $0.007$ seconds per step.
The rows with both factors trained are at the same rank as the frozen rows: they train as many parameters as the two frozen configurations together, so they are not the matched free LoRA of the body, whose rank is set by the parameter count.
With sequences of 2,048 tokens, Qwen2.5-7B ran out of memory in every configuration; the minor band was not measured.

\begin{table}[h!]
\centering\small
\caption{\textbf{$A$ frozen and $B$ frozen cost the same memory and time per step.}
Peak allocated memory (model weights included) and time of one training step, medians of three repetitions, 384 tokens, batch of 2, dominant band. ``Both trained'': both factors trained at the same rank.}
\label{tab:memtime}
\begin{tabular}{llrrrr}
\toprule
Model & Configuration & Rank & Parameters & Peak memory (GiB) & Step time (s) \\
\midrule
Qwen2.5-1.5B & $A$ frozen & 8 & 5{,}160{,}960 & 4.558 & 0.2463 \\
 &  & 32 & 20{,}643{,}840 & 4.833 & 0.2570 \\
 &  & 64 & 41{,}287{,}680 & 5.071 & 0.2778 \\
\addlinespace
 & $B$ frozen & 8 & 4{,}071{,}424 & 4.550 & 0.2497 \\
 &  & 32 & 16{,}285{,}696 & 4.800 & 0.2589 \\
 &  & 64 & 32{,}571{,}392 & 5.008 & 0.2772 \\
\addlinespace
 & both trained & 8 & 9{,}232{,}384 & 4.588 & 0.2569 \\
 &  & 32 & 36{,}929{,}536 & 4.954 & 0.2675 \\
 &  & 64 & 73{,}859{,}072 & 5.314 & 0.2965 \\
\addlinespace
Qwen2.5-7B & $A$ frozen & 8 & 11{,}124{,}736 & 16.080 & 0.8174 \\
 &  & 32 & 44{,}498{,}944 & 16.559 & 0.8353 \\
 &  & 64 & 88{,}997{,}888 & 17.187 & 0.8714 \\
\addlinespace
 & $B$ frozen & 8 & 9{,}060{,}352 & 16.065 & 0.8148 \\
 &  & 32 & 36{,}241{,}408 & 16.497 & 0.8345 \\
 &  & 64 & 72{,}482{,}816 & 17.065 & 0.8644 \\
\addlinespace
 & both trained & 8 & 20{,}185{,}088 & 16.148 & 0.8311 \\
 &  & 32 & 80{,}740{,}352 & 16.829 & 0.8595 \\
 &  & 64 & 161{,}480{,}704 & 17.727 & 0.9066 \\
\bottomrule
\end{tabular}
\end{table}
\endblock

\def\wantedblock{opendetail}\immediate\write\secposfile{block sec9_discussion opendetail}
\def\wantedblock{costsdetail}\immediate\write\secposfile{block sec9_discussion costsdetail}
\def\wantedblock{costsend}\immediate\write\secposfile{block sec9_discussion costsend}
\def\wantedblock{resultsdetail}\immediate\write\secposfile{block sec1_intro resultsdetail}
\def\wantedblock{practicaldetail}\immediate\write\secposfile{block sec9_discussion practicaldetail}
\else
\section{Robustness and side effects in detail}
\label{app:robust}
\def\wantedblock{twoagainst}\immediate\write\secposfile{block sec8_robustness twoagainst}
\subsection{Controls}
\label{app:controls}
\def\wantedblock{controlstable}\immediate\write\secposfile{block appendix controlstable}
\subsection{Results against the thesis}
\label{par:floor}
\def\wantedblock{length}\immediate\write\secposfile{block sec8_robustness length}
\def\wantedblock{sixteen}\immediate\write\secposfile{block appendix sixteen}
\def\wantedblock{budgetfig}\immediate\write\secposfile{block appendix budgetfig}
\def\wantedblock{againstmore}\immediate\write\secposfile{block sec8_robustness againstmore}
\subsection{Other robustness results and side effects}
\label{app:otherrobust}
\def\wantedblock{robustsum}\immediate\write\secposfile{block appendix robustsum}
\def\wantedblock{letterformat}\immediate\write\secposfile{block appendix letterformat}
\def\wantedblock{forgetrktab}\immediate\write\secposfile{block appendix forgetrktab}
\paragraph{Transfer to a neighbouring task.}
\label{app:transfer}
\def\wantedblock{transfer}\immediate\write\secposfile{block appendix transfer}

\section{Every setting and rank}
\label{app:rankexp}

\def\wantedblock{ranksum}\immediate\write\secposfile{block appendix ranksum}
\def\wantedblock{ocsum}\immediate\write\secposfile{block appendix ocsum}
\def\wantedblock{oclevelstab}\immediate\write\secposfile{block sec4_results oclevelstab}
\def\wantedblock{gsmtable}\immediate\write\secposfile{block sec4_results gsmtable}
\def\wantedblock{gsm8b}\immediate\write\secposfile{block sec4_results gsm8b}
\def\wantedblock{rankxvimore}\immediate\write\secposfile{block appendix rankxvimore}

\subsection{The diverged run at rank 16 on the format task}
\label{app:collapse}
\def\wantedblock{rankxvidetail}\immediate\write\secposfile{block sec4_results rankxvidetail}
\def\wantedblock{collapse}\immediate\write\secposfile{block appendix collapse}

\subsection{Where A \texorpdfstring{$-$}{-} B is not established at ranks 16 to 64}
\label{app:notest}
\def\wantedblock{notestsum}\immediate\write\secposfile{block appendix notestsum}
\def\wantedblock{latecvr}\immediate\write\secposfile{block appendix latecvr}

\section{How the OpenBookQA result depends on the protocol}
\label{app:supported}
\def\wantedblock{resultsprot}\immediate\write\secposfile{block sec1_intro resultsprot}

\section{The settings where the main result holds}
\label{app:holds}
\def\wantedblock{supporteddetail}\immediate\write\secposfile{block sec9_discussion supporteddetail}
\paragraph{Above rank 8.}
\def\wantedblock{opendetail}\immediate\write\secposfile{block sec9_discussion opendetail}
\def\wantedblock{costs}\immediate\write\secposfile{block sec9_discussion costs}
\def\wantedblock{costsdetail}\immediate\write\secposfile{block sec9_discussion costsdetail}
\def\wantedblock{costsend}\immediate\write\secposfile{block sec9_discussion costsend}
\def\wantedblock{practicaldetail}\immediate\write\secposfile{block sec9_discussion practicaldetail}
\def\wantedblock{resultsdetail}\immediate\write\secposfile{block sec1_intro resultsdetail}
\fi

\section{Deviation registry}
\label{app:deviations}
\def\wantedblock{deviationsintro}\immediate\write\secposfile{block appendix deviationsintro}
\def\wantedblock{deviationssum}\immediate\write\secposfile{block appendix deviationssum}
\begin{small}
\begin{longtable}{p{6cm}rrr}
\caption{\textbf{Learning-rate selections that ended on an unresolved tie or at the edge of the grid, in each experiment.} Unresolved tie: the two best rates stayed within 0.005 of validation accuracy after three more seeds. At the edge: the selected rate is the smallest or largest rate tried. Edge step: the grid was extended by one step, two in SGD and the LoRA-FA initialisation (\Cref{app:rules}). An edge step is counted whichever rate was then selected, so a row can have more edge steps than selections at the edge. At sixteen times the budget, three selections stayed at the edge without the extra step, kept and declared. The seven rows below the horizontal rule were added after this count was first made. \dag: chosen after related results (\Cref{tab:posthoc}).}\label{tab:ties}\\
\toprule
experiment & unresolved ties & at the edge & edge steps \\
\midrule
\endfirsthead
\toprule
experiment & unresolved ties & at the edge & edge steps \\
\midrule
\endhead
A frozen on activation bands & 0 & 2 & 2 \\
ARC, ranks 2 to 64 & 9 & 5 & 22 \\
alpha scaling, rank 16\dag & 5 & 0 & 0 \\
$\alpha/r$ sweep & 5 & 0 & 0 \\
rotation inside the band & 0 & 0 & 1 \\
core-only adapters & 4 & 1 & 5 \\
eight-task mixture, rank 16\dag & 4 & 0 & 0 \\
eight-task mixture, rank 32\dag & 3 & 0 & 0 \\
eight-task mixture, rank 2 & 2 & 0 & 1 \\
eight-task mixture, rank 8 & 3 & 0 & 0 \\
unfreezing a factor, rank 16\dag & 3 & 0 & 0 \\
training-data size & 3 & 0 & 4 \\
task and model crossed & 3 & 1 & 1 \\
ranks 16 and 32, replication\dag & 7 & 1 & 5 \\
GaLore projection side & 0 & 2 & 5 \\
instruction-tuned model & 3 & 0 & 1 \\
input-side overlap & 2 & 0 & 1 \\
LoRA-FA initialisation & 0 & 1 & 4 \\
LoRA+ & 1 & 0 & 0 \\
letter prompt & 1 & 0 & 2 \\
Mistral-7B, ranks 2 and 32 & 1 & 0 & 3 \\
Mistral-7B, rank 64\dag & 1 & 1 & 5 \\
module type & 0 & 0 & 1 \\
output complexity & 7 & 0 & 2 \\
output overlap, rank 2 & 2 & 0 & 0 \\
output source & 7 & 1 & 7 \\
output overlap, rank 16\dag & 2 & 0 & 0 \\
output overlap, rank 8\dag & 2 & 0 & 0 \\
Qwen3-1.7B-Base, rank 2 & 1 & 0 & 3 \\
rank 1 (format task\dag) & 3 & 1 & 2 \\
ranks 4 and 8 & 5 & 0 & 2 \\
OpenBookQA, ranks 2 to 64 & 9 & 1 & 2 \\
Qwen2.5-7B, rank 16\dag & 1 & 0 & 0 \\
Qwen2.5-7B, two more tasks\dag & 1 & 0 & 1 \\
output complexity, short budgets & 6 & 0 & 0 \\
SGD & 1 & 0 & 2 \\
format task, ranks 2 to 64 & 2 & 1 & 2 \\
square modules & 3 & 0 & 1 \\
factor stitching & 4 & 12 & 14 \\
\midrule
longer training at rank 16 & 4 & 0 & 1 \\
four times the budget, format task & 2 & 0 & 0 \\
sixteen times the budget, format task & 0 & 3 & 2 \\
GSM8K, Llama-3.2-1B & 2 & 0 & 0 \\
GSM8K, Llama-3.1-8B\dag & 2 & 0 & 0 \\
input-side test & 2 & 1 & 2 \\
full fine-tuning & 0 & 0 & 2 \\
\midrule
all experiments & 128 & 34 & 108 \\
\bottomrule
\end{longtable}
\end{small}

\iftmlr
\section{The diverged run at rank 16 on the format task}
\label{app:collapse}
\def\wantedblock{collapse}\immediate\write\secposfile{block appendix collapse}

\section{What a frozen \texorpdfstring{$B$}{B} needs: the overlap analysis}
\label{sec:mechanism}
\label{app:overlap}
\def\wantedblock{opening}\immediate\write\secposfile{block sec7_mechanism opening}\block{opening}
$B$ frozen does better the more its basis overlaps the output directions that a free LoRA learns on the same task.
On OpenBookQA at rank 2, the rank correlation between the overlap and accuracy is $+0.902$.
The overlap analysis is an oracle, not a method: the learned directions come from free LoRAs trained on the same data, which a method does not have.
\Cref{sec:discussion} proposes a hypothesis for why freezing $A$ costs little; its one test, on the input side, leaves it open. The learned input directions help on one pair only (\Cref{app:mech}).
\endblock
\block{figmech}
\begin{figure}[!t]
\centering
\paperfig[width=\linewidth]{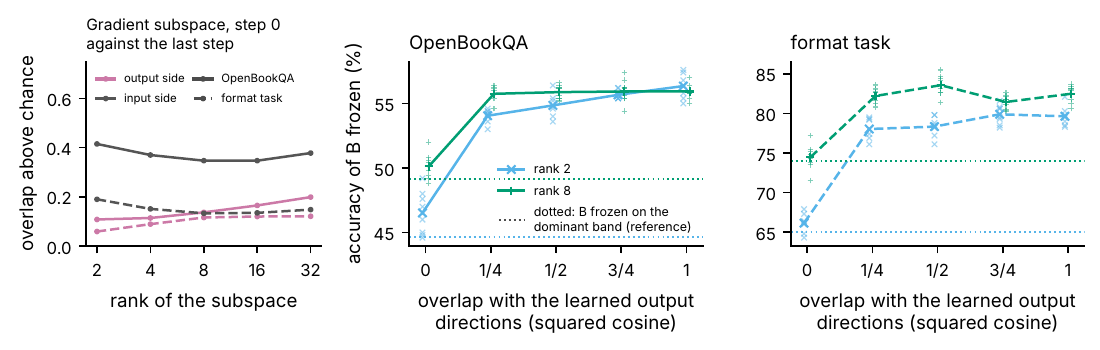}
\caption{\textbf{Accuracy of $B$ frozen rises with its overlap with the learned output directions.}
Middle and right: accuracy against the overlap (squared cosine $c^2$) between the frozen basis and the learned output directions $U_L$, from 0 (orthogonal) to 1 (the learned directions).
The panels show OpenBookQA with Qwen2.5-1.5B and the format task with Llama-3.2-1B, at rank 2 and at rank 8\posthoc.
Small markers: one run per seed, eight per level; lines: means.
Dotted lines: accuracy of $B$ frozen on the dominant band at the same rank, in the same colour, a reference level; the text measures how much of its gap to the matched free LoRA the overlap recovers.
Left\posthoc: how much of the leading subspace of the weight gradient is kept between step 0 and the last step, on the output side and on the input side, as the overlap above chance (1: unchanged; \Cref{app:gradsides}). One free LoRA per task.
The learned directions come from free LoRAs trained on the same data: an oracle construction, not a method.
\posthoc\ Post hoc and descriptive.}
\label{fig:mechanism}
\end{figure}
\endblock
\block{learned}
\paragraph{The learned output directions are the leading directions of a free LoRA's update.}
\endblock
\block{learnedconstruct}
For each main pair and module, we average the update $\Delta W = BA$ of the pair's highest-rank free LoRA over the eight seeds.
We keep its $r$ leading left singular vectors, where $r$ is the rank of the frozen $B$ we build.
\endblock
\block{learned2}
We call these vectors the \emph{learned output directions} ($U_L$).
They are a property of a trained model, so a method that freezes $B$ before training cannot know them.
\endblock
\block{overlap}
\paragraph{The more the frozen output basis overlaps the learned one, the better $B$ frozen does.}
\endblock
\block{overlapconstruct}
We freeze $B$ on $U(c) = c\,U_L + \sqrt{1 - c^2}\,U_\perp$, with $U_\perp$ a random basis orthogonal to $U_L$.
Every direction then makes the same angle with $U_L$, with squared cosine $c^2$, which we call the \emph{overlap}.
We use five levels from 0 to 1, with eight seeds each.
\endblock
\block{overlap2}
The test of this experiment was fixed in advance at rank 2.
At rank 2, accuracy rises with the overlap on both main pairs, and the rank correlation is established on both.
It is also established on both main pairs at rank 8 (post hoc) (\Cref{tab:tests}).
At rank 16 it is established on both main pairs too, $+0.544$ on OpenBookQA and $+0.509$ on the format task.
The recovery is step-shaped, not gradual.
On OpenBookQA at rank 2, an overlap of one quarter ($c^2 = 1/4$) already recovers a fraction $0.910$ of the accuracy gap.
This gap is the one between $B$ frozen on the dominant band and the matched free LoRA (descriptive; \Cref{fig:mechanism}).
Because we set the overlap, the analysis is an intervention, not a correlation between two observed quantities.
Neither band lies close to the output direction a free LoRA learns. Against the output direction of the matched free LoRA (rank 1, same seed), the frozen $B$ has a mean squared cosine of $0.004$ (dominant) and $0.012$ (minor) on OpenBookQA, and $0.002$ on both bands of the format task; averaged over modules, whose chance level $2/d_\text{out}$ differs, that is at most $2.4$ times chance; the $B$ trained with $A$ frozen reaches $0.205$ and $0.117$ on OpenBookQA (descriptive).
\endblock
\block{core}
\paragraph{With both outer factors frozen, the output basis again limits the adapter on the format task.}
Another intervention freezes both factors and trains only a $16 \times 16$ core between them, as LoRA-XS~\citep{balazy2024loraxs} does.
Each side is set to the spectral, a random or the learned basis.
The configurations with a learned basis are oracles in the same sense, built from free LoRAs that saw no test item.
The test fixed in advance compares two configurations.
One has the learned basis on the output side and the spectral basis on the input side; the other has the opposite.
On the format task with Llama-3.2-1B at rank 16, the configuration with the learned output basis beats the other one by $+7.31$ points.
On OpenBookQA, no difference is detected.
As fixed in advance, we therefore make the claim for the format task only (the other comparisons follow).
\endblock
\block{others}
\paragraph{The other experiments leave the advantage of A frozen unexplained.}
Unfreezing a factor shows that keeping $B$ frozen costs more than keeping $A$ frozen on both pairs and bands at rank 2.
The difference is about as large as A $-$ B itself.
Once both factors are trained, no difference is detected from which factor started on the band, and the cost of a frozen $B$ is not additive across module groups (\Cref{app:robust}).
With both oracles, no difference is detected between $A$ frozen on the learned input directions and $B$ frozen on the learned output directions.
\Cref{sec:discussion} gives a hypothesis for the advantage, which its one test leaves open, based on the gradient analysis of \Cref{app:gradsides}.
\endblock

\def\wantedblock{figmech}\immediate\write\secposfile{block sec7_mechanism figmech}
\def\wantedblock{learned}\immediate\write\secposfile{block sec7_mechanism learned}
\def\wantedblock{learnedconstruct}\immediate\write\secposfile{block sec7_mechanism learnedconstruct}
\def\wantedblock{learned2}\immediate\write\secposfile{block sec7_mechanism learned2}
\def\wantedblock{overlap}\immediate\write\secposfile{block sec7_mechanism overlap}
\def\wantedblock{overlapconstruct}\immediate\write\secposfile{block sec7_mechanism overlapconstruct}
\def\wantedblock{overlap2}\immediate\write\secposfile{block sec7_mechanism overlap2}
\def\wantedblock{overlapdetail}\immediate\write\secposfile{block appendix overlapdetail}

\section{Freezing \texorpdfstring{$A$}{A} where the inputs do not go}
\label{app:activation}
\def\wantedblock{activation}\immediate\write\secposfile{block appendix activation}
\else
\section{What a frozen \texorpdfstring{$B$}{B} needs: an overlap analysis}
\label{sec:mechanism}
\def\wantedblock{opening}\immediate\write\secposfile{block sec7_mechanism opening}
\def\wantedblock{figmech}\immediate\write\secposfile{block sec7_mechanism figmech}
\subsection{The learned output directions and the overlap}
\label{app:overlap}
\def\wantedblock{learned}\immediate\write\secposfile{block sec7_mechanism learned}
\def\wantedblock{learnedconstruct}\immediate\write\secposfile{block sec7_mechanism learnedconstruct}
\def\wantedblock{learned2}\immediate\write\secposfile{block sec7_mechanism learned2}
\def\wantedblock{overlap}\immediate\write\secposfile{block sec7_mechanism overlap}
\def\wantedblock{overlapconstruct}\immediate\write\secposfile{block sec7_mechanism overlapconstruct}
\def\wantedblock{overlap2}\immediate\write\secposfile{block sec7_mechanism overlap2}
\def\wantedblock{overlapdetail}\immediate\write\secposfile{block appendix overlapdetail}
\subsection{Core-only adapters}
\label{app:core}
\def\wantedblock{core}\immediate\write\secposfile{block sec7_mechanism core}
\subsection{Unfreezing a factor, and the input side}
\label{app:mech}
\def\wantedblock{mechsum}\immediate\write\secposfile{block appendix mechsum}
\subsection{Freezing \texorpdfstring{$A$}{A} where the inputs do not go}
\label{app:activation}
\def\wantedblock{activation}\immediate\write\secposfile{block appendix activation}
\subsection{Which side of the gradient is more reproducible}
\label{app:gradsides}
\def\wantedblock{gradsum}\immediate\write\secposfile{block appendix gradsum}
\def\wantedblock{lateig}\immediate\write\secposfile{block appendix lateig}
\fi

\section{Llama-3.1-8B and rate ties}
\label{app:ties}
\def\wantedblock{ratetiesdetail}\immediate\write\secposfile{block appendix ratetiesdetail}

\iftmlr
\section{Details of the mechanism experiments}
\label{app:mech}
\def\wantedblock{mechintro}\immediate\write\secposfile{block appendix mechintro}
\def\wantedblock{others}\immediate\write\secposfile{block sec7_mechanism others}
\def\wantedblock{mechsum}\immediate\write\secposfile{block appendix mechsum}
\def\wantedblock{lateig}\immediate\write\secposfile{block appendix lateig}

\section{Core-only adapters: the other comparisons}
\label{app:core}
\def\wantedblock{coreintro}\immediate\write\secposfile{block appendix coreintro}
\def\wantedblock{core}\immediate\write\secposfile{block sec7_mechanism core}
\def\wantedblock{coreother}\immediate\write\secposfile{block appendix coreother}

\section{Which side of the gradient is more reproducible}
\label{app:gradsides}
\def\wantedblock{gradsum}\immediate\write\secposfile{block appendix gradsum}

\section{How the configurations relate to MiCA and PiCa}
\label{app:methodarms}
\def\wantedblock{methodsum}\immediate\write\secposfile{block appendix methodsum}
\def\wantedblock{tabmethods}\immediate\write\secposfile{block sec6_methods tabmethods}\block{opening}
So far the ranks and recipes were ours.
Does the frozen factor still matter for MiCA and PiCa (\Cref{sec:configs}) at their own ranks and recipes, which start at rank 16?
\endblock
\block{setups}
\paragraph{MiCA is run in three setups, to separate MiCA's recipe from our code and our rates.}
(1) \emph{MiCA's recipe}: the PEFT library~\citep{mangrulkar2022peft} at rank 16, the rank MiCA uses for Llama-2, with its published learning rate and no rate selection\iftmlr\else\ (\Cref{tab:methods})\fi.
(2) \emph{Our protocol}: our own code, $B$ frozen on the minor band, and the rate selected per configuration (\Cref{sec:rates}).
(3) \emph{The tuned-rate check}: the PEFT library at rank 16, with a rate selected for each method.
\endblock
\block{setupsdetail}
MiCA runs on our seven adapted projections, not the query and value projections of its paper.
With these projections, MiCA differs from our configuration only in its recipe, and the matched free LoRA stays matched.
On MiCA's configuration at its recipe, our code and the PEFT implementation differ by less than one point (descriptive).
\endblock
\block{flipped}
\paragraph{On the format task, flipped MiCA beats MiCA (not Holm-robust).}
If the factor still matters at rank 16, as \Cref{sec:rank} suggests, MiCA should do better with its other factor frozen.
\emph{Flipped MiCA} tests this: it freezes $A$ on the minor band and trains with MiCA's recipe.
We compare it with MiCA at two matchings: the same rank 16, and about MiCA's number of trainable parameters.
\endblock
\block{flippeddetail}
\Cref{tab:methods} gives each method's parameters, so the first flipped row compares at equal rank and the second at about MiCA's parameter count (its rank and the ratios are given below).
Flipped MiCA runs in our code and MiCA in the PEFT library; the comparison of the two codes is above.
Both run without MiCA's adapter dropout of 0.05, so dropout is not part of the comparison.
Differences are computed on unrounded means.
\endblock
\block{flipped2}
On the format task with Llama-3.2-1B it does: flipped MiCA wins by $+3.48$ and $+3.50$ points at the two matchings (\Cref{tab:methods}).
Both results are established, and neither is Holm-robust.
On OpenBookQA with Qwen2.5-1.5B, the flip is inconclusive: we detect no difference, and the intervals allow at most $5.83$ points.
With Qwen2.5-7B on eight commonsense tasks, the flip changes almost nothing: $+0.37$ points, at most $1.05$ (\Cref{app:flip7}).
\iftmlr\else This experiment extends the rank-16 results of \Cref{sec:rank} (\Cref{tab:tests}).\fi
\endblock
\block{flippedfree}
Flipped MiCA beats MiCA on the format task, but the matched free LoRA, with MiCA's number of trainable parameters, stays above it (descriptive; \Cref{tab:methods}).
\endblock
\block{trails}
\paragraph{Under our protocol, MiCA's configuration trails the matched free LoRA in all seven comparisons, one of them Holm-robust.}
\Cref{tab:gap} gives the seven comparisons, on four models: the free LoRA leads by $+3.09$ to $+6.27$ points.
All seven intervals exclude zero, three of them barely.
Of the four in the Holm family, the main pairs at ranks 16 and 32, only OpenBookQA at rank 16 is Holm-robust.
\iftmlr\else The free LoRA has slightly fewer trainable parameters on OpenBookQA, which favours MiCA's configuration.
On the format task it has slightly more, and we do not separate the effect of size from that of the frozen factor there (\Cref{tab:g1}). \fi
The two other setups give the same sign, with gaps of other sizes\iftmlr\ (\Cref{app:methodarms}, which also gives the size differences)\fi.
\endblock
\block{trailsother}
With tuned rates in the PEFT library, the matched free LoRA leads MiCA on OpenBookQA at rank 16.
At MiCA's recipe, it leads on both tasks (descriptive, \Cref{tab:methods}).
\endblock
\block{trails2}
These tasks are not the setting in which MiCA was proposed (\Cref{sec:background}).
\endblock
\block{minorcost}
\paragraph{On OpenBookQA, we detect no difference between $B$ frozen on the dominant band and the matched free LoRA.}
Is that deficit due to MiCA's frozen factor or to its band?
On OpenBookQA, the evidence points to the band.
At ranks 16 to 64, $B$ frozen on the dominant band stays close to the matched free LoRA: the differences lie between $-1.62$ and $+1.40$ points, and none is established (last block of \Cref{tab:gap}).
With Qwen2.5-1.5B, $B$ frozen does better on the dominant band than on the minor band at ranks 16 and 32, by $+4.45$ and $+5.87$ points.
For MiCA, then, the factor matters on the format task with a 1B model (not Holm-robust), and the band matters on OpenBookQA.
With one 7B model on the eight-task mixture, we detect no effect of the factor in MiCA's recipe.
\endblock
\block{tabgap}
\begin{table}[!ht]
\centering
\small
\caption{\textbf{Under our protocol, the matched free LoRA beats MiCA's configuration in all seven comparisons; on OpenBookQA we detect no difference with $B$ frozen on the dominant band.}
Each row is the matched free LoRA minus MiCA's configuration, in accuracy points.
The first row is Holm-robust, rows 2 to 4 are not, and the 7B rows of this comparison are not part of the correction.
Barely: the lower bound is within 0.5 points of zero.
Last block: the same for $B$ frozen on the dominant band.}
\label{tab:gap}
\begin{tabular}{llrll}
\toprule
Task and model & rank & difference & interval & reading \\
\midrule
OpenBookQA, Qwen2.5-1.5B & 16 & $+5.85$ & $[+2.37, +9.38]$ & free LoRA ahead; Holm-robust \\
OpenBookQA, Qwen2.5-1.5B & 32 & $+4.25$ & $[+0.85, +7.90]$ & free LoRA ahead \\
Format task, Llama-3.2-1B & 16 & $+4.78$ & $[+1.89, +7.90]$ & free LoRA ahead \\
Format task, Llama-3.2-1B & 32 & $+3.09$ & $[+0.02, +6.03]$ & free LoRA ahead, barely \\
OpenBookQA, Qwen2.5-7B & 16 & $+6.27$ & $[+2.83, +9.67]$ & free LoRA ahead \\
OpenBookQA, Qwen2.5-7B & 32 & $+3.75$ & $[+0.35, +7.13]$ & free LoRA ahead, barely \\
OpenBookQA, Mistral-7B & 32 & $+3.90$ & $[+0.25, +7.70]$ & free LoRA ahead, barely \\
\midrule
\multicolumn{5}{l}{Free LoRA minus $B$ frozen on the dominant band} \\
OpenBookQA, Qwen2.5-1.5B & 16 & $+1.40$ & $[-1.68, +4.52]$ & interval includes zero \\
OpenBookQA, Qwen2.5-1.5B & 32 & $-1.62$ & $[-4.55, +1.28]$ & interval includes zero \\
OpenBookQA, Qwen2.5-1.5B & 64 & $-1.28$ & $[-4.02, +1.42]$ & interval includes zero \\
OpenBookQA, Mistral-7B & 32 & $+0.45$ & $[-2.20, +3.10]$ & interval includes zero \\
OpenBookQA, Mistral-7B & 64 & $-0.62$ & $[-3.32, +2.05]$ & interval includes zero \\
\bottomrule
\end{tabular}
\end{table}
\endblock
\block{pica}
\paragraph{PiCa beats the matched free LoRA only at its published learning rate, in two of three settings.}
For PiCa we ask whether its gain survives a comparison with a free LoRA of its size, at PiCa's published recipe.
We run it at rank 256, a rank used in PiCa's paper and above the ranks 1 to 64 of our design, and at MiCA's rank 16.
Here the matched free LoRA has about PiCa's number of trainable parameters.
\endblock
\block{picaimpl}
We run PiCa in our implementation, which we checked against the official code.
We use its published rate, schedule and batch.
\endblock
\block{picashared}
With sharing, PiCa at rank 256 trains the parameters of a LoRA of much lower rank (the counts are given below).
The update of each layer can still reach rank 256.
\endblock
\block{pica2}
At PiCa's published rate, which both methods use, we test three of the four settings.
OpenBookQA at rank 16 is left out, because the free LoRA gains less than 2 points over the base model there.
PiCa leads the matched free LoRA by $6.05$ points in two settings: OpenBookQA at rank 256, and the format task at rank 16.
The two experiments are separate, and their equal values are a coincidence.
On the format task at rank 256, we detect no difference (the interval runs from PiCa ahead by $4.23$ points to the free LoRA ahead by $2.70$).
With rates selected as in our protocol, on two seeds kept apart from the eight reported ones, we detect no difference in any setting (\Cref{app:picatable}).
\iftmlr\else The two sets of tests, at the published rate and at selected rates, share no correction for multiple tests.
Our rule, fixed in advance, is to state the weaker of the two when they differ. \fi
So PiCa's lead depends on the learning rate, in line with \mbox{\citet{lee2026learningrate}}, who find that LoRA variants reach about the peak accuracy of vanilla LoRA once the rate is tuned.
\endblock
\block{picasteps}
PiCa's recipe gives every configuration few optimiser steps: 15 on our packed OpenBookQA training set and 150 on the format task.
At 15 steps, in the rank-256 setting, the free LoRA still gains $+5.65$ points over the base model.
\iftmlr\else At rank 16 it ends below the base model ($-0.42$), so it does not learn the task and no test is read.
On ARC-Challenge at rank 256, its gain over the base model is $-1.03$ points (no test).\fi
\endblock
\block{pica3}
At PiCa's parameter count, on the dominant band, $A$ frozen beats $B$ frozen on the format task at rank 16 by $+6.78$ points at the published rate.
\iftmlr In every other case no difference is detected (\Cref{app:methodarms}).\fi
\endblock
\block{picaab}
With selected rates, the same comparison is not established (the interval is given with the other Part B results above).
At rank 256 no difference is detected on either pair (\Cref{tab:tests}).
\endblock
\block{pica4}
At PiCa's own rank 256, we detect no difference between PiCa and $B$ frozen on the dominant band at PiCa's parameter count, on either pair.
So at PiCa's rank and recipe, we do not claim that $A$ frozen beats $B$ frozen.
\endblock
\block{tabmethods}
\begin{table}[t]
\centering
\small
\caption{\textbf{At MiCA's recipe, the matched free LoRA scores highest, flipped MiCA next and MiCA lowest, on both tasks.}
Test accuracy (\%), mean over eight seeds, at the rank 16 MiCA uses for Llama-2 and at MiCA's published learning rate.
Flipped MiCA freezes $A$ on the minor band instead of $B$.
Only the differences between flipped MiCA and MiCA are tests (\Cref{sec:methods}): established on the format task, not on OpenBookQA.
The rest is descriptive.
OpenBookQA: Qwen2.5-1.5B, 500 test questions. Format task: Llama-3.2-1B, 614 test items.}
\label{tab:methods}
\begin{tabular}{lrrrr}
\toprule
& \multicolumn{2}{c}{OpenBookQA} & \multicolumn{2}{c}{Format task} \\
\cmidrule(lr){2-3}\cmidrule(lr){4-5}
Method (frozen factor, band) & parameters & accuracy & parameters & accuracy \\
\midrule
MiCA ($B$, minor) & 8,142,848 & 52.40 & 5,242,880 & 79.13 \\
Flipped MiCA, same rank ($A$, minor) & 10,321,920 & 54.37 & 6,029,312 & 82.61 \\
Flipped MiCA, MiCA's parameter count ($A$, minor) & 8,386,560 & 54.43 & 5,275,648 & 82.63 \\
Matched free LoRA (none frozen) & 8,078,336 & 55.90 & 5,636,096 & 84.39 \\
\bottomrule
\end{tabular}
\end{table}
\endblock
\block{scopedetail}
At rank 32, A $-$ B is not established on either band of OpenBookQA, and on the format task only the dominant band of the replication is established (post hoc).
At rank 64, A $-$ B is not established on either band of either pair.
\endblock
\block{pica4detail}
At rank 16, A $-$ B is established on the minor band of both main pairs (post hoc), suggestive on the format task.
On the dominant band, it is not established on OpenBookQA, and established (post hoc) on the format task but driven by one diverged run (\Cref{app:collapse}).
On OpenBookQA at rank 16, the matched free LoRA did not clear the 2-point floor at the published rate, so that regime is uninformative.
This is a run of PiCa's method and recipe in our setting, not a reproduction of its tables.
\endblock
\block{trailsdetail}
They cover OpenBookQA with Qwen2.5-1.5B and with Qwen2.5-7B, and the format task, at ranks 16 and 32.
The free LoRA has slightly fewer trainable parameters on OpenBookQA and slightly more on the format task (\Cref{tab:g1}: values and ratios).
On OpenBookQA, this size difference favours MiCA's configuration.
On the format task it favours the free LoRA, and we do not separate the effect of size from the effect of the frozen factor there.
\endblock
\block{picareading}
We read PiCa's gain at its published rate as under-training of the free LoRA, not as evidence about frozen factors.
PiCa trains for the same 15 or 150 steps, and our data do not show why so few steps would hold back the free LoRA and not PiCa.
They show only that the gap is no longer established once each configuration has its own selected rate (Part B).
\endblock
\block{gapcoinc}
\iftmlr\paragraph{Two values of \Cref{tab:gap} that recur.}\fi
Four values of \Cref{tab:gap} also appear as A $-$ B in \Cref{tab:results}. $+4.78$ (format task) is a coincidence, since it comes from another experiment (rank 8). $+6.27$ (Qwen2.5-7B, rank 16) recurs because $A$ frozen on the minor band and the free LoRA have the same mean there. $+3.90$ and $+1.40$ recur too, each time in a different setting.
Other values recur across experiments: $78.32$, $54.37$, $+7.93$, $+9.62$, $+10.32$ and $+11.07$ each come from two distinct readings, and we checked each pair against its logs.
Further repeats, such as $+8.30$, $+4.82$, $+3.48$, $+4.38$ and $+8.70$, also belong to different experiments.
In \Cref{tab:gap}, the Qwen2.5-1.5B rows at ranks 16 and 32 come from the replication, the rank-64 row from the first experiment.
\endblock

\def\wantedblock{picatable}\immediate\write\secposfile{block appendix picatable}
\def\wantedblock{lateflip}\immediate\write\secposfile{block appendix lateflip}
\fi

\iftmlr
\section{Further related work}
\label{app:related}
\def\wantedblock{bandsmore}\immediate\write\secposfile{block sec2_background bandsmore}\block{opening}
Two lines of work meet in frozen-subspace methods (\Cref{sec:configs}): one asks which factor to train, the other which band of the weight to use.
\iftmlr\Cref{app:related} covers VeRA, DoRA and small singular values.\else We review them in turn.\fi
\endblock
\block{factors}
\subsection{Earlier work freezes a random $A$, not a spectral subspace}
\citet{zhu2024asymmetry} showed that training $B$ matters more than training $A$, and that a random, untrained $A$ can perform close to a trained one.
LoRA-FA~\citep{zhang2023lorafa} freezes $A$ at its random initialisation to save memory, and FFA-LoRA~\citep{sun2024ffalora} does the same for stability in private federated learning.
These works compare a frozen random $A$ with LoRA; none of them tests a spectral band.
We do: we compare $A$ frozen with $B$ frozen on the same spectral subspace.
Two controls then ask whether the band is needed at all, by replacing it with random directions.
The first freezes $A$ at its random initialisation, as in LoRA-FA, and the second freezes $A$ or $B$ on random orthonormal bases, on the format task with eight seeds.
This second control is weaker, because its learning rates are fixed, not selected.
In both, $A$ frozen still beats $B$ frozen, by $+9.62$ to $+18.42$ points in the first and $+11.07$ in the second, so in these settings the advantage does not need a spectral band\iftmlr\ (\Cref{tab:controls})\else\ (\Cref{tab:controls,tab:results})\fi.
The band is not neutral, though: on the format task at rank 2, we observe that the minor band lowers the accuracy of $B$ frozen ($57.92\,\%$ against $64.96\,\%$) but not that of $A$ frozen ($78.32\,\%$ against $77.08\,\%$).
So choosing the band does not remove the deficit of $B$ frozen, and on this task the asymmetry of \citet{zhu2024asymmetry} also holds for frozen-subspace methods (\Cref{sec:rank} adds OpenBookQA).
\endblock
\block{bands}
\subsection{Frozen-subspace methods differ in which factor they freeze}
PiSSA~\citep{meng2024pissa} and MiLoRA~\citep{wang2024milora} initialise both factors on the dominant or the minor singular components of $W_0$ and train both, so the band only sets the starting point.
LoRA-RLPO and LoRA-RLMO~\citep{rlvrortho2026}, proposed for reinforcement learning with verifiable rewards, start only $A$ on the dominant or the minor right singular vectors, with $B = 0$, and also train both factors.
The methods below keep one factor frozen on a band (\Cref{sec:configs}), and they differ in the factor they keep.
MiCA~\citep{rudiger2026mica} reports that its configuration learns more new knowledge than LoRA.
We test it outside that setting, on our tasks and models.
PiCa~\citep{hwang2026pica} freezes $r$ leading singular vectors, which it calls $U_r$, and trains a factor shared across layers.
Its paper writes the layer as $Y = X(W_0 + U_r B)$, with the inputs $X$ multiplied from the left, so $U_r$ lies on the input side: in our notation, PiCa is $A$ frozen on the dominant band (\Cref{app:methodarms}).
LoRA-XS~\citep{balazy2024loraxs} freezes both factors on the dominant singular vectors and trains a small $r \times r$ core between the two frozen factors.
Its output-side factor is $U_r \Sigma_r$: the $r$ leading left singular vectors, scaled by the diagonal matrix $\Sigma_r$ of their singular values.
LoRA-SB~\citep{ponkshe2024lorasb} sets such frozen factors from an approximation of the first update of full fine-tuning.
\iftmlr\else\Cref{sec:mechanism} asks whether the input basis or the output basis limits such an adapter.\fi
\paragraph{What has been missing.}
No earlier study crosses the frozen factor with the band on the same tasks, models and training budgets.
Without that, a difference between two methods could come from the factor, from the band, or from both.
\endblock
\block{bandsmore}
VeRA~\citep{kopiczko2024vera} freezes random matrices shared across layers and trains scaling vectors.
DoRA~\citep{liu2024dora} trains both factors on the direction of the weight, with a separate magnitude.
Neither method raises the choice of a frozen factor.
Small singular values can matter for the behaviour of transformers~\citep{staats2025small}.
LoRA and full fine-tuning also change the singular structure of the weights differently~\citep{shuttleworth2024illusion}.
\endblock

\else
\section{How the configurations relate to MiCA and PiCa}
\label{app:methodarms}
\def\wantedblock{methodsum}\immediate\write\secposfile{block appendix methodsum}
\def\wantedblock{tabmethods}\immediate\write\secposfile{block sec6_methods tabmethods}
\paragraph{The seven comparisons of \texorpdfstring{\Cref{tab:gap}}{the gap table}.}
\def\wantedblock{trailsdetail}\immediate\write\secposfile{block sec6_methods trailsdetail}
\def\wantedblock{trails2}\immediate\write\secposfile{block sec6_methods trails2}
\def\wantedblock{g1}\immediate\write\secposfile{block tables/tab_contrasts g1}
\def\wantedblock{counts}\immediate\write\secposfile{block appendix counts}
\paragraph{A \texorpdfstring{$-$}{-} B at the ranks of MiCA and PiCa.}
\def\wantedblock{pica4detail}\immediate\write\secposfile{block sec6_methods pica4detail}
\def\wantedblock{scopedetail}\immediate\write\secposfile{block sec6_methods scopedetail}
\def\wantedblock{picatable}\immediate\write\secposfile{block appendix picatable}
\def\wantedblock{lateflip}\immediate\write\secposfile{block appendix lateflip}

\section{Further related work}
\label{app:related}
\def\wantedblock{bandsmore}\immediate\write\secposfile{block sec2_background bandsmore}
\fi

\section{Further experiments}
\label{app:pending}
\def\wantedblock{further}\immediate\write\secposfile{block appendix further}

\iftmlr
\section{Data, splits and budgets}
\label{app:data}
\def\wantedblock{datasum}\immediate\write\secposfile{block appendix datasum}
\def\wantedblock{budgetslowdetail}\immediate\write\secposfile{block sec3_protocol budgetslowdetail}
\def\wantedblock{tasks2why}\immediate\write\secposfile{block sec3_protocol tasks2why}
\def\wantedblock{tasks2detail}\immediate\write\secposfile{block sec3_protocol tasks2detail}
\def\wantedblock{scoring}\immediate\write\secposfile{block sec3_protocol scoring}
\def\wantedblock{letterprompt}\immediate\write\secposfile{block sec3_protocol letterprompt}
\def\wantedblock{budgetscounts}\immediate\write\secposfile{block sec3_protocol budgetscounts}
\def\wantedblock{budgetrule}\immediate\write\secposfile{block sec3_protocol budgetrule}
\def\wantedblock{reprodetail}\immediate\write\secposfile{block appendix reprodetail}

\section{Memory and time}
\label{app:memtime}
\def\wantedblock{memtime}\immediate\write\secposfile{block sec9_discussion memtime}
\else
\section{Data, splits and budgets}
\label{app:data}
\def\wantedblock{datasum}\immediate\write\secposfile{block appendix datasum}
\def\wantedblock{scoring}\immediate\write\secposfile{block sec3_protocol scoring}
\def\wantedblock{letterprompt}\immediate\write\secposfile{block sec3_protocol letterprompt}
\def\wantedblock{budgetslowdetail}\immediate\write\secposfile{block sec3_protocol budgetslowdetail}
\def\wantedblock{reprodetail}\immediate\write\secposfile{block appendix reprodetail}

\section{Memory and time}
\label{app:memtime}
\def\wantedblock{memtime}\immediate\write\secposfile{block sec9_discussion memtime}
\fi
\secmark{end}

\end{document}